\pdfoutput=1
\documentclass[times,twocolumn,final]{elsarticle}
\UseRawInputEncoding

\usepackage{medima}
\usepackage{framed,multirow}
\usepackage{amssymb}
\usepackage{latexsym}
\usepackage{subfigure}
\usepackage[T1]{fontenc}
\usepackage{graphicx}
\usepackage{multirow}
\usepackage{nicematrix}
\usepackage{url}
\usepackage{xcolor}
\usepackage{hyperref}
\hypersetup{
    colorlinks=true,
    linkcolor=blue,
    filecolor=magenta,      
    urlcolor=cyan,
    }

\journal{Medical Image Analysis}

\begin{document}

\verso{Li, Chen, Tang \textit{et~al.}}

\begin{frontmatter}

\title{Transforming medical imaging with Transformers? A comparative review of key properties, current progresses, and future perspectives}

\author[1]{Jun \snm{Li} \fnref{fn1}}
\author[2]{Junyu \snm{Chen} \fnref{fn1}}
\author[3]{Yucheng \snm{Tang} \fnref{fn1}}
\fntext[fn1]{Contribute equally to this work.}

\author[1]{Ce \snm{Wang}}

%\author[2]{Eric C. \snm{Frey}}
\author[3]{Bennett A. \snm{Landman}}

\author[4,1]{S. Kevin \snm{Zhou} \corref{cor1}}
\cortext[cor1]{Corresponding author. E-mail address: skevinzhou@ustc.edu.cn.}

\address[1]{Key Lab of Intelligent Information Processing of Chinese Academy of Sciences (CAS), Institute of Computing Technology, CAS, Beijing 100190, China}
\address[2]{Russell H. Morgan Department of Radiology and Radiological Science, Johns Hopkins Medical Institutes, Baltimore, MD, USA}
\address[3]{Department of Electrical and Computer Engineering, Vanderbilt University, Nashville, TN, USA}
\address[4]{School of Biomedical Engineering \& Suzhou Institute for Advanced Research Center for Medical Imaging, Robotics, and Analytic Computing \& Learning (MIRACLE), University of Science and Technology of China, Suzhou 215123, China}

\begin{abstract}
Transformer, one of the latest technological advances of deep learning, has gained prevalence in natural language processing or computer vision. Since medical imaging bear some resemblance to computer vision, 
it is natural to inquire about the status quo of Transformers in medical imaging and ask the question: can the Transformer models transform medical imaging? In this paper, we attempt to make a response to the inquiry. After a brief introduction of the fundamentals of Transformers, especially in comparison with convolutional neural networks (CNNs), and highlighting key defining properties that characterize the Transformers, we offer a comprehensive review of the state-of-the-art Transformer-based approaches for medical imaging and exhibit current research progresses made in the areas of medical image segmentation, recognition, detection, registration, reconstruction, enhancement, etc. In particular, what distinguishes our review lies in its organization based on the Transformer's key defining properties, which are mostly derived from comparing the Transformer and CNN, and its type of architecture, which specifies the manner in which the Transformer and CNN are combined, all helping the readers to best understand the rationale behind the reviewed approaches. We conclude with discussions of future perspectives.
\end{abstract}


\end{frontmatter}

\section{Introduction} \label{sec.intro} 
%(1 pages, 10 references, SK Zhou)
Medical imaging~\citep{beutel2000handbook} is a non-invasive technology that acquires signals by leveraging the physical principles of sound, light, electromagnetic wave, etc., from which visual images of internal tissues of the human body are generated. There are many widely used medical imaging modalities, including ultrasound, digital radiography, computed tomography (CT), magnetic resonance imaging (MRI), and optical coherent tomography (OCT).  According to a report published by EMC\footnote{``The Digital Universe Driving Data Growth in Healthcare,'' published by EMC with research and analysis from IDC (12/13).}, about 90\% of all healthcare data are medical images, which undoubtedly become a critical source of evidence for clinical decision making, such as diagnosis and intervention.

Artificial intelligence (AI) technologies that process and analyze medical images have gained prevalence in scientific research and clinical practices in recent years~\citep{zhou2019handbook}. This is mainly due to the surge of deep learning (DL)~\citep{lecun2015deep}, which has achieved superb performances in a multitude of tasks, { including classification \citep{he2016deep, hu2018squeeze, huang2017densely}, object detection \citep{girshick2014rich, wang2017fast}, and semantic segmentation \citep{zhao2017pyramid, chen2017deeplab}}. The convolutional neural networks (CNNs or ConvNets) are DL methods customarily designed for image data. {The earliest applications of CNNs in medical imaging go back to the 1990s \citep{lo1995artificialA, lo1995artificialB, sahiner1996classificationC}. Though they showed encouraging results, it was not until the last decade that CNNs began to exhibit state-of-the-art performances and widespread deployment in medical image analysis. Ever since U-Net \citep{ronneberger2015u} won the 2015 ISBI cell tracking challenge, CNNs have taken the medical image analysis research by storm. Up till today, U-Net and its variants continue to demonstrate outstanding performance in many fields of medical imaging \citep{isensee2021nnu, zhou2022dudodr, cui2019pet}. The recurrent neural networks (RNNs) \citep{zhou2019handbook} and deep reinforcement learning (DRL) \citep{zhou2021deep} are employed for medical image analysis. More recently, Transformer~\citep{vaswani2017attention} has shown great potential in medical imaging applications as it has flourished in natural language processing and is flourishing in computer vision. }

Regarding homogeneity and heterogeneity of natural and medical images representations, it is motivated to investigate the status quo of Vision Transformer for medical imaging. It remains unclear whether Vision Transformers are better than CNNs for understanding medical images, and whether Transformers can transform medical imaging? {In this paper, we highlight the properties of Vision Transformers and present a comparative review for Transformer-based medical image analysis. Given that, the survey is confined to Vision Transformer, Unless stated otherwise, "Transformer" and "Transformer-based" referred in this paper represents "Vision Transformer", models with vanilla Language Transformer base blocks integrated, and applied in image analysis tasks.
}

%After a brief introduction of the fundamentals of Transformer, especially in comparison with convolutional neural networks, and highlighting key defining properties that characterize the Transformer, we offer a comprehensive review of the state-of-the-art Transformer-based approaches for medical imaging and exhibit current research progresses made in the areas of medical image segmentation, recognition, registration, reconstruction, enhancement, etc. In particular, what distinguishes our review lies in its organization based on the Transformer's key defining properties and its wide use of comparison as a means of explanation, both helping the readers to best understand the rationale behind the reviewed approaches.  We conclude with discussions of future perspectives.

%\subsection{paper organization} %(1 table)
%We start with providing our readers basic knowledge about of the principle of Transformer and then proceed with a thorough coverage of the latest of how Transformer is used in medical imaging. 

We organize the rest of
paper to include the following: {(i) a brief introduction to CNN and RNN for medical image analysis; (Section~\ref{sec.cnnrnn})} (ii) an introduction to Transformer with its general principle, key properties, and its main differences from a CNN (Section~\ref{sec.theory}); (iii) current progresses of state-of-the-art Transformer methods for solving medical imaging tasks, including medical image segmentation, recognition, classification, detection, registration, reconstruction, and enhancement, which is the main part (Section~\ref{sec.progress}); (iv) yet-to-solve challenges and future potential of Transformer in medical imaging (Section~\ref{sec.future}).

\section{CNN and RNN for Medical Image Analysis} \label{sec.cnnrnn}
\subsection{CNNs for medical imaging}
{We begin by briefly outlining the applications of CNNs in medical imaging and discussing their potential limitations. CNNs are specialized in analyzing data with a known grid-like topology (e.g., images). This is due to the fact that the convolution operation imposes a strong prior on the weights, compelling the same weights to be shared across all pixels. As the exploration of deep CNN architectures has intensified since the development of AlexNet for image classification in 2012 \citep{AlexNet12}, the first few successful efforts at deploying CNNs for medical imaging lay in the application of medical image classifications. These network architectures often begin with a stack of convolutional layers, pooling operations, and follow by a fully connected layer for producing a vector reflecting the probability of belonging to a certain class \citep{roth2014new, roth2015anatomy, cirecsan2013mitosis, brosch2013manifold, xu2014deep, malon2013classification, cruz2013deep, li2014deep}. In the meanwhile, similar architectures have been used for medical image segmentation \citep{ciresan2012deep, prasoon2013deep, zhang2015deep, xing2015automatic, vivanti2015automatic} and registration \citep{wu2013unsupervised, miao2016cnn, simonovsky2016deep} by performing the classification task on a pixel-by-pixel basis.}

{
In 2015, Ronneberger et al. introduced U-Net \citep{ronneberger2015u}, which is built based on the concept of the fully convolutional network (FCN) \citep{long2015fully}. In contrast to previous encoder-only networks, U-Net employs a decoder composed of successive blocks of convolutional layers and upsampling layers. Each block upsamples the previous feature maps such that the final output has the same resolution as the input. U-Net represents a substantial advance over previous networks. First, it eliminated the need for laborious sliding-patch inferences by having the input and output be full-sized images. Moreover, because the input to the network is a full-sized image as opposed to a small patch, U-Net has a better understanding of contextual information presented in the input. Although many other CNN architectures have demonstrated superior performances (e.g., HyperDense-Net \citep{dolz2018hyperdense} and DnCNN \citep{zhang2017beyond, cheng2019dilated, kim2018penalized}), the U-Net-like encoder-decoder paradigm has remained the \textit{de facto} choice when it comes to CNNs for pixel-level tasks in medical imaging. Many variants of such a kind have been proposed and demonstrated promising results on various applications, including segmentation \citep{isensee2021nnu, zhou2018unet++, oktay2018attention, gu2020net, zhang2020hifunet}, registration \citep{balakrishnan2019voxelmorph, dalca2019unsupervised, zhao2019unsupervised, zhao2019recursive}, and reconstruction \citep{han2018framing, cui2019pet}.}

{
Despite the widespread success of CNNs in medical imaging applications over the last decade, there are still inherent limitations within the architecture that prevent CNNs from reaching even greater performance. The vast majority of current CNNs deploy rather small convolution kernels (e.g., $3\times3$ or $5\times5$). Such a locality of convolution operations results in the CNNs being biased toward local spatial structures \citep{zhou2021convnets, naseer2021intriguing, dosovitskiy2020image}, which makes them less effective at modeling the long-range dependencies required to better comprehend the contextual information presented in the image. Extensive efforts have been made to address such limitations by expanding the theoretical receptive fields (RFs) of CNNs, with the most common methods including increasing the depth of the network \citep{SimonyanZ14a}, introducing recurrent- \citep{liang2015recurrent} or skip-/residual-connections \citep{he2016deep}, introducing dilated convolution operations \citep{Yu16multi, devalla2018drunet}, deploying pooling and up-sampling layers \citep{ronneberger2015u, zhou2018unet++}, as well as performing cascaded or two-stage framework \citep{isensee2021nnu, gao2019focusnet, gao2021focusnetv2}. Despite these attempts, the first few layers of CNNs still have limited RFs, making them unable to explicitly model the long-range spatial dependencies. Only at the deeper layers can such dependencies be modeled implicitly. However, it was revealed that as the CNNs deepen, the influence of faraway voxels diminishes rapidly \citep{luo2016understanding}. The effective receptive fields (ERFs) of these CNNs are, in fact, much smaller than their theoretical RFs, even though their theoretical RFs encompass the entire input image.
}

\subsection{RNN's role for medical image analysis}
{RNNs are originated to solve sequence modeling tasks such as language models or longitudinal/spatial-temporal data. The pioneering study long short-term memory (LSTM)~\citep{hochreiter1997long} calculate gradients with the gates module and loops. In medical image analysis, LSTM and RNN-based models~\citep{zhang2018segmentation} are used to handle segmentation, classification, and detection tasks. Distance-LSTM~\citep{gao2019distanced}, capable of modeling time distances between longitudinal scans, is good at learning intra-scan feature variabilities. The ResUNet with RNN model~\citep{alom2018recurrent}, aims at segmentation task, introduces a feature accumulation module for improving feature representations.~\citep{gao2018fully} combined CNNs with LSTM to learn spatial temporal representations of brain MRI slices. And~\citep{bai2018recurrent} captures aortic sequences for segmentation by fusing FCN with LSTM. 
Comparing to CNN and Transformers, RNNs have special ability for modeling medical images, such as capturing global spatial-temporal feature relationships. The major limitation for using RNNs are that learning 3D medical data is resources intensive, RNNs demand high-resolution for capturing pixel-wise variations. Therefore, RNN performs as a unique tool for specific tasks when understanding sequential data (e.g., longitudinal data presents).}

\subsection{Motivations behind using Transformers}
{Transformers, as alternative network architecture to CNNs, has recently demonstrated superior performances in many computer vision tasks \citep{dosovitskiy2020image, liu2021swin, wu2021cvt, zhu2020deformable, wang2021pyramid, chu2021twins, yuan2021tokens, dong2022cswin}. The core element of Transformers is the self-attention mechanism, which is not subject to the same limitations as convolution operations, making them better at capturing explicit long-range dependencies\citep{article}. Transformers have other appealing features, such as they scale up more easily \citep{liu2022convnet} and are more robust to corruption \citep{naseer2021intriguing}. Additionally, their weak inductive bias enables them to achieve better performance than CNNs with the aid of large-scale model sizes and datasets \citep{liu2022convnet, Zhai2022Scaling, dosovitskiy2020image, raghu2021vision}. Existing Transformer-based models have shown encouraging results in several medical imaging applications \citep{chen2021transunet, hatamizadeh2022unetr, chen2021transmorph, zhang2021transct}, prompting a surge of interest in further developing such models~\citep{shamshad2022transformers, liu2022medical, parvaiz2022vision, matsoukas2021time}.
%(as shown in Fig. \ref{fig:miccai_papers}). 
This paper provides an overview of Transformer-based models developed for medical imaging applications and highlights their key properties, advantages, shortcomings, and future directions. In the next section, we briefly review the fundamentals of Transformers.}

\section{Fundamentals of Transformer}   \label{sec.theory} 
%(1.5pages, 20 refs, Jun Li/Yucheng)
 %1page

{Language Transformer}~\citep{vaswani2017attention} is a neural network based on self-attention mechanisms and feed-forward module to compute representations and global dependencies. Recently, large {Language Transformer} models employed self-supervised pre-training has demonstrated improved efficiency and scalability, such as BERT~\citep{devlin2018bert} and GPT~\citep{radford2018improving, radford2019language, brown2020language} in natural language processing (NLP). In addition, Vision Transformer (ViT)~\citep{dosovitskiy2020image} partition and flatten images to sequences and implement Transformer for modeling visual features in a sequence-to-sequence paradigm. 
%The self-attention mechanism in Transformer shows a greater potential in vision tasks when compared with a CNN.
Below, we first give a detailed introduction to Vision Transformer, focusing on self-attention and its general pipeline. Next, we summarize the characteristics of convolution and self-attention and how the two interact. Lastly, we include key properties of Transformer from manifold perspectives.

\begin{figure}
\centering
\includegraphics[width=0.45\textwidth]{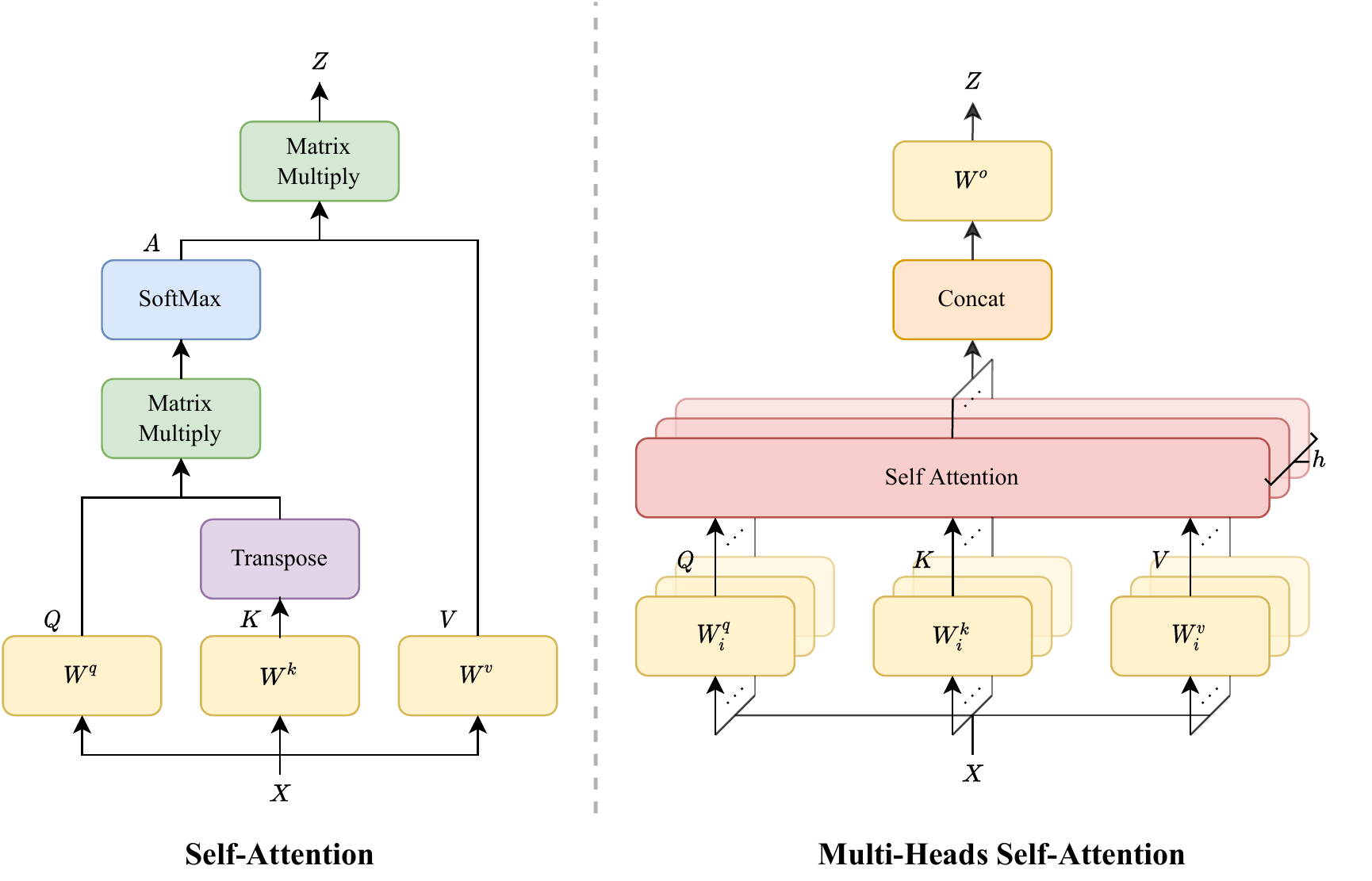}
\caption{Details of a self-attention mechanism (left) and a multi-head self-attention (MSA) (right). Compared to self-attention, the MSA conducts several attention modules in parallel. The independent attention features are then concatenated and linearly transformed to the output.} \label{fig:self-attention}
\end{figure}

\subsection{Self-attention in Transformer}
Humans choose and pay \textit{attention} to part of the information unintentionally when observing, learning and thinking. The attention mechanism in neural networks is a mimic to this physiological signal processing process~\citep{bahdanau2014neural}. A typical attention function computes a weighted aggregation of features, filtering and emphasizing the most significant components or regions ~\citep{bahdanau2014neural, xu2015show, dai2017deformable, hu2018squeeze}.

\subsubsection{Self-attention}
Self-attention (SA)~\citep{bahdanau2014neural} is a variant of attention mechanism (Figure~\ref{fig:self-attention} (left)), which is designed for capturing the internal correlation in data or features. Firstly, it maps the input $X \in \mathbb{R}^{n \times c}$ into a query $Q \in \mathbb{R}^{n \times d}$, a key $K \in \mathbb{R}^{n \times d}$, and a value $V \in \mathbb{R}^{n \times d}$, using three learnable parameters $W^q$, $W^k$, and $W^v$, respectively:
\begin{equation}
\begin{aligned}
    Q = X \times W^q, \quad W^q \in \mathbb{R}^{c \times d}, \\
    K = X \times W^k, \quad W^k \in \mathbb{R}^{c \times d}, \\
    V = X \times W^v, \quad W^v \in \mathbb{R}^{c \times d}. \\
\end{aligned}
\label{eq:qkv}
\end{equation}
Then, the similarity and correlation between query $Q$ and key $K$ is normalized, attaining an attention distribution $A \in \mathbb{R}^{n \times n}$:
\begin{equation}
    A(Q, K) = {\rm Softmax} (\frac{Q\times K^\top}{\sqrt{d}}).
\label{eq:self-att}
\end{equation}
The attention weight is applied to value $V$, giving the output $Z\in \mathbb{R}^{n\times d} $ of a self-attention block:
\begin{equation}
    Z = {\rm SA}(Q, K, V) = A(Q, K) \times V.
\end{equation}
In general, the key $K$ acts as an embedding matrix that ``memorizes" data, and the query $Q$ is a look-up vector. The affinity between the query $Q$ and the corresponding key $K$ defines the attention matrix $A$. The output $Z$ of a self-attention layer is computed as a sum of value $V$, weighted by $A$. The matrix $A$ calculated in~(\ref{eq:self-att}) connects all elements, thereby leading to a good capability of handling long-range dependencies in both NLP and CV tasks.

%attention usually applied on output of a hidden layer, while SA focus more on input and their relations, output after utilizing this information, and this is why the paper is named by attention is all you need.

%query works like an embedding matrix represent patterns from current data; key works as an embedding matrix of templates memorised from dataset ; current data exploit query to retrieve the most relative key in templates, resulting a weight vector then applied in value of itself.
%gradually mapping semantics relation in context. 
%enlarge parameters to a more robust transformation of features. 
%O(1) distance between elements.

\subsubsection{Multi-head self-attention (MSA)}
Multiple self-attention blocks, namely multi-head self-attention (Figure~\ref{fig:self-attention} (right)), are performed in parallel to produce multiple output maps. The final output is typically a concatenation and projection of all outputs of SA blocks, which can be given by:
\begin{equation}
\begin{aligned}
     Z_i = {\rm SA}(X\times W^q_i, X\times W^k_i, X\times W^v_i), \\
     {\rm MSA}(Q,K,V) = {\rm Concat}[Z_1,~\dots,~Z_{h}] \times W^o.
\end{aligned}
\end{equation}
where $h$ denotes the total number of heads and $W^o \in \mathbb{R}^{hd\times c}$ is a linear projection matrix, aggregating the outputs from all attention heads. $W^q_i$, $W^k_i$ and $W^v_i$ are parameters of the $i^{th}$ attention head. MSA projects $Q$, $K$ and $V$ into multiple sub-spaces that compute similarities of context features. Note that it is not necessarily true that a larger number of heads accompanies with better performance~\citep{voita2019analyzing}.

%“Multi-head attention allows the model to jointly attend to information from different representation subspaces at different positions." divide query key and value into multiple sub-space. independent sub models learn different types of patterns, and then aggregates them. not the more heads, better the performance.

\begin{figure}
\centering
\includegraphics[width=0.45\textwidth]{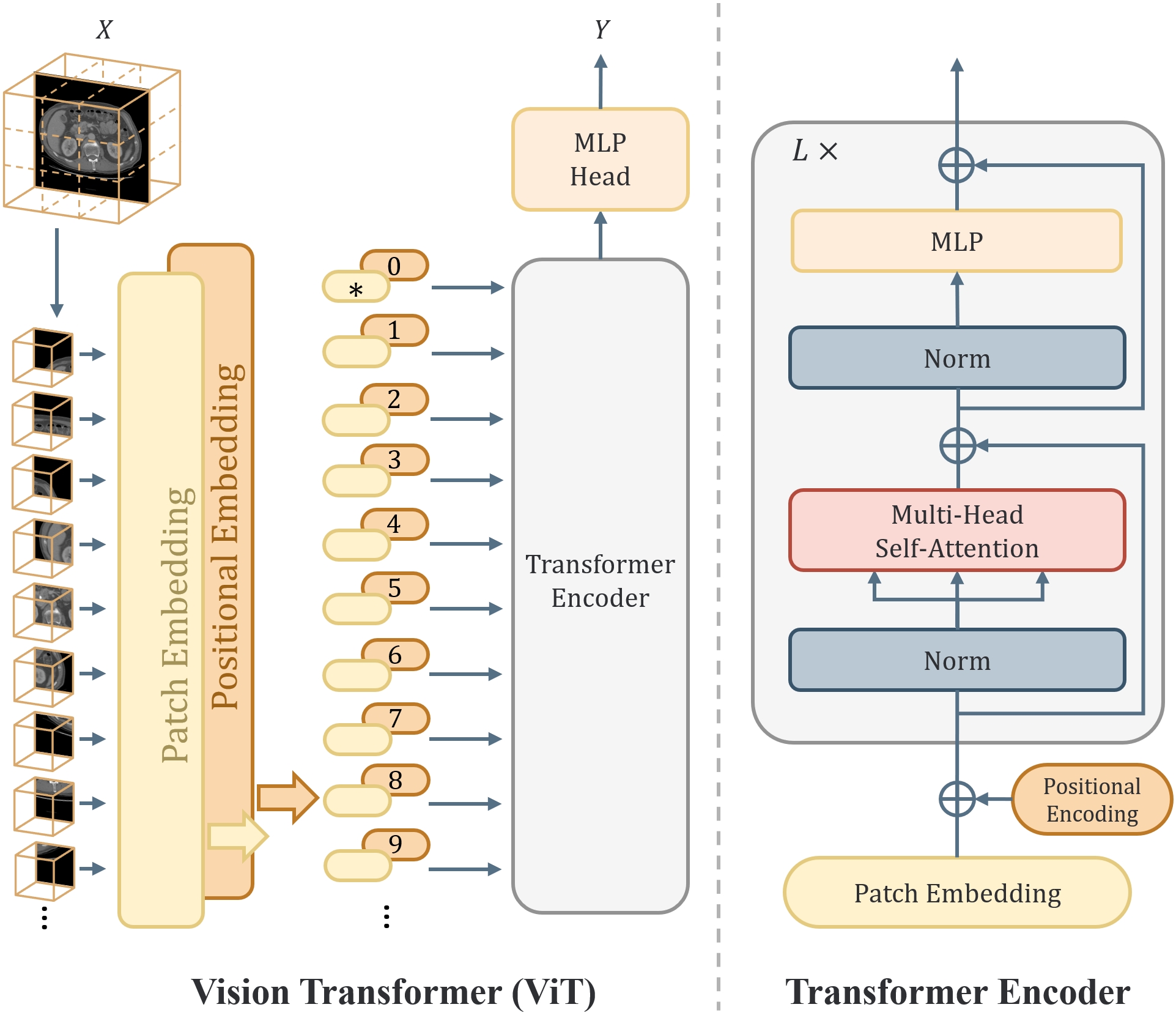}
\caption{Overview of Vision Transformer (left) and illustration of the Transformer encoder (right). The partitioning strategy splits an image into several fixed-size patches, and modeled as sequences with efficient Transformer implementation from NLP.} \label{fig:vit-pipeline}
\end{figure}

\subsection{Vision Transformer pipeline}
%nlp to cv, introduce the figure of pipeline. image -> patches -> embedding & position embedding -> N x Transformer block -> MLP head and the output.
% the basic Transformer block & embedding
% discussions on patch
% discussions on pe
\subsubsection{Overview}
A typical design of a Vision Transformer consists of a Transformer encoder and a task-specific decoder, depicted in Figure~\ref{fig:vit-pipeline} (left). Take the processing of 2D images for instance. Firstly, the image $X\in\mathbb{R}^{C\times H\times W }$ is split into a sequence of $N$ non-overlapping patches $\{X_1,X_2,\dots,X_N\}; ~X_i \in \mathbb{R}^{C \times P \times P}$, where $C$ is the number of channels, $[H, W]$ denotes the image size, and $[P, P]$ is the resolution of a patch.
{
Next, each patch is vectorized and then linearly projected into tokens:
\begin{equation}
    \hat{\mathbf{x}} = \{X_1\mathbf{E},X_2\mathbf{E}, \dots,X_N\mathbf{E}\},\ \mathbf{E}\in \mathbb{R}^{CP^2\times D},
\end{equation}
where $D$ is the embedding dimension. Then, a positional embedding, $\mathbf{E}_{pos}$, is added so that the patches can retain their positional information:
\begin{equation}
\label{eqn:token}
    \mathbf{x} = \hat{\mathbf{x}}+\mathbf{E}_{pos},\ \mathbf{E}_{pos}\in \mathbb{R}^{N\times D}.
\end{equation}
}%where $N$ denotes the number of patches.} 
The resulting tokens are fed into a Transformer encoder as shown in Figure~\ref{fig:vit-pipeline} (right), which consists of $L$ stacked base blocks. {Each base block consists of a multi-head self-attention and a multi-layer perceptron (MLP), with Layer-Norm (LN).} The feature can be formulated as:
\begin{equation}
    \begin{aligned}
        Z^\prime_l &= {\rm MSA}({\rm LN}(Z_{l-1})) + Z_{l-1}, \quad &l\in [1,\dots, L], \\
        Z_l &= {\rm MLP}({\rm LN}(Z^\prime_l)) + Z^\prime_l, \quad &l\in [1,\dots, L].
    \end{aligned}
\end{equation}

\subsubsection{Non-overlapping patch generation}
ViT adapts a standard Transformer in vision tasks, with the fewest modifications as possible. Therefore, the patches $\{X_1,~\dots,~X_n\}$ are generated in a non-overlapping style. On one hand, non-overlapping patches partially break the internal structure of an image~\citep{han2021Transformer}. MSA blocks integrate information from various patches, alleviating this problem. On the other hand, there is no computational redundancy when feeding non-overlapping patches into Transformer.

% patch: 
% trade-off of quadratic computation requirement.

% overlap or no-overlap patches:
% directly applying a standard Transformer to images, with fewest possible modifications; 
% no computation redundancy when no-overlap; 
% MSAs fuse information from various patches, (Early Convolutions Help Transformers See Better);
% inject pixel information is helpful to fix image structures cut out by patches (Transformer in Transformer.);
% smaller patch works better but computationally more expensive.
\subsubsection{Positional embedding}
%While multi-head self-attention is permutation-invariant, the relative position of patch embeddings is not considered. The absence of positional information leads a performance drop~\citep{dosovitskiy2020image}. A learnable parameter is typically empoyed to memorize position information for the partitioned sequences, the resulting feature is embedded as vectors that serves as input to the encoder ~\citep{dosovitskiy2020image,chu2021conditional}. For instance, the 1D position embedding considers the partitioned image as a sequence of patches in the raster order. The 2D learnable embedding takes the grid of patches in two dimensions. The relative positional embeddings model the relative distance among patches to maintain the spatial information. 
{Transformers tokenize and analyze each patch individually, resulting in the loss of positional information on each patch in relation to the whole image, which is undesired given that the position of each patch is imperative for comprehending the context in the image. Positional embeddings are proposed to encode such information into each patch such that the positional information is preserved throughout the network. Moreover, positional embeddings serve as the manually introduced inductive bias in Transformers. In general, there are three types of positional embedding: sinusoidal, learnable, and relative. The first two encode absolute positions from 1 to the number of patches, while the last encodes relative positions/distances between patches. In the following subsections, we briefly introduce each of the positional embeddings.}
\paragraph{Sinusoidal positional embedding} {To encode the position of each patch, we might intuitively assign an index value between 1 and the total number of patches to each patch. Yet, an obvious issue arises: if the number of patches is large, there may be a significant discrepancy in the index values, which hinders network training. Here, the key idea is to represent different positions using sinusoids of different wavelengths. For each patch position $n$, the sinusoidal positional embedding is defined as \citep{vaswani2017attention}:
\begin{equation}
    \begin{split}
        \mathbf{E}_{sin}(n, 2d)&={\rm sin}(\frac{n}{10000^{2d/D}})\\
        \mathbf{E}_{sin}(n, 2d+1)&={\rm cos}(\frac{n}{10000^{2d/D}})
    \end{split},
\end{equation}
where $d=1,\dots,\lfloor\frac{D}{2}\rfloor$.
}
\paragraph{Learnable positional embedding} {Instead of encoding the exact positional information onto the patches, a more straightforward way is to deploy a learnable matrix, $\mathbf{E}_{lrn}$, and let the network learn the positional information on its own. This is known as the learnable positional embedding.}
\paragraph{Relative positional embedding} {Contrary to using a fixed embedding for each location, as is done in sinusoidal and learnable positional embeddings, relative positional embedding encodes the relative information according to the offset between the elements in $Q$ and $K$ being compared in the self-attention mechanism \citep{raffel2020exploring}. Many relative positional embedding approaches have been developed, and this is still an active field of research \citep{Shaw2018self, raffel2020exploring, Dai2019TXL, huang-etal-2020-improve, wang2020axial, wu2021rethinking}. However, the basic principle stays the same, in which they encode information about the relative position of $Q$, $K$, and $V$ through a learnable or hard-coded additive bias during the self-attention computation.
}

% position encoding:
% learn-able parameters;
% information is incorporated as fixed [52] or trainable [18] positional embeddings. They are added before the first Transformer block to the patch tokens, which are then fed to the stack of Transformer blocks.
% an inductive bias about the 2D structure of the images is manually injected; no PE leads performance drop (DeiT experiments)
% a word is a clear semantic element in NLP tasks; (Conditional Positional Encodings for Vision Transformers)

\subsubsection{Multi-layer perceptrons}
{In the conventional Transformer design (e.g., the original ViT \citep{dosovitskiy2020image} and Transformer \citep{vaswani2017attention}), the MLP comes after each self-attention module. MLP is a crucial component since it injects inductive bias into Transformer, while the self-attention operation lacks inductive bias. This is because MLP is local and translation-equivariant, but self-attention computation is a global operation. The MLP is comprised of two feed-forward networks with an activation (typically a GeLU) in between:
\begin{equation}
    {\rm MLP}(x)=\phi(xW_1+b_1)W_2+b_2,
\end{equation}
where $x$ denotes the input, and $W$ and $b$ denote, respectively, the weight matrix and the bias of the corresponding linear layer. The dimensions of the weight matrices, $W_1$ and $W_2$, are typically set as $D\times4D$ and $4D\times D$ \citep{dosovitskiy2020image, vaswani2017attention}. Since the input is a matrix of flattened and tokenized patches ({\it i.e.}, Eqn. (\ref{eqn:token})), applying $W$ to $x$ is analogous to applying a convolutional layer with a kernel size of $1\times 1$. Consequently, the MLPs in the Transformer are highly localized and and equivariant to translation.  
}

\subsection{Transformer vs. CNNs} % 0.5page
{CNNs provide promising results for image analysis, while Vision Transformer has shown comparable even superior performance when pre-training or scaled datasets are available~\citep{dosovitskiy2020image}. This raises a question on the differences about how Transformers and CNNs understand images.
The receptive field of CNNs gradually expands when the nets go deeper, therefore the features extracted in lower stages are quite different from those in higher stages~\citep{raghu2021vision}. Features are analyzed and represented layer-by-layer, with global information injected. Besides, the increasing receptive field size of neurons and the pooling operations brings equivalence and invariance in translation ~\citep{jaderberg2015spatial, kauderer2017quantifying}, which empowers CNNs to exploit samples and parameters more effectively. Beyond that, the locality and weight sharing confers CNNs the advantages in capturing local structures. Considering the limited receptive field, CNNs are limited in catching long-distance relationships among image regions.
In Transformer model, the MSA provides a global receptive field even with the lowest layer of ViT, resulting in similar representations in different number of blocks~\citep{raghu2021vision}. The MSA block of each layer is capable of aggregating features in a global perspective, reaching a good understanding of long-distance relationships. The 16 by 16 sequences length is in natural large receptive field that can lead to better global feature modeling. In 3D transformers for volumetric data, this advantage is even obvious, the use of patch size 16$\times$16$\times$16 is intuitive and beneficial for high dimensional, high resolution medical images, as anatomical context are crucial for medical deep learning.}

\subsubsection{Combining Transformer and CNN}
\begin{figure}
\centering
\includegraphics[width=0.48\textwidth]{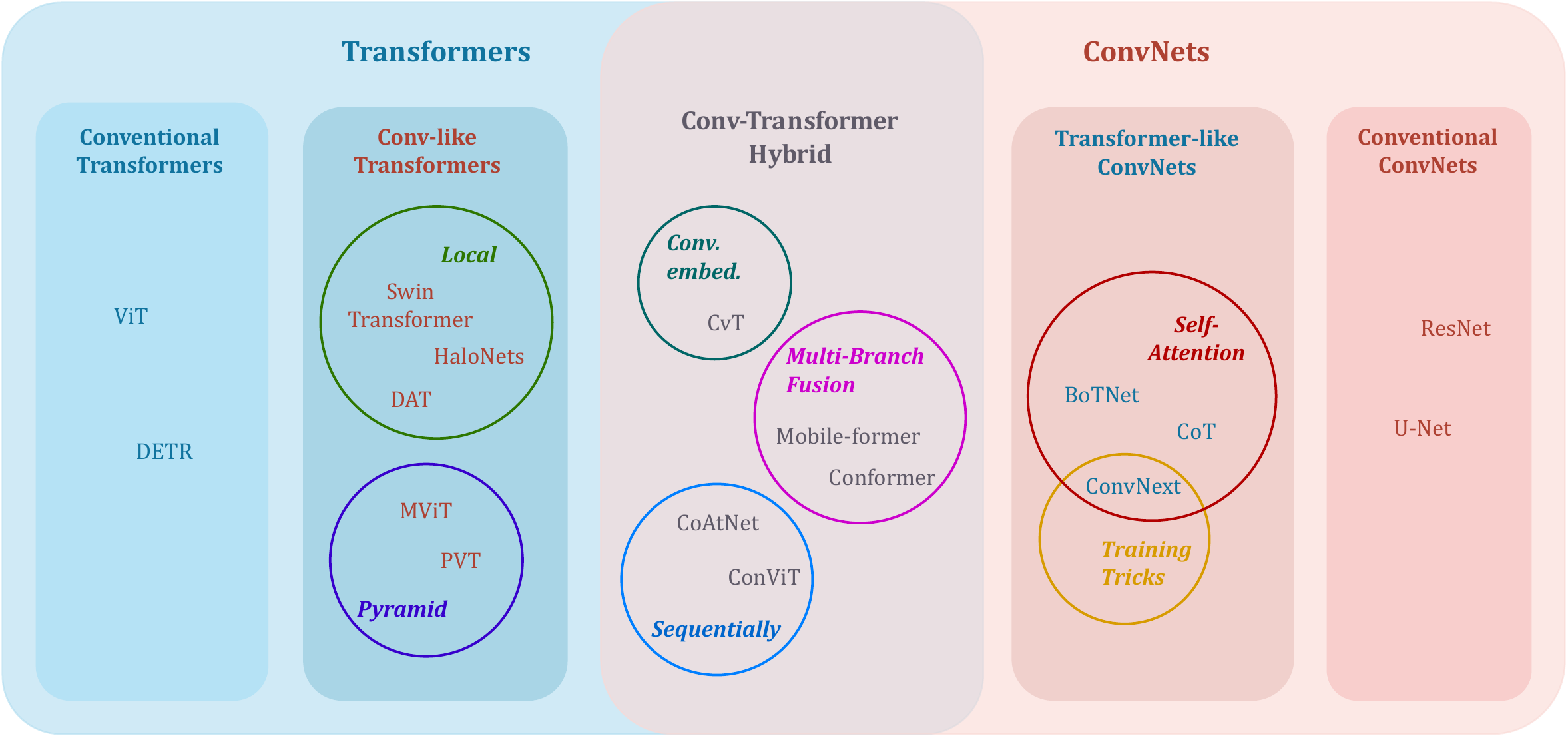}
\caption{Taxonomy of typical approaches in combining CNNs and Transformer.} \label{fig:vit-types}
\end{figure}

To embrace the benefits from \underline{conventional CNNs} (\textit{e.g.}, ResNet~\citep{he2016deep} and U-Net~\citep{ronneberger2015u}) and \underline{conventional Transformers} (\textit{e.g.}, the original ViT~\citep{dosovitskiy2020image} and DETR~\citep{carion2020end}), multiple works have been done in combining the strengths of CNNs and Transformer, which can be included into three types, and we illustrate them one by one in the following paragraphs. Additionally, Fig.~\ref{fig:vit-types} contains a taxonomy of typical methods that combine CNN and Transformer.

\underline{Conv-like Transformers:} This type of model introduces some convolutional properties into conventional Vision Transformer. The building blocks are still MLPs and MSAs, while arranged in a convolutional style. For example, in Swin Transformer~\citep{liu2021swin}, HaloNets~\citep{vaswani2021scaling}, and DAT~\citep{xia2022vision}, the self-attention is performed within a local window hierarchically and neighboring windows are merged in subsequent layers. Hierarchical multi-scale framework in MViT ~\citep{fan2021multiscale} and pyramid structures in PVT~\citep{wang2021pyramid} guide a Transformer to increase the capacity of intermediate layers progressively.
%  (MViT, swin, halonet, DAT, PVT), performs the self-attention within a local window hierarchically and merging neighboring windows in subsequent layers. boost the performance, while reducing computational complexity and memory requirement.

\underline{Transformer-like CNNs:} This type of model introduces the traits of Vision Transformers into CNNs. The building blocks are convolutions, while arranged in a more Vision Transformer way. Thus, this type of models are excluded in the introduction about Transformer models in Section~\ref{sec.progress}. Specifically, the self-attention mechanism is assembled to convolutions, like in CoT~\citep{li2021contextual} and BoTNet~\citep{srinivas2021bottleneck}, making a full exploration of neighboring context that compensates the CNNs' weakness in capturing long-range dependencies. ConvNext~\citep{liu2022convnet} \textit{modernizes} a ResNet by exploiting a depth-wise convolution as a substitute of self-attention, and following the training tricks from Swin Transformer~\citep{liu2021swin}.

\underline{Conv-Transformer hybrid:} A straightforward way of combining CNNs and Transformers is to employ them both in an attempt of leveraging both of their strengths. So the building blocks are convolutions, MLPs and MSAs. This is done by keeping self-attention modules to catch long-distance relationships, while utilizing the convolution to project patch embeddings in CvT~\citep{wu2021cvt}. Another type of methods is the multi-branch fusion, like Conformer~\citep{peng2021conformer} and Mobile-former~\citep{chen2021mobile}, which typically fuses the feature maps from two parallel branches, one from CNN and the other from Transformer, such that the information provided by both architectures is retained throughout the decoder. Analogously, convolutions and Transformer blocks are arranged sequentially in ConViT~\citep{d2021convit} and CoAtNet~\citep{dai2021coatnet}, and representations from convolutions are aggregated by MSAs in a global view.
%Another representative example of the hybrid models is the \underline{multi-branch fusion}, which typically fuses the feature maps from two parallel branches, one from CNN and the other from Transformer, such that the information provided by both architectures is retained throughout the decoder [\textit{some references here}].
\begin{figure}
\centering
\includegraphics[width=0.98\columnwidth]{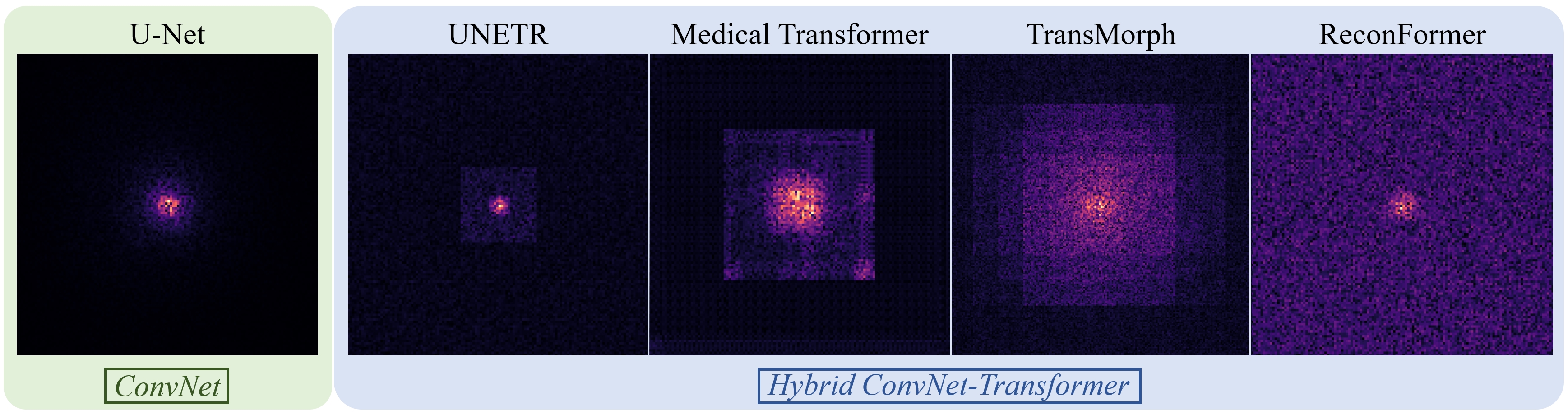}
\caption{Effective receptive fields (ERFs) \citep{luo2016understanding} of the well-known CNN, U-Net \citep{ronneberger2015u}, versus the hybrid Transformer-CNN models, including UNETR \citep{hatamizadeh2019deep}, Medical Transformer \citep{valanarasu2021medical}, TransMorph \citep{chen2021transmorph}, and ReconFormer \citep{guo2022reconformer}. The ERFs are computed at the last layer of the model prior to the output. The $\gamma$ correction of $\gamma=0.4$ was applied to the ERFs for better visualization. Despite the fact that its theoretical receptive field encompasses the whole image, the pure CNN model, U-Net \citep{ronneberger2015u}, has a limited ERF, with gradient magnitude rapidly decreasing away from the center.  On the other hand, all Transformer-based models have large ERFs that span over the entire image.} \label{fig:ERFs}
\end{figure}
\subsubsection{The role of MSA}
Arguably, the success of a Vision Transformer is brought by MSA. However, recent works show that 
%Recently there are some works put forward discrepancies in why Vision Transformers are successful. The main moot point is that 
the role of self-attention block is not that much irreplaceable in extracting global features. The MSA works as a trainable aggregation of feature maps~\citep{park2022vision}, whose function can be covered by MLPs repeatedly applied across spatial or channels in several MLP-mixer like models~\citep{tolstikhin2021mlp,touvron2021resmlp,yu2021metaformer}, or large kernel depth-wise convolutions~\citep{liu2022convnet,trockman2022patches,han2021connection}, or plain pooling operators to conduct spatial smoothing~\citep{yu2021metaformer}. 
In~\citep{tolstikhin2021mlp,yu2021metaformer,liu2022convnet}, researchers raise skeptical arguments, ascribing the performance gains to the design of pipeline, not MSAs. %If consider all these attempts comprehensively, we suggest that its not a precise statement. 
A perspective of Transformer and CNNs is that convolutions in CNNs and MLPs in Transformers both learn the patterns derived from images, and pooling in CNNs and all operations aforementioned in Transformers are aimed at fusing and integrating feature maps from previous layers. The differences lie in (i) \textit{fusion trainability}, when comparing MSAs with pooling, (ii) \textit{fusion field size}, when comparing original MSAs in ViT with those in Swin Transformer, and (iii) \textit{fusion method}, when comparing depth-wise convolutions with MLPs.

\subsection{Key properties}
\label{key_properties}

\begin{figure}
\centering
\includegraphics[width=0.98\columnwidth]{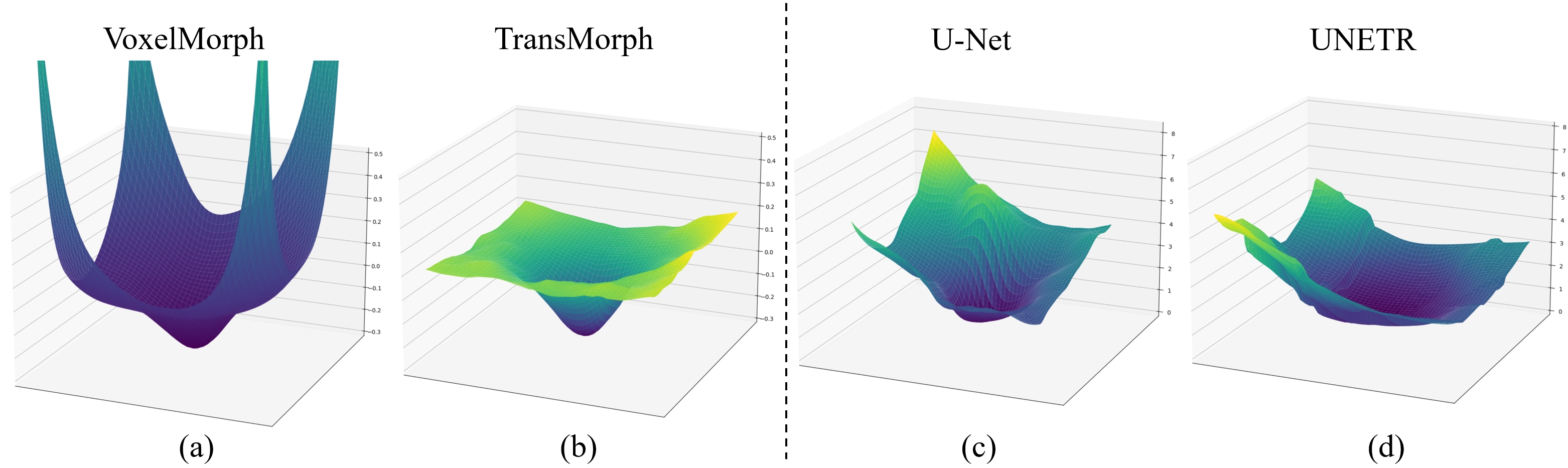}
\caption{Loss landscapes for the models based on CNNs versus Transformers. The left and right panels depict, respectively, the loss landscapes for registration and segmentation models. The left panel shows loss landscapes generated based on normalized cross-correlation loss and a diffusion regularizer; the right panel shows loss landscapes created based on a combination of Dice and cross-entropy losses. Transformer-based models, such as (b) TransMorph \citep{chen2021transmorph} and (d)  UNETR \citep{hatamizadeh2022unetr}, exhibit flatter loss landscapes than CNN-based models, such as (a) VoxelMorph \citep{balakrishnan2019voxelmorph} and (c) U-Net \citep{ronneberger2015u}.} \label{fig:loss-landscape}
\end{figure}

From the basic theory and architecture design of Transformer, researchers are yet to figure out why Transformer works better than say CNN in many scenarios. Below are some key properties associated with Transformers from the perspectives of modeling and computation.

\subsubsection{Modeling}

\begin{itemize}
\item[\hypertarget{M1}{$M_1$}:] \underline{Long-range dependency}. The MSA module connects all patches with a constant distance, and it is proved in \citep{joshi2020Transformers} that a Transformer model is equivalent to a graph neural network (GNN). It promises Transformer with large theoretical and effective receptive fields (as shown in~Fig. \ref{fig:ERFs}), and possibly brings better understanding of contextual information and long-range dependency than CNNs. %This properties lead a good capturing of long-range dependency in vision tasks. %(self-attentive, adaptive convolution; low-pass filtering)

\item[\hypertarget{M2}{$M_2$}:] \underline{Detail modeling}. Images are projected into embeddings by MLPs in Transformers. The embeddings of local patches are refined and adjusted progressively at the same scale. Features in CNNs, like ResNet and U-Net, are resized by pooling and strided-convolution operations. Features are at different detailing stages over scales. Dense modeling and trainable aggregation of features in Transformers can preserve contextual details along with more semantic information injected when deeper layers are reached~\citep{li2022exploring}.
%In uses dense modeling of local patches to preserve more context details while the pooling and strided-convolution operations somehow loses details. Global modeling ability helps preserve details to some extent.

\item[\hypertarget{M3}{$M_3$}:] \underline{Inductive bias}. {The convolutions in CNNs exploit the relations from the locality of pixels and apply the same weights across the entire image. This inherent inductive bias leads to faster convergence of CNNs and better performances in small datasets~\citep{d2021convit}. On the other hand, because computing self-attention is a global operation, Transformers in general have a weaker inductive bias than CNNs~\citep{cordonnier2019relationship}. The only manually injected inductive bias in original ViT~\citep{dosovitskiy2020image} is the positional embedding. Therefore, Transformers lack the inherent properties of locality and scale-invariance, making them more data-demanding and harder to train \citep{dosovitskiy2020image, touvron2021training}. However, the reduced inductive bias may improve the performance of Transformers when trained on a larger-scale dataset. See \ref{sec:ind_bias} for further details.}

\item[\hypertarget{M4}{$M_4$}:] \underline{Loss landscape}. The self-attention operation of Transformer tends to promote a flatter loss landscape~\citep{park2022vision}, even for hybrid CNN-Transformer models, as shown in Fig. \ref{fig:loss-landscape}. {This results in improved performance and better generalizability compared to CNNs when trained under the same conditions. See \ref{sec:loss_land} for further details.}

\item[\hypertarget{M5}{$M_5$}:] \underline{Noise robustness}. Transformers are more robust to common corruptions and perturbations, such as blurring, motion, contrast variation, and noise~\citep{bhojanapalli2021understanding, xie2021segformer}.

%\item[$M_6$:] \underline{Graph equivalence}. It is shown \citep{joshi2020Transformers} that a Transformer model is equivalent to a graph, so various concepts could be represented as vertexes, which is suitable for multi modalities fusion in medical tasks.

\end{itemize}

\subsubsection{Computation}

\begin{itemize}
\item[\hypertarget{C1}{$C_1$}:] \underline{Scaling behavior}. Transformers show the same scaling properties in NLP and CV ~\citep{Zhai2022Scaling}. The Transformer models achieve higher performance when their computation, model capacity, and data size scale up together.
%\textcolor{red}{[yet to be specified]}

\item[\hypertarget{C2}{$C_2$}:] \underline{Easy integration}. It is easy to integrate Transformers and CNNs into one computational model. As shown in Section~\ref{sec.theory}.C and future sections, there are multiple ways of integrating them, resulting in flexible architecture designs that are mainly grouped into Conv-like Transformers, Transformer-like CNNs, and Conv-Transformer hybrid. 

% \item[$C_3$:] \underline{Pipeline components}. The patch-encode-learn pipeline also works in MLP-based architectures. Pooling-based Transformers should also discussed. it is the patch-encode-learn pipeline achieves a success, and the learn part could be replaced by other archs (pooling+MLP or purely MLP). 

\item[\hypertarget{C3}{$C_3$}:] \underline{Computational intensiveness}. {While promising results may be obtained with Transformers, typical Transformers (e.g., ViT \citep{dosovitskiy2020image, zhai2021scaling}) require a significant amount of time and memory, particularly during training. See \ref{sec:comp_complex} for further details.}

%\item[$C_5$:] \underline{Positional encoding}. various methods and their effects. \textcolor{red}{[P2 is yet to be specified]}

\end{itemize}

% original MSA in ViT does a trainable fusion in global view, while Swin-T does that locally. MLPs and depth-wise convolutions 

% rethinking MSA:
% MSA not that good, can be replaced by MLP, pooling or Convs. (MLPMixer, ConvMixer, Metaformer, ConvNext)
% MSA is a trainable spatial smoothing. pooling and depthwise-conv also works in same way, an information selection process.

% pipeline brings the promotion.
% (patches are all you need)

\section{Current Progresses} \label{sec.progress}

\begin{figure}[tbp]
\centering
\subfigure[]{
    \includegraphics[width=0.32\linewidth]{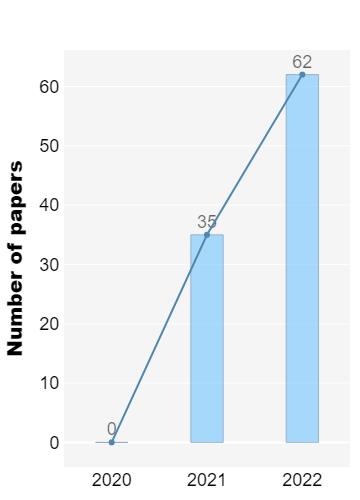}
    \label{fig:miccai_papers}
}
% 
% \includegraphics[width=0.4\textwidth]{figures/MICCAI_Transformer.jpg}
% \caption{The number of papers accepted to the MICCAI conference from 2020 to 2022 whose titles included the word "Transformer".} \label{fig:miccai_papers}
% \end{figure}
% 
% \begin{figure}[htbp]
\centering
% % \begin{minipage}[t]{0.48\textwidth}
\subfigure[]{
\includegraphics[width=0.62\linewidth]{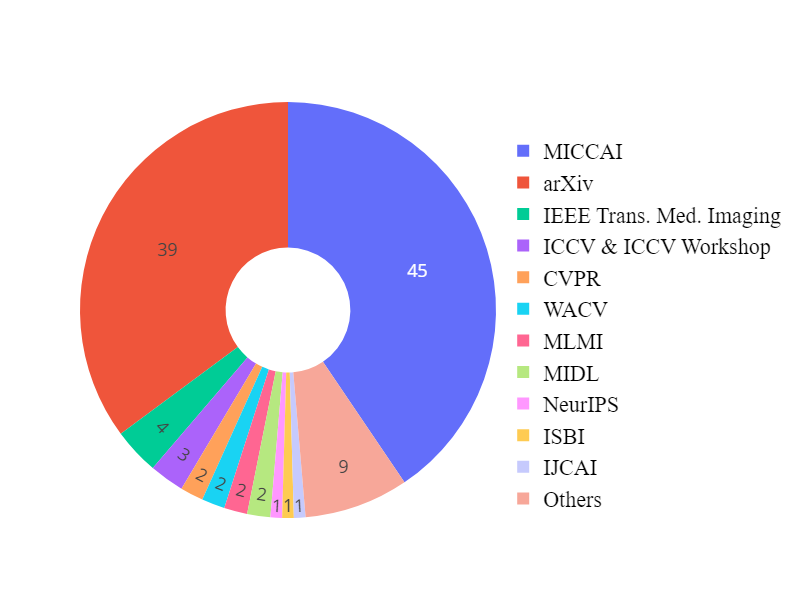}
\label{fig:paper_source}
}
% \includegraphics[width=0.46\textwidth]{figures/paper_source_n.png}
% \caption{Sources of all 114 selected papers.} \label{fig:paper_source}
% \end{minipage}
\caption{(a) The number of papers accepted to the MICCAI conference from 2020 to 2022 whose titles included the word "Transformer". (b) Sources of all 114 selected papers.}
\end{figure}

\begin{figure*}
\centering
\includegraphics[width=0.98\textwidth]{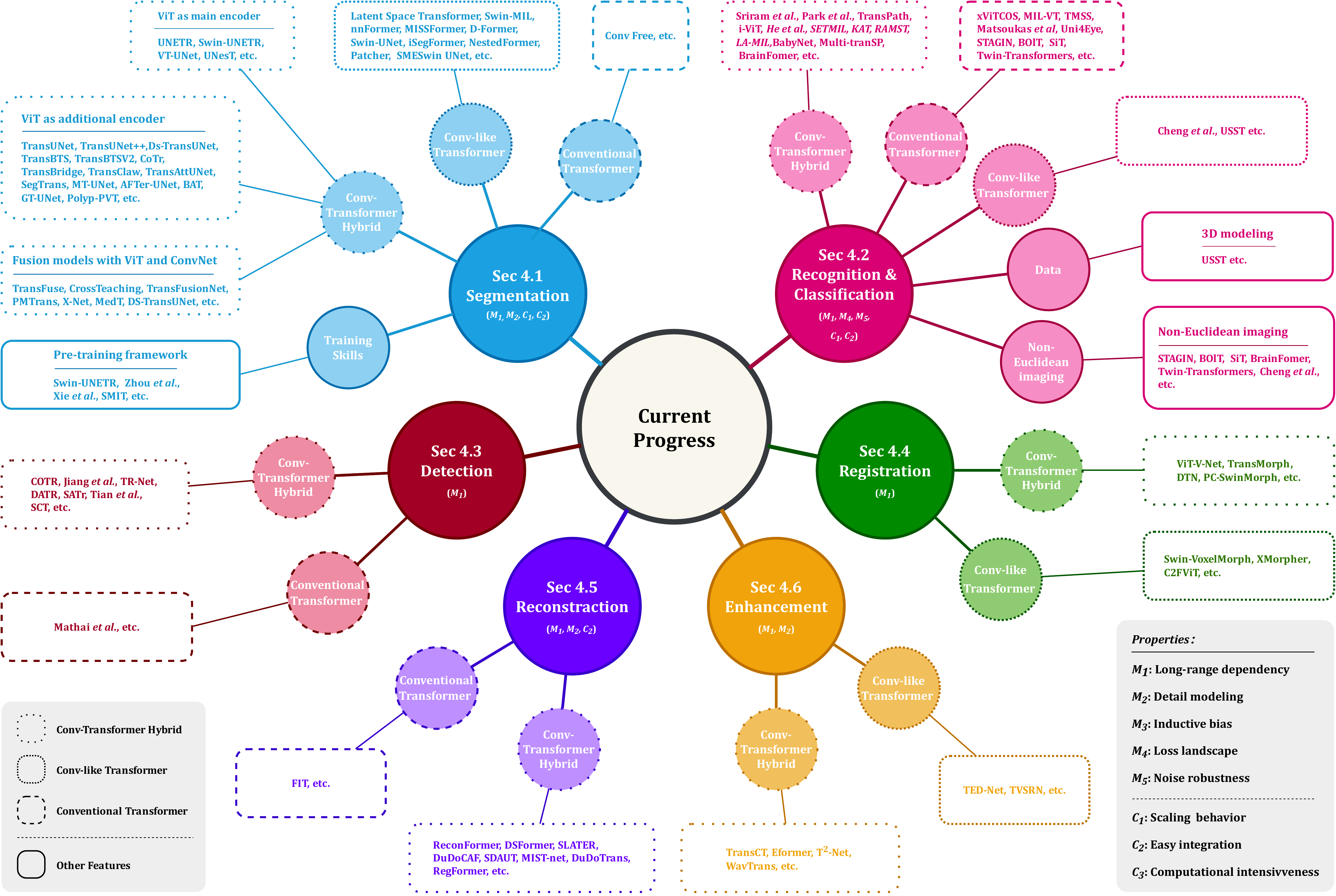}
\caption{An overview of Transformers applied in medical tasks in segmentation, recognition \& classification, detection, registration, reconstruction, and enhancement.} \label{fig:current_progress}
\end{figure*}

{As shown in Fig.~\ref{fig:miccai_papers}, Vision Transformers has received intensive study in present. We introduce the criteria of inclusion/exclusion for selecting research papers in this review. Fig.~\ref{fig:paper_source} shows the graphic summary of Transformers in medical image analysis papers. In particular, we investigate articles on IEEE, PubMed, Xplore, Springer, Science direct, proceedings of conferences including medical imaging conferences such as MICCAI, IPMI, ISBI, RSNA, SPIE, etc. Finally, we search manuscripts and project references on google scholar. In the result of search queries, we have found over 2000 transformer related papers, most of these contributions are from language studies or natural image analysis. We build our survey concepts from the self-attention paper, and vision transformer, which are keys milestones for exploring transformer in medical studies.  Finally, we set the criteria of legitimacy for this survey only about medical application with transformers. As shown in Fig.~\ref{fig:paper_source}, we demonstrate categorization of our selected papers based on tasks in medical domain. In the figure, we show percentage of article sources from conferences, journals, and pre-print platforms. The list of our selected papers, covering a wide range of topics including medical image segmentation, recognition \& classification, detection, registration, reconstruction, and enhancement, is by no means exhaustive. %For each reference, we also provide the task with its concerned image modality and anatomy and offer some remarks when appropriate.
Fig.~\ref{fig:current_progress} gives an overview of the current applications of vision Transformers, and below we present a literature summary for each topic with the use of key properties indicated accordingly.}

%It is worth mentioning that there are two prior papers~\cite{} that survey the literature of medical imaging with vision Transformer. Our paper is different from them in we supply a list of defining properties and its wide use of comparison as a means of explanation, both helping the readers to best understand the rationale behind the reviewed approaches.

\subsection{Medical image segmentation}  \label{sec.segmentation} 
%(2pages, 1 fig, 20 refs, Yucheng/Jun Li)
\begin{table*}[]
 \centering
\resizebox{\textwidth}{!}{
\begin{tabular}{lllrp{2cm}p{7cm}rp{7cm}}
\hline
Reference  &\multicolumn{1}{l}{Architecture} & \multicolumn{1}{l}{2D/3D}   & \multicolumn{1}{r}{\#Param}   & \multicolumn{1}{l}{{Modality}} & \multicolumn{1}{l}{Dataset}  &\multicolumn{1}{r}{ViT as Enc/Inter/Dec} & \multicolumn{1}{l}{Highlights}  \\ \hline
CoTr~\citep{xie2021cotr}   & Conv-Transformer Hybrid &3D   &46.51M  &CT & Multi-organ (BTCV~\citep{landman2015miccai}) & No/Yes/No & The Transformer block with deformable module captures deep features in the bottleneck. \\
SpecTr~\citep{yun2021spectr}  & Conv-Transformer Hybrid  &3D &N  &Microscopy & Cholangiocarcinoma~\citep{zhang2019multidimensional} & No/Yes/No &Hybrid Conv-Transformer encoder with spectral normalization.\\
TransBTS~\citep{wang2021transbts}   &Conv-Transformer Hybrid  &3D  &32.99M  &MRI &Brain Tumor~\citep{baid2021rsna} & No/Yes/No & 3D Transformer blocks for encoding bottleneck features.\\
UNETR~\citep{hatamizadeh2021unetr}  &Conv-Transformer Hybrid  &3D &92.58M  &CT, MRI &Multi-organ (BTCV~\citep{landman2015miccai}), Brain Tumor, Spleen (MSD~\citep{simpson2019large})  & Yes/No/No  & The 3D Transformer directly encodes image into features, and use of CNN decoder for capturing global information.\\
BiTr-UNet~\citep{jia2021bitr}   &Conv-Transformer Hybrid &3D  &N  &MRI &Brain Tumor~\citep{baid2021rsna} & No/Yes/No  & The bi-level Transformer blocks are used for encoding two level bottleneck features of acquired CNN feature maps. \\
VT-UNet~\citep{peiris2021volumetric}   &Conv-Transformer Hybrid &3D  &20.8M  &MRI, CT  &Brain tumor, Pancreas, Liver (MSD~\citep{simpson2019large}) & Yes/Yes/Yes &The encoder directly embeds 3D volumes jointly capture local/global information, the decoder introduces parallel cross-attention expansive path.\\
Swin UNETR~\citep{tang2022self, hatamizadeh2022swin}   &Conv-Transformer Hybrid &3D &61.98M  &CT, MRI &Multi-organ (BTCV)~\citep{landman2015miccai}, MSD 10 tasks~\citep{simpson2019large}   & Yes/No/No & The 3D encoder with swin-Transformer direclty encodes the 3D CT/MRI volumes with a CNN-based decoder for better capturing global information.\\
HybridCTrm~\citep{sun2021hybridctrm}  &Conv-Transformer Hybrid &3D   &N  &MRI &MRBrainS~\citep{mendrik2015mrbrains}, iSEG-2017~\citep{wang2019benchmark}   & Hybrid/N.A./No & A hybrid architecture encodes images from CNN and Transformer in parallel.\\
UNesT~\citep{yu2022characterizing}  &Conv-Transformer Hybrid  &3D  &87.30M  &CT &Kidney Sub-components (RenalSeg, KiTS~\citep{heller2021state})  & Yes/No/No &The use of hierarchical Transformer models for efficiently capturing multi-scale features with a 3D block aggregation module. \\
Universal~\citep{jun2021medical}  &Conv-Transformer Hybrid  &3D   &N  &MRI &Brain Tumor~\citep{baid2021rsna}  & No/Yes/No & The proposed model takes advantages of three views of 3D images and fuse 2D features to 3D volumetric segmentation. \\
% G, Li \textit{et al.}~\citep{li2022transbtsv2}  & Conv-Transformer Hybrid   &3D  &15.30M  &MRI, CT &Brain tumor~\citep{baid2021rsna}, Liver/kidney tumor (LiTS~\citep{bilic2019liver}, KiTS~\citep{heller2021state})  & No/Yes/No &The deformable bottleneck module is used in the Transformer blocks modeling bottleneck features to capture more shape-aware representations.  \\
PC-SwinMorph~\citep{liu2022pc}  &Conv-Transformer Hybrid &3D    &N  &MRI & Brain (CANDI~\citep{kennedy2012candishare}, LPBA-40~\citep{shattuck2008construction}) & No/No/Hybird & The designed patch-based contrastive and stitching strategy enforce a better fine detailed alignment and richer feature representation. \\ 
TransBTSV2~\citep{li2022transbtsv2}  & Conv-Transformer Hybrid   &3D  &15.30M  &MRI, CT &Brain Tumor~\citep{baid2021rsna}, Liver/Kidney Tumor (LiTS~\citep{bilic2019liver}, KiTS~\citep{heller2021state})  & No/Yes/No &The deformable bottleneck module is used in the Transformer blocks modeling bottleneck features to capture more shape-aware representations.  \\
GDAN~\citep{lin2022geometry}  & Conv-Transformer Hybrid   &3D  &N/A  &CT &Aorta  & No/Yes/No &Geometry-constrained module and deformable self-attention module are designed to guide segmentation.  \\
VT-UNet~\citep{peiris2022robust}  & Conv-Transformer Hybrid   &3D  &N/A  &MRI,CT &Brain Tumor~\citep{baid2021rsna}  & Yes/Yes/No &The self-attention mechanism to simultaneously encode local and global cues, the decoder employs a parallel self and cross attention formulation
to capture fine details for boundary refinement.  \\
ConTrans~\citep{lin2022contrans}  & Conv-Transformer Hybrid   &2D  &N/A  &Endoscopy, Microscopy, RGB, CT &Cell (Pannuke), (Polyp, CVC-ClinicDB~\citep{bernal2015wm}, CVC-ColonDB~\citep{bernal2012towards}, ETIS-Larib~\citep{silva2014toward}, Kvasir~\citep{jha2020kvasir}), Skin (ISIC~\citep{codella2018skin})  & Yes/Yes/No &Spatial-Reduction-Cross-Attention (SRCA) module is embedded in the decoder to form a comprehensive fusion of these two distinct feature representations and eliminate the semantic divergence between them.  \\
DA-Net~\citep{wang2022net}  & Conv-Transformer Hybrid   &2D  &N/A  &MRA images &Retina Vessels (DRIVE~\citep{staal2004ridge} and CHASE-DB1~\citep{fraz2012ensemble}) & No/Yes/No &Dual
Branch Transformer Module (DBTM) that can simultaneously and fully enjoy
the patches-level local information and the image-level global context.  \\
EPT-Net~\citep{liu2022edge}  & Conv-Transformer Hybrid   &3D  &N/A  &Intracranial Aneurysm &Intracranial Aneurysm (IntrA~\citep{yang2020intra}) & No/Yes/No &Dual stream transformer (DST), outeredge context dissimilation (OCD) and inner-edge hard-sample excavation (IHE) help the semantics stream produce sharper boundaries.   \\
Latent Space Transformer~\citep{li2022parameter}   & Conv-like Transformer  &3D &N/A  &CT, MRI & LiTS~\citep{bilic2019liver}, CHAOS~\citep{kavur2019}  & Yes/Yes/Yes & It intentionally make the large patches overlap to enhance intra-patch communication. \\
Swin-MIL~\citep{Qian2022TransformerBM}  & Conv-like Transformer  & 2D & N/A  & Microscopy & Haematoxylin and Eosin (H\&E)  & Yes/Yes/No &  A novel weakly supervised method for pixel-level segmentation in histopathology images, which introduces Transformer into the MIL framework to capture global or long-range dependencies. \\
\cline{1-8}
\\
Segtran~\citep{li2021medical} &Conv-Transformer Hybrid   &2D/3D  &166.7M  &Fundus, Colonoscopy, MRI &Disc/Cup (REFUGE20~\citep{orlando2020refuge}), Polyp, Brain Tumor  & No/Yes/N.A. &The use of squeeze and expansion block for contextualized features after acquiring visual and positional features of CNN. \\
MT-UNet~\citep{wang2021mixed} &Conv-Transformer Hybrid &2D    &N  &CT, MRI & Multi-organ (BTCV~\citep{landman2015miccai}), Cardiac (ACDC~\citep{bernard2018deep})  & No/Yes/No & The proposed mixed Transformer module simultaneously learns inter- and intra- affinities used for modeling bottleneck features.\\
TransUNet++~\citep{wang2022multiscale}  &Conv-Transformer Hybrid &2D    &N  &CT, MRI &Prostate, Liver tumor (LiTS~\citep{bilic2019liver}) & No/Yes/No &The feature fusion scheme at decoder enhances local interaction and context. \\
RT-Net~\citep{huang2022rtnet} &Conv-Transformer Hybrid   &2D   &N  &Fundus &Retinal (IDRiD~\citep{porwal2018indian}, DDR) & No/yes/No &The dual-branch architecture with global Transformer block and relation Transformer block enables detection of small size or blurred border.  \\
TransUNet~\citep{chen2021transunet}  &Conv-Transformer Hybrid &2D  &105.28M  &CT, MRI & Multi-organ (BTCV~\citep{landman2015miccai}), Cardiac (ACDC~\citep{bernard2018deep})  & No/Yes/No & Transformer blocks for encoding bottleneck features. \\
U-Transformer~\citep{petit2021u}  & Conv-Transformer Hybrid   &2D &N  &CT & Pancreas (TCIA)~\citep{holger2016turkbey}, Multi-organ  & No/Yes/No & The U-shape design with multi-head self-attention for bottleneck features and multi-head cross attention in the skip connections.\\
MBT-Net~\citep{zhang2021multi} & Conv-Transformer Hybrid   &2D  &N  &Microscopy &Corneal Endothelium cell (TM-EM300, Alizarine~\citep{ruggeri2010system})  & No/Yes/No & The design of hybrid residual Transformer model captures multi-branch global features.  \\
MCTrans~\citep{ji2021multi}   & Conv-Transformer Hybrid  &2D &7.64M  &Microscopy, Colonoscopy, RGB &Cell (Pannuke), (Polyp, CVC-ClinicDB~\citep{bernal2015wm}, CVC-ColonDB~\citep{bernal2012towards}, ETIS-Larib~\citep{silva2014toward}, Kvasir~\citep{jha2020kvasir}), Skin (ISIC~\citep{codella2018skin})  & No/Yes/No & The Transformer blocks are used for encoding bottleneck features in a UNet-like model.\\
Decoder~\citep{li2021more}  & Conv-Transformer Hybrid  &2D  &N  &CT, MRI &Brain tumor (MSD~\citep{simpson2019large}), Multi-organ (BTCV~\citep{landman2015miccai})  & No/No/Yes & The first study of evaluate the effect of using Transformer for decoder in the medical image segmentation tasks.\\
UTNet~\citep{gao2021utnet} & Conv-Transformer Hybrid   &2D  &9.53M  &MRI &Cardiac~\citep{campello2021multi}  & Hybrid/Hybrid/No & The design of a hybrid architecture in the encoder with convolutional and Transformer layers. \\
TransClaw UNet~\citep{chang2021transclaw}  &Conv-Transformer Hybrid &2D   &N  &CT &Multi-organ (BTCV~\citep{landman2015miccai})  & No/Yes/No &The Transformer blocks are used as additional encoder for strengthening global connection of CNN encoded features. \\
TransAttUNet~\citep{chen2021transattunet} &Conv-Transformer Hybrid  &2D   &N  &RGB, X-ray, CT, Microscopy &Skin (ISIC~\citep{codella2018skin}), Lung (JSRT~\citep{shiraishi2000development}, Montgomery~\citep{jaeger2014two}, NIH~\citep{tang2019xlsor}), (Clean-CC-CCII~\citep{he2020benchmarking}), Nuclei (Bowl, GLaS~\citep{malik2020instance})  & No/Yes/No & The model contains a co-operation of Transformer self-attention and global spatial attention for modeling semantic information. \\
LeViT-UNet(384)~\citep{xu2021levit}  &Conv-Transformer Hybrid  &2D  &52.17M  &CT, MRI &Multi-organ (BTCV~\citep{landman2015miccai}), Cardiac (ACDC~\citep{bernard2018deep})  & No/Yes/No & The lightweight design of Transformer blocks as second encoder.\\
Polyp-PVT~\citep{dong2021polyp} & Conv-Transformer Hybrid  &2D   &N  &Endoscopy  &Polp (Kvasir~\citep{jha2020kvasir}, CVC-ClinicDB~\citep{bernal2015wm}, CVC-ColonDB~\citep{bernal2012towards}, Endoscene~\citep{vazquez2017benchmark}, ETIS~\citep{silva2014toward})  & Yes/No/No & The Transformer encoder directly learns the image patches representation.\\
COTRNet~\citep{shen2021automated} &Conv-Transformer Hybrid   &2D  &N  &CT &Kidney (KITS21~\citep{heller2021state})  & Hybrid/N.A./No & The U-shape model design has the hybrid of CNN and Transformers for both encoder and decoder.\\
TransBridge~\citep{deng2021transbridge}  &Conv-Transformer Hybrid   & 2D & 11.3M  & Echocardiograph & Cardiac (EchoNet-Dynamic)~\citep{ouyang2020video}  & No/Yes/No & The Transformer blocks are used for capturing bottleneck features for bridging CNN encoder and decoder. \\
GT UNet~\citep{li2021gt}  & Conv-Transformer Hybrid  &2D   &N  &Fundus &Retinal (DRIVE~\citep{staal2004ridge}) & Hybrid/N.A./No & The design of hybrid grouping and bottleneck structures greatly reduces computation load of Transformer. \\
BAT~\citep{wang2021boundary} & Conv-Transformer Hybrid   &2D  &N  &RGB & Skin (ISIC~\citep{codella2018skin}, PH2~\citep{mendoncca2013ph})  & No/Yes/No & The model proposes a boundary-wise attention gate in Transformer for capturing prior knowledge.\\
AFTer-UNet~\citep{yan2022after}   & Conv-Transformer Hybrid  & 2D & 41.5M  & CT & Multi-organ (BTCV~\citep{landman2015miccai}), Thorax (Thorax-85~\citep{chen2021deep}, SegTHOR~\citep{lambert2020segthor})  & No/Yes/No & The proposed axial fusion mechanism enables intra- and inter-slice communication and reduced complexity.\\ 
\hline
\hline
\end{tabular}%
}
% \caption{The summarized review of Transformer-based model for medical image segmentation. }
%\label{tab:seg01}
\end{table*}

\begin{table*}[]
 \centering
\resizebox{\textwidth}{!}{
\begin{tabular}{lllrp{2cm}p{7cm}rp{7cm}}
Conv Free~\citep{karimi2021convolution}   & Conventional Transformer  &3D &N  & CT, MRI & Brain cortical~\citep{bastiani2019automated} plate, Pancreas, Hippocampus (MSD~\citep{simpson2019large})   & Yes/Yes/N.A. & 3D Transformer blocks as encoder without convolution layers\\
nnFormer~\citep{zhou2021nnformer}  &Conv-like Transformer  &3D  &158.92M  &CT, MRI &Brain tumor~\citep{baid2021rsna}, Multi-organ (BTCV~\citep{landman2015miccai}), Cardiac (ACDC~\citep{bernard2018deep})  & Yes/Yes/Yes & The 3D model with pure Transformer as encoder and decoder. \\
MISSFormer~\citep{huang2021missformer}  &Conv-like Transformer  &2D  &N  &MRI, CT &Multi-organ (BTCV~\citep{landman2015miccai}), Cardiac (ACDC~\citep{bernard2018deep})  & Yes/Yes/Yes & The U-shape design with patch merging and expanding modules as encoder and decoder. \\
D-Former~\citep{wu2022d}   &Conv-like Transformer  &2D &44.26M  &CT, MRI & Multi-organ (BTCV~\citep{landman2015miccai}), Cardiac (ACDC~\citep{bernard2018deep})  & Yes/Yes/Yes & The 3D network contains local/global scope modules to increase the scopes of information interactions and reduces complexity. \\
Swin-UNet~\citep{cao2021swin}  & Conv-like Transformer   &2D &N  &CT, MRI & Multi-organ (BTCV~\citep{landman2015miccai}), Cardiac (ACDC~\citep{bernard2018deep})  & Yes/Yes/Yes & The pure Transformer U-shape segmentation model design enables the use for both encoder and decoder\\ 
iSegFormer~\citep{liu2022isegformer}   &Conv-like Transformer  &3D &N/A  &MRI & Knee (OAI-ZIB~\citep{ambellan2019automated})  & Yes/Yes/Yes & It contains a memory-efficient Transformer thatcombines a Swin Transformer with a lightweight multilayer perceptron (MLP).
decoder. \\
NestedFormer~\citep{xing2022nestedformer}   &Conv-like Transformer  &3D &N/A  &MRI & BraTS2020~\citep{baid2021rsna}, MeniSeg  & Yes/Yes/Yes & A novel Nested Modality-Aware Transformer (NestedFormer) to explicitly explore the intra-modality and inter-modality relationships of multi-modal MRIs for brain tumor segmentation. \\
Patcher~\citep{ou2022patcher}   &Conv-like Transformer  &3D &N/A  &MRI, Endoscopy & Stroke Lesion, Kvasir-SEG~\citep{jha2020kvasir}  & Yes/Yes/Yes & This design allows Patcher to benefit from both the coarse-to-fine feature extraction common in CNNs and the superior spatial relationship modeling of Transformers. \\
SMESwin UNet~\citep{wang2022smeswin}   &Conv-like Transformer  &2D &N/A  &Microscopy & GlaS~\citep{malik2020instance}  & Yes/Yes/Yes & Fuse multi-scale semantic features and attentions maps by designing a compound structure with CNN and ViTs (named MCCT), based on Channel-wise Cross fusion Transformer (CCT) . \\
\hline
DS-TransUNet~\citep{lin2021ds}  & Conv-Transformer Hybrid  &2D  &N  &Colonoscopy, RGB, Microscopy &Polyp~\citep{jha2020kvasir}, Skin (ISIC~\citep{codella2018skin}), Gland (GLaS~\citep{malik2020instance})  & Yes/Yes/Yes & The use of swin Transformer as both encoder and decoder forms the U-shape design of segmentation model. \\
MedT~\citep{valanarasu2021medical}  &Conv-Transformer Hybrid  &2D  &N  &Ultrasound, Microscopy & Brain~\citep{valanarasu2020learning}, Gland~\citep{sirinukunwattana2017gland}, Multi-organ Nuclei (MoNuSeg~\citep{kumar2019multi})  & Yes/No/No & A fusion model with a global and local branches as encoders.\\
PMTrans~\citep{zhang2021pyramid}  &Conv-Transformer Hybrid &2D   &N  &Microscopy, CT &Gland (GLAS~\citep{malik2020instance}), Multi-organ Nuclei (MoNuSeg~\citep{kumar2019multi}), Head (HECKTOR~\citep{andrearczyk2020overview})  & Hybrid/No/No & The pyramid design of structure enables multi-scale Transformer layers for encoder image features.\\
TransFuse~\citep{zhang2021transfuse}  &Conv-Transformer Hybrid   &2D &26.3M  &Endoscopy, RGB, X-ray, MRI &Polp (Kvasir~\citep{jha2020kvasir}, ClinicDB~\citep{bernal2015wm}, ColonDB~\citep{bernal2012towards}, EndoScene~\citep{vazquez2017benchmark}, ETIS~\citep{silva2014toward}),  Skin (ISIC~\citep{codella2018skin}), Hippocampus, Prostate (MSD~\citep{simpson2019large})  & Hybrid/N.A./No & A CNN branch and a Transformer branch encoded features are fused by a BiFusion module to the decoder for segmentation.\\
CrossTeaching~\citep{luo2021semi}   &Conv-Transformer Hybrid  &2D  &N  &MRI &ACDC~\citep{bernard2018deep}  & Hybrid/Hybrid/Hybrid & The two branch network employs advantage of UNet and Swin-UNet.  \\
TransFusionNet~\citep{meng2021exploiting}   &Conv-Transformer Hybrid  &2D  &N  &CT &Liver Tumor (LiTS~\citep{bilic2019liver}), Liver Vessels (LTBV)~\citep{huang2018robust}, Multi-organ (3Dircadb~\citep{soler20103d})  & Hybrid/N.A./No & The Transformer- and CNN-based encoders extract both features directly from input and fuse to the CNN decoder. \\
% Mixed, Wang et al.~\citep{wang2021mixed}     & Conv-Transformer-Conv  & Mixed Transformer U-Net For Medical Image Segmentation\\
X-Net~\citep{li2021x}  &Conv-Transformer Hybrid &2D   &N  &Microscopy, Endoscopy &Nuclei (Bowl~\citep{caicedo2019nucleus}, TNBC~\citep{naylor2018segmentation}), Polyp (Kvasir~\citep{jha2020kvasir})   & Hybrid/Hybrid/Hybrid & The use of CNN reconstruction model and Transformer segmentation model with mixed representations. \\ 
\hline
\\
T-AutoML~\citep{yang2021t}  &Net Architecture Search &3D   &16.96M  &CT &Liver, Lung tumor (MSD~\citep{simpson2019large}) & N.A.  &The first medical architecture search framework designed for Transformer-based models. \\ \\
~\citep{xie2021unified}  &Pre-training Framework   &2D/3D  &N  &CT, MRI, X-ray, Dermoscopy &JSRT~\citep{shiraishi2000development}, ChestXR~\citep{wang2017chestx}, BTCV~\citep{landman2015miccai}, RICORD~\citep{tsai2021rsna}, CHAOS~\citep{kavur2019}, ISIC~\citep{codella2018skin}  & No/yes/No & The unified pre-training Framework of 3D and 2D images for Transformer models \\
~\citep{zhou2022self}  &Pre-training Framework   &3D   &N  &CT, MRI, X-ray &Lung (ChestX-ray14~\citep{wang2017chestx}), Multi-organ (BTCV~\citep{landman2015miccai}), Brain Tumor (MSD~\citep{simpson2019large})  & Yes/No/No & The masked autoencoder scheme adapts the pre-training framework for medical images. \\
~\citep{tang2022self, hatamizadeh2022swin} &Pre-training Framework    &3D &N  &CT, MRI &Multi-organ (BTCV~\citep{landman2015miccai}), MSD 10 tasks~\citep{simpson2019large}  & Yes/No/No & Very large-scale medical image pre-training framework with Swin Transformers.\\
SMIT~\citep{jiang2022self} &Pre-training Framework    &3D &N  &CT, MRI &Covid19, Kidney Cancer, BTCV~\citep{landman2015miccai}  & Yes/No/No & Self-distillation learning with masked image modeling method to perform SSL for vision transformers (SMIT) is applied to 3D multi-organ segmentation from CT and MRI. It contains a dense pixel-wise regression within masked patches called masked image prediction\\

\hline
\end{tabular}%
}
\caption{The summarized review of Transformer-based model for medical image segmentation. "N" denotes not reported or not applicable on number of model parameters. "N.A." denotes for not applicable for intermediate blocks or decoder module. }
\label{tab:segmentation}
\end{table*}

\begin{figure*}
\centering
\includegraphics[width=0.98\textwidth]{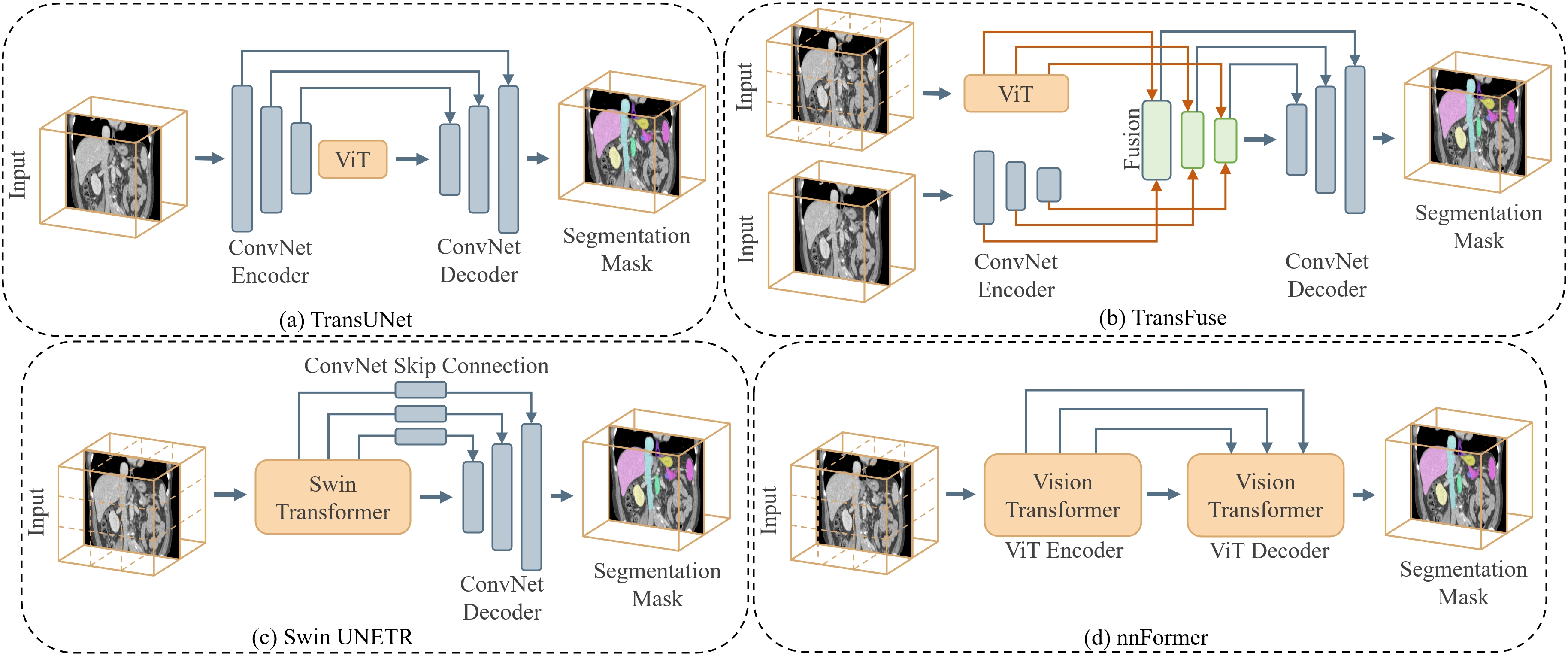}
\caption{Typical Transformer-based U-shaped segmentation model architectures. (a) The TransUNet~\citep{chen2021transunet}-like structure uses Transformer as additional encoder modeling bottleneck features. (b) The Swin UNETR~\citep{tang2022self} uses the Transformer as the main encoder and CNN decoder to construct the hybrid network. (c) The TransFuse~\citep{zhang2021transfuse} fuses CNN and Transformer encoders together to connect the decoder. (d) The nnFormer~\citep{zhou2021nnformer}-like structure uses a pure Transformer for both encoder and decoder.} \label{fig:seg_model_desgin}
\end{figure*}

In general, {Transformer-based models} outperform ConvNets for solving medical image segmentation tasks. The main reasons are as follows: 
\begin{itemize}
\item The ability of modeling longer range dependencies of context in high dimensional and high resolution medical images. [Property \hyperlink{M1}{$M_1$}]
\item The scalability and robustness of ViT and Swin Transformer strengthen the dense prediction for pixel-wise segmentation ~\citep{liu2021swin}. [Property \hyperlink{M2}{$M_2$}]
\item {The superior scaling behavior of Transformers over ConvNets and the lack of convolutional inductive bias in Transformers make them more advantageous to large-scale self-supervised pre-training on medical image datasets~\citep{tang2022self, Zhai2022Scaling}. [Property \hyperlink{C1}{$C_1$} and \hyperlink{M3}{$M_3$}]}
\item Network architecture design is flexible by mixing Transformer and CNN modules. [Property \hyperlink{C2}{$C_2$}]
\end{itemize}
Though it has demonstrated superior performance, the use of Transformers for medical image segmentation has challenges in transferring the representation capability from language domain to image modalities. Compared to word tokens that are modeled as the basic embedding, visual features are at variant scales. This multi-scale problem can be significant in dense prediction tasks with higher resolution of voxels in medical images. However, for the current Transformer backbones, the learnt embedding is commonly at a fixed scale, which is intractable for segmentation tasks, especial on large-scale medical radiography, microscopy, fundus, endoscopy or other imaging modalities. To adapt the vanilla Transformer models for medical image segmentation, recent researchers proposed solutions that utilize the components of ViT into particular segmentation models. In the following, we summarize and discuss recent works on how Transformer blocks are used in the segmentation models. Table~\ref{tab:segmentation} provides a summary list of all reviewed segmentation approaches along with their information about associated architecture type, model size, dataset, method highlight, etc. 
{As one of the most classical approaches in medical segmentation, U-Net~\citep{ronneberger2015u} is widely chosen for comparison by its followers. The U-shaped architecture and skip-connections in U-Net has proved its effectiveness in leveraging hierarchical features.}
%Although Transformer-based segmentation models have gained prevalence, the U-Net~\citep{ronneberger2015u} remains a prominent backbone architecture for them. 
Fig.~\ref{fig:seg_model_desgin} presents some typical Transformer-based U-shaped segmentation model architectures. 

\begin{figure*}
\centering
\includegraphics[width=0.9\textwidth]{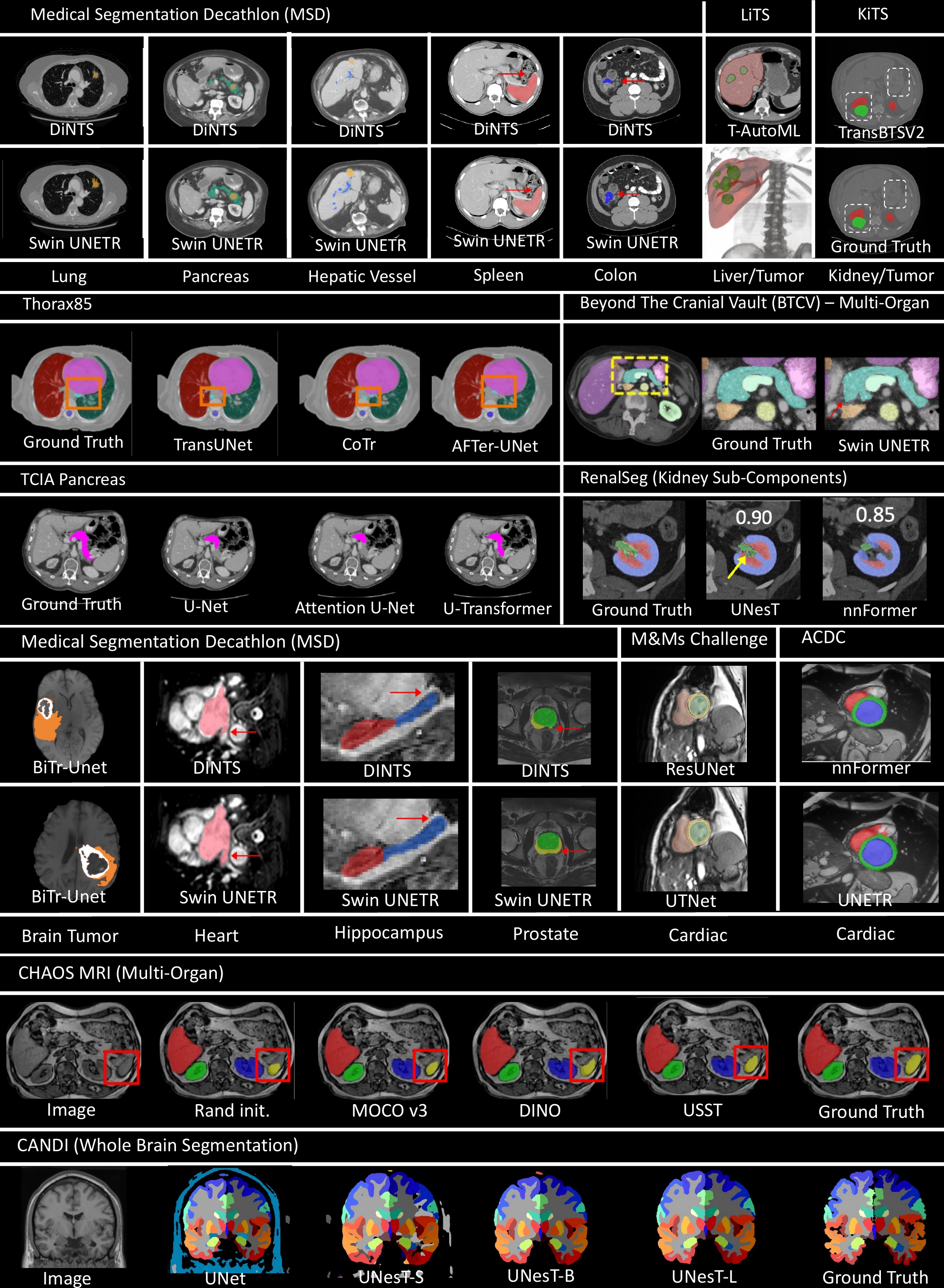}
\caption{Visualization of CT/MRI segmentation and comparison on public datasets between Transformer-based and baseline models. Transformer-based models includes Swin UNETR~\citep{tang2022self}, T-AutoML~\citep{yang2021t}, TransBTSV2~\citep{li2022transbtsv2}, AFTer-UNet~\citep{yan2022after}, U-Transformer~\citep{petit2021u}, UNesT~\citep{yu2022characterizing}, BiTr-Unet~\citep{jia2021bitr}, UTNet~\citep{gao2021utnet}, nnFormer~\citep{zhou2021nnformer}, MOCOv3~\citep{chen2021empirical}, and DINO~\citep{caron2021emerging}, USST~\citep{xie2021unified}. Baseline models contain the DiNTS~\citep{he2021dints}, ResUNet~\citep{zhang2018road}, and AttentionUNet~\citep{oktay2018attention}} \label{fig:quali_1}.
\end{figure*}

\begin{figure*}
\centering
\includegraphics[width=0.85\textwidth]{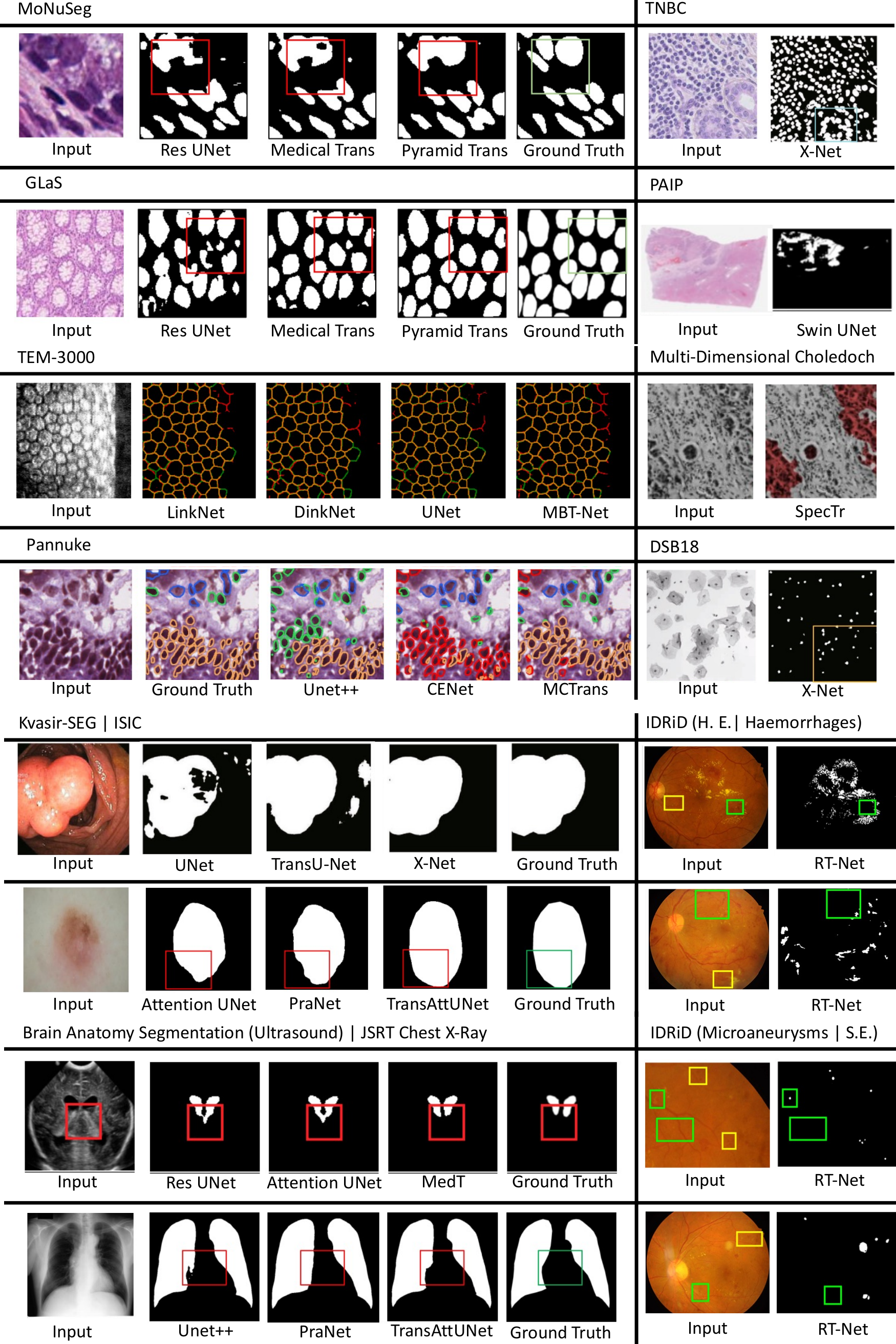}
\caption{Transformer segmentation to other medical image modalities such as endoscopy, microscopy, retinopathy, ultrasound, X-ray, and camera images. The comparison methods include Pyramid Trans~\citep{zhang2021pyramid}, MBT-Net~\citep{zhang2021multi}, MCTrans~\citep{ji2021multi}, X-Net~\citep{li2021x}, TransAttUnet~\citep{chen2021transattunet}, MedT~\citep{valanarasu2021medical}, Swin-UNet~\citep{nguyen2021evaluating}, SpecTr~\citep{yun2021spectr}, RT-Net~\citep{huang2022rtnet}, and ConvNet-based models (ResUNet~\citep{zhang2018road}, UNet~\citep{ronneberger2015u}, UNet++~\citep{zhou2018unet++}, and AttentionUNet~\citep{oktay2018attention}).} \label{fig:quali_2}
\end{figure*}

\underline{ViT as main encoder:} The Vision Transformers reformulate the segmentation problem as a 1D sequence-to-sequence inference task and to learn medical context from the embedded patches. A major advantage of the sequence-to-sequence modeling strategy is the larger receptive fields compared to CNNs~\citep{dosovitskiy2020image}, resulting in stronger representation capability with longer range dependencies. By employing these properties, models that directly use Transformer for generating the input sequences and tokenized patches are proposed~\citep{hatamizadeh2022unetr, tang2022self,peiris2021volumetric,yu2022characterizing}. ~\citep{hatamizadeh2022unetr} and ~\citep{peiris2021volumetric} introduce the volumetric model that utilizes the global attention-based Vision Transformer as the main encoder and then connects to the CNNs-based decoder or expand modules. ~\citep{tang2022self, hatamizadeh2022swin} demonstrate the use of shifted-window (Swin) Transformer, which presents more powerful representation ability, as the major encoder into the `U-shaped' segmentation architecture. The Swin UNETR model achieves state-of-the-art performance on the 10 tasks in Medical Segmentation Decathlon (MSD)~\citep{simpson2019large} and BTCV benchmarks. Similarly, ~\citep{yu2022characterizing} propose a hierarchical Transformer-based segmentation model that utilizes the 3D block aggregation, which achieves the state-of-the-art results on the kidney sub-components segmentation with CT images. 

\underline{ViT as additional encoder:} The second widely-adopted structures for medical image segmentation are to use the Transformer as the secondary encoder after ConvNets. The rationale of this design is the lack of inductive bias such as locality and translation equivariance of Transformers. In addition, the use of CNN as the main encoder can bring the computational benefit %of data and modeling global features, and 
as it is computationally expensive to calculate global self-attention among voxels in high-resolution medical images. 
One earlier adoption of $12$ layers ViT for the bottleneck features is the TransUNet~\citep{chen2021transunet}, which follows the 2D UNet~\citep{ronneberger2015u} design and incorporates the Transformer blocks in the middle structure. TransUNet++~\citep{wang2022multiscale} and Ds-TransUNet~\citep{lin2021ds} propose an improved version of the design that achieves promising results for CT segmentation tasks. For volumetric medical segmentation, TransBTS~\citep{wang2021transbts} and TransBTSV2~\citep{li2022transbtsv2} introduce the Transformer to model spatial patch embedding for the bottleneck feature. CoTr~\citep{xie2021cotr}, TransBridge~\citep{deng2021transbridge}, TransClaw~\citep{chang2021transclaw}, and TransAttUNet~\citep{chen2021transattunet} study the variant of attention blocks in the Transformer, such as the deformable mechanism that enables attention on a small set of key positions. SegTrans~\citep{li2021medical} exploits the squeeze and expansion block for modeling contextual features with Transformers for hidden representations. MT-UNet~\citep{wang2021mixed} uses a mixed structure for learning inter- and intra- affinities among features. More recently, several studies such as AFTer-UNet~\citep{yan2022after}, BAT~\citep{wang2021boundary}, GT-UNet~\citep{li2021gt}, and Polyp-PVT~\citep{dong2021polyp} focus on using grouping, boundary-aware or slice communication modules for improved robustness in ViT.

\underline{Fusion models with ViT and ConvNet:} While Transformers show the superiority of modeling long-range dependencies, its lack of capability of capturing {local feature} remains a challenge. Instead of cascading the Conv and Transformer blocks, researchers propose to leverage ViT and ConvNet as encoders that both take medical image as inputs. Afterwards, the embedded features are fused to connect to the decoder. The multi-branch design benefits from the advantages of learning global/local information for ViT and Convnet in parallel and then stacking representations in a sequential manner. TransFuse~\citep{zhang2021transfuse} uses a bi-fusion paradigm, in which the features from the two branches are fused to jointly make inference. CrossTeaching~\citep{luo2021semi} employs a semi-supervised learning with UNet and Swin Transformer for medical segmentation. TransFusionNet~\citep{meng2021exploiting} uses the CNN as the decoder to bridge the fused featured learnt from Transformer and ConvNet. PMTrans~\citep{zhang2021pyramid} introduces a pyramid structure for a multi-branch encoder with Transformers. X-Net~\citep{li2021x} demonstrates a dual encoding-decoding X-shape network structure for pathology images. MedT~\citep{valanarasu2021medical} designs model encoders with a CNN global branch and a local branch with gated axial self-attention. DS-TransUNet~\citep{lin2021ds} proposes to split the input image into
non-overlapping patches and then use two branches of encoder that learn feature representations at different scales; the final output is fused by Transformer Interactive Fusion (TIF) module.

\underline{Pure Transformer:} In addition to hybrid models, networks with pure Transformer blocks have been shown to be effective at modeling dense predictions such as segmentation. The nnFormer~\citep{zhou2021nnformer} proposes to use 3D Transformer that exploits the combination of interleaved convolutions and self-attention operations. The nnFormer also replaces the skip connection with a skip attention mechanism and it outperforms nnUNet significantly. MISSFormer~\citep{huang2021missformer} is a pure Transformer network with a feed-forward enhanced Transformer block with a context bridge. It models local features at different scales for leveraging long-range dependencies. D-Former~\citep{wu2022d} envisions an architecture with a D-Former block, which contains the dynamic position encoding block (DPE), local scope modules (LSMs), and the global scope modules (GSMs). The design employs a dilated mechanism that directly processes 3D medical images and improves the communication of information without increasing the tokens in self-attention.  Swin-UNet~\citep{cao2021swin} utilizes the advantages of shifted window self-attention Transformer blocks to construct a U-shaped segmentation network for 2D images. The pure Transformer architecture also uses the Transformer block as the expansion modules to upsample feature maps. However, current pure Transformer-based segmentation model are commonly of large model size, resulting in challenges of design robustness and scalability.

\underline{Pre-training framework for medical segmentation:} {Based on the empirical studies of Vision Transformer, the self-attention blocks commonly require pre-training data at a large scale to learn a more powerful backbone~\citep{dosovitskiy2020image}. Compared to CNNs, Transformer models are more data-demanding at different scales~\citep{Zhai2022Scaling}, effective and efficient ViT models are typically pre-trained by appropriate scales of dataset. However, adapting from natural images to a medical domain remain a challenge as the context gap is large. In addition, generating expert annotation of medical images is nontrivial, expensive and time-consuming; therefore it is difficult to collect large-scale annotated data in medical image analysis. Compared to the fully supervised dataset, raw medical images without expert annotation are easier to obtain. Hence, transfer learning, which aims to reuse the features of already trained ViT on different but related tasks, can be employed. To further improve the robustness and efficiency of ViT in medical image segmentation, several works are proposed to learn in a self-supervised manner a model of feature representations without manual labels.
Self-supervised Swin UNETR~\citep{tang2022self} collects a large-scale of CT images (5,000 subjects) for pre-training the Swin Transformer encoder, which derives significant improvement and state-of-the-art performance for BTCV~\citep{landman2015miccai} and Medical Segmentation Decathlon (MSD)~\citep{antonelli2021medical}. The pre-training framework employs multi-task self-supervised learning approaches including image inpainting, contrastive learning and rotation prediction.  Self-supervised masked autoencoder (MAE)~\citep{zhou2022self} investigates the MAE-based self pre-training paradigm designed for Transformers, which enforces the network to predict masked targets by collecting information from the context. Furthermore, the unified 2D/3D pre-training~\citep{xie2021unified} aims to construct a teacher-student framework to leverage unlabeled medical data. The approach designs a pyramid Transformer U-Net as the backbone, which takes either 2D or 3D patches as inputs depending on the embedding dimension.}

\underline{Segmentation Transformers for different imaging modalities:}
{Medical image modalities are of potential challenges with deep learning tools. The medical segmentation decathlon~\citep{antonelli2021medical}, a challenge dataset designed for general purpose segmentation tools, contains multiple radiological modalities including dynamic CTs, T1w, T2w, and FLAIR MRIs. In addition, pathology images, endoscopy intervention data, or videos are also challenging medical segmentation scenarios. Upon image modalities with Transformer model, for only CT studies, CoTr~\citep{xie2021cotr}, U-Transformer~\citep{petit2021u}, TransClaw~\citep{chang2021transclaw}, COTRNet~\citep{shen2021cotr}, AFTerNet~\citep{yan2022after}, TransFusionNet~\citep{meng2021exploiting}, T-AutoML~\citep{yang2021t}, etc. conduct experiments on extensive evaluation. Among a large number of methods, researchers attempt to explore general segmentation approaches that can at least handle volumetric data both in CT and MRI, for which UNETR~\citep{hatamizadeh2022unetr}, VT-UNet~\citep{peiris2021volumetric}, SwinUNETR~\citep{tang2022self}, UNesT~\citep{yu2022characterizing}, MT-UNet~\citep{wang2021mixed}, TransUNet~\citep{chen2021transunet},, TransClaw~\citep{chang2021transclaw}, LeViT-UNet~\citep{xu2021levit}, nnFormer~\citep{zhou2021nnformer}, MISSformer~\citep{huang2021missformer}, D-Former~\citep{wu2022d}, Swin-UNet~\citep{cao2021swin}, and some pre-training workflows are proposed. Regarding pathology images, SpecTr~\citep{yun2021spectr}, MBT-Net~\citep{zhang2021multi}, MCTrans~\citep{ji2021multi}, MedT~\citep{valanarasu2021medical}, and X-Net~\citep{li2021x} are some pioneering works. Finallt, SegTrans~\citep{li2021contextual}, MCTrans~\citep{ji2021multi}, Polyp-PVT~\citep{dong2021polyp}, DS-TransUNet~\citep{lin2021ds}, and TransFuse~\citep{zhang2021transfuse} can model endoscopy images or video frames.}

\subsection{Medical image recognition and classification} \label{sec.recog} 
%(0.5page, Junyu Chen / Ce Wang) 
\begin{table*}[!ht]
 \centering
\resizebox{\textwidth}{!}{
\begin{tabular}{p{6.5cm}llllp{4cm}p{2cm}p{7cm}p{7cm}}
\hline
Reference  &\multicolumn{1}{l}{Architecture} & \multicolumn{1}{l}{2D/3D}  & \multicolumn{1}{l}{Pre-training} & \multicolumn{1}{l}{\#Param} & \multicolumn{1}{l}{Classification Task}  & \multicolumn{1}{l}{Modality} & \multicolumn{1}{l}{Dataset}  & \multicolumn{1}{l}{Highlights}  \\
\hline
~\citep{sriram2021covid}  & Conv-Transformer Hybrid & 2D & Pre-trained CNN & N  & COVID-19 Prognosis & X-ray  & CheXpert~\citep{irvin2019chexpert}, NYU COVID~\citep{shamout2021artificial} & A pre-trained CNN backbone extracts features from individual image, and a Transformer is applied to the extracted features from a sequence of images for prognosis. \\
~\citep{park2021vision}  & Conv-Transformer Hybrid & 2D & Pre-trained CNN & N & COVID-19 Diagnosis & X-ray  & CheXpert~\citep{irvin2019chexpert} & A pre-trained CNN backbone is integrated with ViT for classification. \\
TransPath~\citep{wang2021transpath}  & Conv-Transformer Hybrid & 2D & Pre-trained CNN + ViT & N & Histopathological Image Classification & Microscopy & TCGA~\citep{tomczak2015cancer}, PAIP~\citep{kim2021paip}, NCT-CRC-HE~\citep{kather2019predicting}, PatchCamelyon~\citep{bejnordi2017diagnostic}, MHIST~\citep{wei2021petri} & The entire network is pre-trained prior to the downstream tasks. The TAE module is introduced to the ViT in order to aggregate token embeddings and subsequently excite the MSA output. \\
i-ViT~\citep{gao2021instance}  & Conv-Transformer Hybrid & 2D & No & N  & Histological Subtyping & Microscopy & AIPath~\citep{gao2021nuclei} & A lightweight CNN is used to extract features from a series of image patches, which is then followed by a ViT to capture high-level relationships between patches for classification.\\
~\citep{he2021global}  & Conv-Transformer Hybrid & 2D & No & N & Brain Age Estimation & MRI & Brain MRI (BGSP~\citep{holmes2015brain}, OASIS-3~\citep{lamontagne2019oasis}, NIH-PD~\citep{evans2006nih}, ABIDE-I~\citep{di2014autism}, IXI*, DLBS~\citep{park2012neural}, CMI~\citep{alexander2017open}, CoRR~\citep{zuo2014open}) & Two CNN backbones, one of which extract features from the whole image and the other from the image patches. Then, a Transformer is used to aggregate the
features from the two backbones for classification.\\
SETMIL~\citep{zhao2022setmil} & Conv-Transformer Hybrid & 2D & Pre-trained CNN &  N & Gene Mutation Prediction, Lymph Node Metastasis Diagnosis & Microscopy  & Whole Slide Pathological Image & A novel spatial encoding wiht Transformer is proposed for multiple instance learning. \\
KAT~\citep{zheng2022kernel} & Conv-Transformer Hybrid & 2D & Pre-trained CNN & N &  Tumor Grading \& Prognosis & Microscopy & Whole Slide Pathological Image & A cross-attention Transformer is proposed to enable information exchange across tokens based on their spatial relationship on the whole slide image. \\
RAMST~\citep{lv2022joint} & Conv-Transformer Hybrid & 2D & Pre-trained CNN & N &  Microsatellite Instability Classification & Microscopy  & Whole Slide Pathological Image & A combination region- and whole-slide-level Transformer is proposed. The Transformer accepts sampled patches per the attention map and combines two levels of information for the final classification. \\
LA-MIL~\citep{reisenbuchler2022local} & Conv-Transformer Hybrid & 2D & Pre-trained CNN & N & Microsatellite Instability Classification, Mutation Prediction & Microscopy &  Whole Slide Pathological Image (TCGA colorectal \& stomach~\citep{weinstein2013cancer}) & A local attention graph-based Transformer is proposed for multiple instance learning, as well as an adaptive loss function to mitigate the class imbalance problem.\\
BabyNet~\citep{plotka2022babynet} & Conv-Transformer Hybrid & 2D+$t$ & No & N & Birth Weight Prediction & Ultrasound & Fetal Ultrasound Video Scans & BabyNet advances a 3D ResNet with a Transformer module to improve the local and global feature aggregation.\\
Multi-transSP~\citep{zheng2022multi} & Conv-Transformer Hybrid & 2D & Pre-trained CNN & N &  Survival Prediction for Nasopharyngeal Carcinoma Patients & CT & In-house CT Scans & A hybrid CNN-Transformer model that combines CT image and tabular data (i.e., clinical text data) is developed for survival prediction of nasopharyngeal carcinoma patients.\\
BrainFormer~\citep{dai2022brainformer} & Conv-Transformer Hybrid & 3D & No & N & Autism, Alzheimer’s Disease, Depression, Attention Deficit
Hyperactivity Disorder, and Headache Disorders Classification & fMRI & ABIDE~\citep{di2014autism}, ADNI~\citep{petersen2010alzheimer}, MPILMBB~\citep{mendes2019functional}, ADHD-200~\citep{bellec2017neuro} and ECHO &  A 3D CNN and Transformer Hybrid network employs CNNs to model local cues and Transformer to capture global relation among distant brain regions.\\
\cline {1-9}\\
xViTCOS~\citep{mondal2021xvitcos}  & Conventional Transformer & 2D & Pre-trained ViT & N & COVID-19 Diagnosis & CT, X-ray & Chest CT (
COVIDx CT-2A~\citep{gunraj2021covid}), Chest X-ray (COVIDx-CXR-2~\citep{pavlova2021covid}, CheXpert~\citep{irvin2019chexpert}) & A multi-stage transfer learning strategy is proposed for fine-tuning pre-trained ViT on medical diagnostic tasks. \\
MIL-VT~\citep{yu2021mil} & Conventional Transformer & 2D & Pre-trained ViT & N & Fundus Image Classification & Fundus & APTOS2019$^\dag$, RFMiD2020~\citep{pachade2021retinal} & Multiple instance learning module is introduced to the pre-trained ViT that learns from both the classification tokens and the image patches. \\
~\citep{matsoukas2021time} & Conventional Transformer & 2D & Pre-trained ViT & N & Dermoscopic, Fundus, and Mammography Image Classification & Fundus, Dermoscopy, Mammography & ISIC2019$^\ddag$, APTOS2019$^\dag$, CBIS-DDSM~\citep{lee2017curated} & This study investigates the effectiveness of pre-training DeiT versus ResNet on medical diagnostic tasks. \\
TMSS~\citep{saeed2022tmss} & Conventional Transformer & 3D & No & N & Survival Prediction for Head and Neck Cancer Patients & PET/CT & HECKTOR~\citep{oreiller2022head} & A Transformer for end-to-end survial prediction and segmentation using PET/CT and electronic health records (i.e., clinical text data). \\
Uni4Eye~\citep{cai2022uni4eye} & Conventional Transformer & 2D/3D  & Pre-trained ViT & N & Ophthalmic Disease Classification & OCT, Fundus & OCTA-500~\citep{li2020image}, GAMMA~\citep{wu2022gamma}, GAMMA~\citep{wu2022gamma}, EyePACS$^\S$, Ichallenge-Ichallenge-PMAMD~\citep{milea2020artificial}, Ichallenge-PM~\citep{fu2018joint}, PRIME-FP20~\citep{Ding2021Weakly} & A self-supervised learning framework is developed to pre-train a Transformer using both 2D and 3D ophthalmic images for ophthalmic disease classification.\\
STAGIN~\citep{NEURIPS2021_22785dd2} & Conventional Transformer & 3D+$t$ & No & 1.2M & Gender, Cognitive Task Classification & fMRI & HCP S1200 ~\citep{van2013wu} & A conventional Transformer encoder is employed to capture the temporal attention over features of functional connectivity from fMRI. \\
BolT~\citep{bedel2022bolt} & Conventional Transformer & 3D+$t$ &  No & N & Gender Prediction, Cognitive Task and Autism Spectrum Disorder Classification & fMRI & HCP S1200~\citep{van2013wu}, ABIDE~\citep{di2014autism} & A cascaded Transformer encodes features of BOLD responses via progressively increased temporally-overlapped window attention. \\
SiT~\citep{dahan2022surface} & Conventional Transformer & 3D & Pretrained ViT & 21.6M & Cortical Surface Patching, Postmenstrual Age (PMA) and Gestational Age (GA) & MRI & dHCP \citep{https://doi.org/10.1002/mrm.26462} & Reformulating surface learning task as seq2seq problem and solving it by ViTs. \\
Twin-Transformers~\citep{yu2022disentangling} & Conventional Transformer & 3D+$t$ & No & N & Brain Networks Identification & fMRI & HCP S1200~\citep{van2013wu} & A Twin-Transformers is proposed to simultaneously capture temporal and spatial features from fMRI. \\
~\citep{cheng2022spherical} & Conv-like Transformer & 3D & No & 6.23M & Cortical Surfaces Quality Assessment & MRI & Infant Brain MRI Dataset & The first work extended Transformer into spherical space. \\
USST~\citep{xie2021unified} & Conv-like Transformer & 2D/3D & Pre-trained ViT & N & COVID-19 Diagnosis, Pneumonia Classification & X-ray, CT& RICORD~\citep{tsai2021rsna}, ChestXR~\citep{CovidGrandChallenge2021} & The unified pre-training framework that allows the pre-training using 3D and 2D images is introduced to Transformers. \\

\hline
*https://brain-development.org/ixi-dataset/\\
$^\dag$https://www.kaggle.com/c/aptos2019-blindness-detection/\\
$^\ddag$https://challenge.isic-archive.com/landing/2019/\\
$^\S$https://https://www.kaggle.com/c/diabetic-retinopathy-detection/\\
\end{tabular}%
}
\caption{The summarized review of Transformer-based model for medical image classification. “”"N.A." denotes for not applicable for intermediate blocks or decoder module. "N" denotes not reported or not applicable on the number of model parameters. "$t$" denotes temporal dimension.}
\label{tab:classification}
\end{table*}

Since the advent of ViT~\citep{dosovitskiy2020image}, it has exhibited exceptional performances in natural image classification and recognition~\citep{wang2021pyramid, liu2021swin, touvron2021training, chu2021twins}. The benefits of ViT over CNN to image classification tasks are likely due to the following properties: 
\begin{itemize}
\item The ability of a \textit{single} self-attention operation in ViT to globally characterize the contextual information in the image provided by its large theoretical and effective receptive field~\citep{ding2022scaling, raghu2021vision}. [Property \hyperlink{M1}{$M_1$}]
\item The self-attention operation tends to promote a more flat loss landscape, which results in improved performance and better generalizability~\citep{park2022vision}. [Property \hyperlink{M4}{$M_4$}]
\item ViT is shown to be more resilient than CNN to distortions (e.g., noise, blur, and motion artifacts), semantic changes, and out-of-distribution samples~\citep{cordonnier2019relationship, bhojanapalli2021understanding, xie2021segformer}. [Property \hyperlink{M5}{$M_5$}]
\item {ViT has a weaker inductive bias than CNN, whose convolutional inductive bias has been shown to be advantageous for learning from smaller datasets~\citep{dosovitskiy2020image}. However, with the help of pre-training using a significant large amount of data, ViT is able to surpass convolutional inductive bias by learning the relevant patterns directly from data. [Property \hyperlink{M3}{$M_3$}]}
\item Related to the previous property, the superior scaling behavior of ViT over CNN with the aid of a large model size and pre-training on large datasets~\citep{liu2022convnet, Zhai2022Scaling}. [Property \hyperlink{C1}{$C_1$}]
\item It is flexible to design different network architectures by mixing Transformer and CNN modules to accommodate different modeling requirements. [Property \hyperlink{C2}{$C_2$}]
\end{itemize}

These appealing properties have sparked an increasing interest in developing Transformer-based models for medical image classification and recognition. The original ViT~\citep{dosovitskiy2020image} achieves superior classification performance with the help of pre-training on large-scale datasets. Indeed, as a result of their weaker inductive bias, pure ViTs are more "data hungry" than CNNs~\citep{park2022vision, liu2021efficient, bao2021beit}. As a result of this discovery, many supervised and self-supervised pre-training schemes for Transformers have been proposed for applications like COVID-19 classification~\citep{park2021vision, xie2021unified, mondal2021xvitcos}, retinal disease classification~\citep{yu2021mil, matsoukas2021time}, and histopathological image classification~\citep{wang2021transpath}. Despite the intriguing potential of these models, obtaining large-scale pre-training datasets is not always practicable for some applications. Therefore, there have been efforts devoted to developing hybrid Transformer-CNN classification models that are less data-demanding~\citep{sriram2021covid, park2021vision, dai2021transmed, he2021global, gao2021instance}. Next we briefly review and analyze these recent works for medical image classification and also list the reviewed works in Table~\ref{tab:classification}.

%\subsubsection
\underline{Hybrid model:}
The earliest use of ViTs for medical image classification is on COVID-19 classification from chest X-rays~\citep{sriram2021covid, park2021vision}. Public datasets like CheXpert~\citep{irvin2019chexpert}, ChestXR~\citep{CovidGrandChallenge2021}, and COVIDx CXR~\citep{Wang2020} provide over 10,000 chest x-ray images. Due to the massive quantity of images in these datasets, they are suitable for network pre-training as well as for evaluating downstream classification tasks. \citep{sriram2021covid} introduce a hybrid CNN-Transformer model for COVID-19 prognosis by analyzing a series of chest X-ray images taken at various time points. Specifically, a MOCO~\citep{he2020momentum, chen2020improved} encoder (a CNN) pre-trained in a self-supervised manner is used to extract features from each X-ray image. The features extracted from multiple images of the same patient are then fed into a Transformer followed by a linear classifier for classification. In their model, only the CNN backbones (\textit{i.e.}, the MOCO encoders) are pre-trained and the Transformer is randomly initialized, whereas the overall network is fine-tuned for the classification task. Similarly, \citep{park2021vision} propose to bridge DenseNet-121~\citep{huang2017densely} with ViT. The DenseNet is pre-trained on the CheXpert dataset using the Probabilistic Class Activation Map (PCAM) pooling operations introduced in~\citep{ye2020weakly}, whilst the ViT is randomly initialized. The overall network is subsequently trained and evaluated on several chest X-ray datasets for COVID-19 diagnosis, where their model outperforms ResNet~\citep{he2016deep} and vanilla ViT~\citep{dosovitskiy2020image} that are trained using the same training strategy. {\citep{zhao2022setmil} propose SETMIL for pathological image analysis. SETMIL begins by embedding the large-sized whole slide image (WSI) in low-resolution position-encoded embeddings via a pre-trained CNN. Then, low-resolution embeddings are subjected to a Transformer-based pyramid multi-scale fusion based on tokens-to-token ViT~\citep{yuan2021tokens} to extract multi-scale context information. A novel spatial encoding Transformer that combines absolute and relative positional embedding is used for the final classification. To achieve a similar objective, \citep{zheng2022kernel} propose KAT, which focuses on establishing the correspondence between tokens and a set of kernels associated with a set of positional anchors on the WSI. A CNN that has been pre-trained is first used to extract features from the non-overlapping patches of the WSI. In the meanwhile, a set of anchor points is extracted using K-means clustering on the feature patches. Then, a set of multi-scale weighting masks for each anchor point is defined and sent together with the feature patches and a set of trainable kernels to a Transformer. The Transformer uses cross-attention between tokens and kernels, and classification is achieved through kernel interaction with the classification token. This reduces the quadratic computational cost of the Transformer and reaches close to linear complexity in relation to the size of the WSI. In \citep{lv2022joint}, Lv \textit{et al.} introduce RAMST for the classification of microsatellite instability. In particular, a feature weight uniform sampling method is presented to learn representative features of image regions, and a Transformer encoder is used to aggregate region-level features with patch-level features extracted by a pre-trained CNN. Meanwhile, Reisenb{\"u}chler \textit{et al.} propose a local attention graph-based Transformer (LA-MIL) for microsatellite instability classification and genetic mutation prediction in whole slide pathological images~\citep{reisenbuchler2022local}. The method starts by tessellating a gigapixel WSI into patches of identical size, removing patches containing background, artifacts, and non-tumor tissue using global thresholding and manual annotations. Then, a CNN that has been pre-trained on histopathological data compresses each patch into a feature vector, and a kNN graph matrix is constructed to describe the spatial relations between patches. A local attention Transformer computes the attention between each patch and its neighbors from the graph matrix. Not only does LA-MIL provide promising performance, but it also permits the visualization of local attention for interpreting the contribution of each patch to the classification prediction. In~\citep{zheng2022multi}, Zheng \textit{et al.} propose Multi-transSP for the survival prediction of nasopharyngeal carcinoma patients from CT and tabular data. Multi-transSP exploits the capabilities of CNNs to extract representative features and the capability of Transformers to fuse features. ResNet18~\citep{he2016deep} first extracts features from the 2D CT slices, which are concatenated with the feature representation of the tabular data generated by a linear layer. The output features are fused by a Transformer, which is then followed by a fully-connected layer to generate a survival prediction. }

Rather than pre-training the CNN backbone of the hybrid model, \citep{wang2021transpath} pre-train the entire CNN-Transformer (designated as TransPath) using a self-supervised learning method, BYOL~\citep{grill2020bootstrap}. In addition, the authors develop a token-aggregation and excitation (TAE) module for use with the MSA output in the ViT~\citep{dosovitskiy2020image}. Specifically, the TAE module first averages all token embeddings, then applies two sets of linear projection and activation functions to excite the averaged embeddings, which are then re-projected to the MSA output. According to~\citep{wang2021transpath}, combining MSA and TAE enables the Transformer to consider sufficient global information since each element in the output is the aggregated outcome of all input tokens. They conduct extensive experiments against several other Transformer-based networks on several benchmark histopathology image classification datasets and demonstrate superior performance.

Several studies suggest that even without pre-training, Transformer may be an effective complement to CNNs for feature extraction in a hybrid model. \citep{gao2021instance} propose the instance-based ViT (i-ViT) for subtyping renal cell carcinoma in histopathological image. Their framework begins by extracting nuclei-containing image patches (regarded as instance-level patches) and the corresponding nuclei grades and sizes from an input histopathology image. The patches are sorted by nucleus grade and size, and a predefined number of patches is concatenated and then used as the input to a light CNN. The output embeddings, along with additional embeddings containing information on the nuclei grades and positions relative to the entire image, are sent into a ViT~\citep{dosovitskiy2020image}. The ViT captures cellular level and cell-layer level features for subtyping. The authors train and assess the i-ViT using a dataset of 1,163 ROIs/pictures taken from 171 whole slide images, and the i-ViT achieves improved performance than the CNN-based baselines. In~\citep{he2021global}, He \textit{et al.} propose a hybrid model for brain age estimation that does not require pre-training. Their model consists of two paths: a global path that extracts global contextual information from the whole brain MRI 2D slice, and a local path that extracts local features from image patches segmented from the 2D slice. Each path has a CNN backbone for generating high-level features from the input image/patches. Following that, a ``global-local Transformer"~\citep{he2021global} is used to aggregate the features from the two paths for brain age estimation. With less than 8,000 training samples, their model trained-from-scratch performs noticeably better in comparison to a range of CNN and Transformer baselines. Although the studies discussed in this paragraph are trained on datasets with limited samples, they still outperform the CNN-based baselines, revealing the promising potential of hybrid models for data-limited applications. {P{\l}otka \textit{et al.} propose BabyNet~\citep{plotka2022babynet} that advances a 3D ResNet-based network with an MHSA module for fetal birth weight prediction. BabyNet is similar to BoT~\citep{srinivas2021bottleneck} in that it replaces the bottleneck convolution block with an MHSA to aggregate local and global feature representations more effectively. Unlike BoT, the MHSA module of BabyNet uses temporal positional embedding for temporal analysis between frames and relative positional embedding for encoding spatial correspondence within frames. BabyNet outperforms several comparative learning-based models with accuracy comparable to human experts.}

%\subsubsection
\underline{Pure ViT:}
The aforementioned models bridge CNN backbones with Transformers. Nevertheless, pure Transformers have also been shown to be effective for medical image classification when pre-trained. \citep{mondal2021xvitcos} develop a multi-stage transfer learning strategy for adapting the original ViT~\citep{dosovitskiy2020image} to COVID-19 classification tasks. Specifically, they adopt the ViT that is trained on ImageNet~\citep{deng2009imagenet, russakovsky2015imagenet} and fine-tune it using images from the target domain. Their method is tested on two publicly available datasets, namely the COVIDx-CT-2A~\citep{COVIDxCT-2A} and CheXpert~\citep{irvin2019chexpert}, and outperforms a variety of baseline methods in terms of classification accuracy. Likewise, ~\citep{yu2021mil} propose MIL-VT that fine-tunes the ViT pre-trained on ImageNet for retinal disease classification. The pre-trained ViT is first fine-tuned on an in-house large-scale fundus image dataset ($>300,000$ fundus images), and subsequently on two publicly available datasets (APTOS~\citep{APTOS2019} and RFMiD2020~\citep{RFMiD2020}) for downstream classification tasks. In the original ViT, only the features corresponding to the ``classification token"~\citep{dosovitskiy2020image} are sent to an MLP for final classification, with the features extracted from the image patches being neglected. Yu \textit{et al.} hypothesize that the features from image patches might contain important complementary information. Thus, they introduce an additional Multiple Instance Learning module (referred to as a ``MIL head"~\citep{yu2021mil}) that aggregates the features extracted from the patches and then performs prediction using the aggregated features. MIL-ViT backpropagates the loss into ViT during training through two paths: one via the MLP classifier in ViT and another via the added ``MIL head". During inference, the final prediction is made by averaging the output logits from the two paths. \citep{matsoukas2021time} compare ResNet~\citep{he2016deep} and DeiT~\citep{touvron2021training} side-by-side with three scenarios: training-from-scratch (\textit{i.e.}, without pre-training), supervised pre-training on ImageNet~\citep{deng2009imagenet}, and self-supervised pre-training on medical images in addition to the supervised pre-training. On three benchmark datasets, they empirically find that ResNet outperforms DeiT when trained from scratch, and this performance gap could be closed with the supervised pre-training. Moreover, they show that DeiT performs slightly better than ResNet with the additional self-supervised pre-training on medical images, further demonstrating the potential of self-supervised pre-training of pure Transformers for medical image classification. {In \citep{saeed2022tmss}, the authors propose TMSS for the joint prediction of a patient's survival risk score and tumor segmentation using PET/CT and electronic health records (EHR). The input PET/CT is evenly divided into patches, linearly embedded, and then concatenated with the linear embedding of the patient's EHR. The output is then fed into a ViT~\citep{dosovitskiy2020image} but without the class token. After that, The output of the ViT is sent to a multi-task logistic regression model that predicts survival risk scores and a CNN decoder that generates the segmentation mask. The model achieves superior performance on the HECKTOR dataset~\citep{oreiller2022head} when compared to competing models.}

\underline{3D modeling:} 
To date, the majority of Transformers for medical image classification has concentrated on 2D applications for various reasons, including reduced computational complexity and the ability to directly use models pre-trained on large-scale natural images ({\it e.g.}, ImageNet). However, since most medical imaging modalities produce 3D images, developing efficient Transformers for 3D classification is anticipated to receive an increased attention in the near future. \citep{xie2021unified} develop a Universal Self-supervised Transformer (USST) that can be pre-trained using both 2D and 3D images jointly. Specifically, the authors propose the switchable patch embedding (SPE) for use in the Pyramid Vision Transformer (PVT)~\citep{wang2021pyramid}, which adapts to the dimensionality of the input image by switching between 2D and 3D patch embedding. The USST pre-training framework is developed based on the student-teacher paradigm, in which both the student and teacher paths share the identical architecture, but the teacher path is updated using an exponential moving average of the weights of the student path. The authors use $>5,000$ 3D CT images and $>100,000$ 2D chest X-rays to pre-train the USST framework. The pre-trained Transformers is then fine-tuned on multiple 2D and 3D classification tasks, with the USST framework considerably outperforming other widely used pre-training frameworks on downstream tasks. {To achieve a similar objective on dimension-independent pre-training, \citep{cai2022uni4eye} propose a self-supervised learning method to pre-train ViT~\citep{dosovitskiy2020image} on both 2D and 3D ophthalmic images for downstream ophthalmic disease classification tasks. A unified patch embedding module is developed to extract a fixed number of 2D/3D patches from the input based on random masking. The extracted patches are then passed to a ViT~\citep{dosovitskiy2020image} and two decoders for self-supervised learning to reconstruct the original and the gradient images by carrying out the masked image modeling task~\citep{he2022masked, xie2022simmim}. This Transformer-based model is pre-trained, fine-tuned, and then evaluated on >$95,000$ ophthalmic images with six different classification tasks, demonstrating state-of-the-art performance on all of the evaluated tasks.}

%\subsubsection
 {
\underline{Non-Euclidean imaging:}
Functional magnetic resonance imaging (fMRI) is widely used to capturing the temporal signal of neural activity. Estimation of brain activity can be measured by functional connectivity (FC), the degree of temporal correlation between regions of the brain. Transformer also shows superiority and potential in analysis of brain connectome. ~\citep{NEURIPS2021_22785dd2} propose a GNN and Transformer hybrid model in gender classification on resting-state fMRI and task decoding for task fMRI, with a dynamic GNN enhanced by an elaborate spatial attention learning the representation of the brain connectome from a single time-step fMRI, and a single-headed Transformer encoder integrating attended features temporally. Transformer together with dynamic GNN is capable to capturing characteristics of functional connectivity which fluctuates over time.  BolT ~\citep{bedel2022bolt} exploits a cascade of Transformer blocks to encode local representations of FC, which is performed on temporally-overlapping windows. BolT comprises a cross-window attention module, with the extent of window overlap progressively, to enhance sensitivity to the diverse time scales of FC features. The integration ability from cascaded Transformer promises BolT to achieve the state-of-the-art in HCP gender prediction and cognitive task classification ~\citep{van2013wu}, and autism spectrum disorder detection task ~\citep{di2014autism}. \citep{dai2022brainformer} take the point that FC feature suffers from the insufficient representation ability and coarse granularity. They proposed BrainFormer, a convolution-transformer hybrid architecture that employs a 3D CNN backbone modeling the detailed and informative features from fMRI volume. BrainFormer inserts CNN-based attention blocks into backbone in shallow layers, capturing the spatial correlation. And it exploits transformer-based attention blocks in deep layers to fuse the global information. The effectiveness and generalizability of this method is evaluated on ABIDE~\citep{di2014autism}, ADNI~\citep{petersen2010alzheimer}, MPILMBB~\citep{mendes2019functional}, ADHD-200~\citep{bellec2017neuro}, and ECHO, with diseases of autism, ALzheimer's disease, depression, attention deficit hyperactivity disorder, and headache disorders. \citep{yu2022disentangling} propose a Twin-Transformers to simultaneously capture temporal and spatial features from fMRI. With brain signal matrix as input, the spatial Transformer focuses on non-overlapping spatial patches and the temporal Transformer takes non-overlapping temporal patches as tokens. In other scenarios in neural imaging, \citep{dahan2022surface} extend ViTs to non-Euclidean manifolds cortical surface and propose the Surface Vision Transformer (SiT) for sequence-to-sequence modelling surfaces with projection to a regularly tessellated icosphere. SiT proves a certain level of transformation invariance without introducing strong inductive bias into framework. \citep{cheng2022spherical} propose a spherical Transformer in quality assessment of cortical surface, represented by triangular meshes and mapped onto a spherical manifold. The spherical Transformer shows its potential in extracting the structural and contextual pattern among vertices.
}

In summary, Transformer-based medical image classification still relies heavily on pre-training using large-scale datasets, either supervised or self-supervised. On the other hand, for applications with limited data availability, initializing Transformers with weights pre-trained on natural images is found to be beneficial for improving performances. However, without pre-training and access to large-scale training data, Transformers may not be more effective than CNNs for medical image classification. Moreover, the majority of the existing Transformer-based models focuses on 2D applications. With a growing research interest in Transformers, we anticipate that further efforts will be directed toward developing Transformer-based models for 3D classification applications.

% Aspects:
%   Intuitions of using Transformers for classification
%       Contextual information & long-range dependencies ✔
%       Dataset size & Model size ✔
%       Pre-training ✔
%       More robust to distortions ✔
%   Transformer architectures
%   2D vs 3D
%   Modalities
%       X-ray
%       CT
%       MRI
%   Pre-training
% References:
%  ~\citep{park2021vision} Pre-trained backbone CNN ✔
%  ~\citep{yu2021mil} Pretrained ViT ✔
%  ~\citep{wang2021transpath} Pretrained ✔
%  ~\citep{mondal2021xvitcos} Pretrained ViT on imagenet and med. images ✔
%  ~\citep{matsoukas2021time} Pretrained ✔
%  ~\citep{sriram2021covid} first classification Trans model, pretraining CNN on large-scale datasets (MIMIC-CXR, CheXpert) ✔
%  ~\citep{he2021global} no pre-training ✔
%  ~\citep{gao2021instance} no pre-training ✔
%  ~\citep{xie2021unified} Pre-trained 2D & 3D Transformers ✔
%  ~\citep{dai2021transmed} hybrid Transformer-CNN model ✔
% Oct. 2022 New:
% \citep{zhao2022setmil} hybrid Transformer-CNN model ✔
% \citep{zheng2022kernel} efficient Transformer for Whole Slide Image ✔
% \citep{lv2022joint} hybrid Transformer-CNN model; multi-scale ✔
% \citep{reisenbuchler2022local} Multi-scale ✔
% \citep{wu2022seatrans} ✔
% \citep{almalik2022self} ?
% \citep{xu2022remixformer} ?
% \citep{plotka2022babynet} ✔
% \citep{zheng2022multi} ✔
% \citep{saeed2022tmss} ✔
% \citep{cai2022uni4eye} pretraining ✔

\subsection{Medical image detection} \label{sec.detection} 
%(0.5page, Junyu Chen / Ce Wang ) 
\begin{table*}[h!]
 \centering
\resizebox{\textwidth}{!}{
\begin{tabular}{p{0.05cm}lllllp{3.5cm}lp{3.5cm}rp{7cm}}
\hline
  & Reference  &\multicolumn{1}{l}{Architecture} & \multicolumn{1}{l}{2D/3D}   & \multicolumn{1}{l}{Pre-training} & \multicolumn{1}{l}{\#Param} & \multicolumn{1}{l}{Detection Task} & \multicolumn{1}{l}{Modality} & \multicolumn{1}{l}{Dataset}  & \multicolumn{1}{r}{ViT as Enc/Inter/Dec} & \multicolumn{1}{l}{Highlights}  \\
\hline\\
\multirow{26}{*}{\rotatebox{90}{Detection}}
&COTR~\citep{shen2021cotr} & Conv-Transformer Hybrid & 2D & No & N & Polyp Detection & Colonoscopy & CVC-ClinicDB~\citep{bernal2015wm}, ETIS-LARIB~\citep{silva2014toward}, CVC-ColonDB~\citep{bernal2012towards} & Yes/No/Yes & Convolution layers embedded between Transformer encoder and decoder to preserve feature structure.\\
&~\citep{mathai2022lymph} & Conventional Transformer & 2D & No & N & Lymph Node Detection & MRI & Abdominal MRI & Yes/N.A./Yes & DETR applied to T2 MRI.\\
&~\citep{jiang2021rdfnet} & Conv-Transformer Hybrid & 2D & No & N & Dental Caries Detection & RGB & Dental Caries Digital Image & No/Yes/No & Augment YOLO by applying Transformer on the features extracted from the CNN encoder.\\
&TR-Net~\citep{ma2021Transformer} & Conv-Transformer Hybrid & 3D & No & N & Stenosis Detection & CTA & Coronary CT Angiography & No/Yes/N.A. & CNN applied to image patches, followed by a Transformer to learn patch-wise dependencies.\\
&DATR~\citep{zhu2022datr} & Conv-Transformer Hybrid & 2D & Pre-trained Swin & N & Landmark Detection & X-ray & Head~\citep{wang2016benchmark}, Hand~\citep{payer2019integrating}, and Chest~\citep{zhu2021you} & Yes/N.A./No & The integration of a learnable diagonal matrix to Swin Transformer enables the learning of domain-specific features across domains.\\
&SATr~\citep{li2022satr} & Conv-Transformer Hybrid & 2D & No & N & Lesion Detection & CT & DeepLesion~\citep{yan2018deeplesion} & No/Yes/No & Introduce slice attention Transformer to commonly used CNN backbones for capturing inter- and intra-slice dependencies.\\
&~\citep{tian2022contrastive} & Conv-Transformer Hybrid & 2D+$t$ & Pre-trained CNN & N & Polyp Detection & Colonoscopy & Hyper-Kvasir~\citep{borgli2020hyperkvasir}, LDPolypVideo~\citep{ma2021ldpolypvideo} & No/Yes/N.A. & A weakly-supervised framework with a hybrid CNN-Transformer model is developed for polyp detection.\\
&SCT~\citep{windsor2022context} & Conv-Transformer Hybrid & 2D & Pre-trained CNN & N & Spinal Cancer Detection & MRI & Whole Spine MRI & No/Yes/N.A. & A Transformer that considers contextual information from the multiple spinal columns and all accessible MRI sequences is used to detect spinal cancer.  \\
\hline
\hline
&Reference  &\multicolumn{1}{l}{Architecture} & \multicolumn{1}{l}{2D/3D}    & \multicolumn{1}{l}{Pre-training} & \multicolumn{1}{l}{\#Param}   & \multicolumn{1}{l}{Registration Task} & \multicolumn{1}{l}{Modality}& \multicolumn{1}{l}{Dataset}  &\multicolumn{1}{r}{ViT as Enc/Inter/Dec} & \multicolumn{1}{l}{Highlights}  \\
\hline\\
\multirow{22}{*}{\rotatebox{90}{Registration}}
&ViT-V-Net~\citep{chen2021vit}  & Conv-Transformer Hybrid & 3D & No & 110.6M & Inter-patient & MRI & Brain MRI & Yes/No/No & ViT applied to the CNN extracted features in the encoder. \\
&TransMorph~\citep{chen2021transmorph} & Conv-Transformer Hybrid & 3D & No & 46.8M  & Inter-patient, Atlas-to-patient, Phantom-to-patient & MRI, CT, XCAT & IXI*, OASIS~\citep{marcus2007open}, Abdominal and Pelvic CT, ~\citep{segars2013population}  & Yes/No/No & Swin Transformer is used as the encoder for extracting features from the concatenated input image pair. \\
&DTN~\citep{zhang2021learning} & Conv-Transformer Hybrid & 3D & No & N  & Inter-patient & MRI & OASIS~\citep{marcus2007open} & No/Yes/No & Separate Transformers are employed to capture inter- and intra-image dependencies within the image pair. \\
&PC-SwinMorph~\citep{liu2022pc} & Conv-Transformer Hybrid & 3D & No & N  & Inter-patient & MRI & CANDI~\citep{kennedy2012candishare}, LPBA-40~\citep{shattuck2008construction} & No/No/Hybird & Patch-based image registration; Swin Transformer is used for stitching the patch-wise deformation fields.\\
&Swin-VoxelMorph~\citep{zhu2022swin} & Conv-like Transformers & 3D & No & N  & Patient-to-atlas & MRI & ADNI~\citep{mueller2005ways}, PPMI~\citep{marek2011parkinson} & Yes/N.A./Yes & Swin-Transformer-based encoder and decoder network for inverse-consistent registration.\\
&XMorpher~\citep{shi2022xmorpher} & Conv-like Transformers & 3D & No & N  & Inter-patient & CT & MM-WHS 2017~\citep{zhuang2016multi}, ASOCA ~\citep{gharleghi2022automated} & Yes/N.A./Yes & Two Swin-like Transformers are used for fixed and moving images, with cross-attention blocks facilitating communication between Transformers.\\
&C2FViT~\citep{mok2022affine} & Conv-like Transformers & 3D & No & N  & Template-matching, patient-to-atlas & MRI & OASIS~\citep{marcus2007open}, LPBA-40~\citep{shattuck2008construction} & Yes/N.A./N.A. & Multi-resolution Vision Transformer is used to tackle the affine registration problem with brain MRI.\\
\hline
&\multicolumn{2}{l}{*https://brain-development.org/ixi-dataset/}
\end{tabular}%
}
\caption{The summarized review of Transformer-based model for medical image detection (upper panel) and registration (lower panel). "N.A." denotes for not applicable for intermediate blocks or decoder module. "N" denotes not reported or not applicable on the number of model parameters. "$t$" denotes temporal dimension.}
\label{tab:detection}
\end{table*}

The use of Transformers for object detection in natural images is pioneered by Carion \textit{et al.} in DETR~\citep{carion2020end}. DETR makes use of both the encoder and decoder from the original {Language Transformer} used in NLP~\citep{vaswani2017attention}, whereas ViT~\citep{dosovitskiy2020image} borrows only the encoder. In computer vision, efforts have been made to augment both Transformer encoder-decoder (i.e., DETR)~\citep{zhu2020deformable, zheng2020end, sun2021rethinking}  and Transformer encoder-only (i.e., ViT)~\citep{beal2020toward, li2022exploring} designs for object detection, all of which have shown demonstrable performances. On the one hand, DETR's Transformer decoder learns to make direct set predictions such that duplicate bounding box predictions are suppressed, eliminating the post-processing procedures for the predictions (\textit{e.g.}, non-maximal suppression). In the field of medical imaging, a few Transformer-based object detection methods have been developed based on DETR~\citep{shen2021cotr, mathai2022lymph}. However, it has been discovered that DETR takes much longer training epochs for convergence than ConvNet-based models~\citep{zhu2020deformable, fang2021you, beal2020toward}. On the other hand, using only the Transformer encoder may benefit from the transferability of the encoders pre-trained on large-scale datasets (\textit{e.g.}, ImageNet~\citep{deng2009imagenet, russakovsky2015imagenet}), thereby accelerating convergence. Furthermore, combining these encoders with ConvNets introduces additional inductive bias, reducing the amount of data needed to construct an effective model. Several attempts have been made in medical imaging that uses Transformer encoders as a component of the feature extractor in conjunction with ConvNets for bounding box prediction~\citep{jiang2021rdfnet, li2022satr} and for applications where the bounding boxes are not needed~\citep{ma2021Transformer, zhu2022datr}. While the advantages of Transformers for image classification remain relevant to object detection (i.e., properties \hyperlink{M1}{$M_1$}, \hyperlink{M4}{$M_4$}, \hyperlink{M4}{$M_3$}, \hyperlink{M5}{$M_5$}, \hyperlink{C1}{$C_1$}, and \hyperlink{C2}{$C_2$}), the main advantage is that:
\begin{itemize}
\item The self-attention mechanism computes globally or with a very large kernel, making Transformer more ideal for comprehending contextual information contained in an image, which is crucial for object detection. [Property \hyperlink{M1}{$M_1$}]
%\item 
%[Property $xxx$]
\end{itemize}

%In the next subsection, we briefly discuss the aforementioned applications of Transformers in medical image detection.

%\subsubsection{Transformers for Medical Image Detection}
{\underline{Transformer as encoder and decoder:}} ~\citep{shen2021cotr} propose a convolution-in-Transformer (COTR) network for polyp detection in colonoscopy. COTR is built on top of DETR with an aim to address the slow convergence issue with DETR. Because the Transformer encoder in DETR operates on flattened image features (i.e., vectors), it may lead the image feature structures to become disorganized. The authors thus embed convolution layers between the Transformer encoder and decoder to reconstruct the flattened vectors into high-level image features. This preserves the feature structures within the network and increases convergence speed. Additionally, DETR is shown to effectively detect lymph nodes in T2 MRI. ~\citep{mathai2022lymph} demonstrate using a publicly available dataset that DETR, with a little tweaking to the loss functions, could surpass multiple state-of-the-art lymph node detection methods by a large margin. 

{\underline{Hybrid CNN and Transformer-encoder:}} ~\citep{jiang2021rdfnet} augment YOLO~\citep{redmon2016you} with a Transformer encoder for dental caries detection. Specifically, a sequence of convolution and pooling operations was followed by a Transformer to extract deep features at a lower resolution. In an identical manner to YOLO, the features at all resolutions are sent into the neck module and subsequently the detection head for bounding box prediction. Their model exhibits improved accuracy and average precision compared with the ConvNet-based baselines. In~\citep{ma2021Transformer}, Ma \textit{et al.} propose TR-Net, a Transformer-based network for detecting coronary artery stenosis in Coronary CT angiography. The authors begin by reconstructing multiplanar reformatted (MPR) images from coronary artery centerlines. The MPR images are then divided into equal-sized cubic volumes, with each volume centered on the coronary artery's centerline. After extracting semantic features from each volume using a shallow ConvNet, the features from all volumes are combined with learnable positional embeddings to preserve the volumes' ordering information. Then, the features are sent to a Transformer encoder to analyze relationships within the volume sequence. The output of the TR-Net is not a bounding box but a probability of each cubic volume having significant stenosis. Similarly, ~\citep{zhu2022datr} use a Transformer as the encoder for multi-anatomy landmark detection ~\cite{liu2010search}. The authors propose a domain-adaptive Transformer (DATR), an anatomy-aware Transformer that is invariant of the Transformer architecture and capable of operating on a variety of anatomical features. DATR is built on the basis of a pre-trained Swin Transformer~\citep{liu2021swin}, which extracts four scales of features and passes them to a ConvNet decoder. The network produces a heatmap with the highest-intensity locations corresponding to the landmarks. Last but not least, ~\citep{li2022satr} propose a slice attention Transformer (SATr) that can be plugged into existing three-slice-input ConvNet backbones to improve the accuracy of universal lesion detection (ULD) in CT. The SATr blocks are introduced between the ConvNet backbone and the feature collector to better model long-distance feature dependencies. Each SATr block calculates self-attention between and within the features of the slices. The authors demonstrate that by simply integrating the SATr into existing three-slice-input ULD models, detection accuracy could be greatly improved and reach the state-of-the-art. ~\citep{tian2022contrastive} propose a weakly-supervised framework to identify polyps from colonoscopy video frames. The authors begin by extracting features from each video frame using a pre-trained I3D network~\citep{carreira2017quo} to produce a feature token. The tokens are then sent to a Transformer for the detection of polyp frames. The authors augment the original ViT~\citep{dosovitskiy2020image} by replacing its linear embedding layers with depth-wise convolutional operations to capture local temporal relationships more effectively. In addition, a novel contrastive snippet mining strategy is proposed to extract hard and easy, normal and abnormal video frames during training for enhanced robustness in detecting subtle polyp tissues. \citep{windsor2022context} propose a context-aware Transformer for spinal cancer detection in multi-sequence spinal MRI. A pre-trained ResNet~\citep{he2016deep} is fed using 2D slices from multiple MRI sequences ({\it e.g.}, T1, T2, STIR, FLAIR, etc.) of multiple spinal columns to extract representative features. The feature vectors for each slice are then aggregated using a lightweight two-layer Transformer, along with additional embedding vectors specifying the level of each input vertebra and the MRI sequence employed. An attention operation is used at the end of the network to merge the features of the same vertebra. Then, the output is converted by a linear layer to produce the prediction for the corresponding vertebra. The authors demonstrate that their method leads to improved accuracy compared with a well-established method, SpineNet~\citep{jamaludin2017automation}.

In summary, the existing works (also as listed in Table \ref{tab:detection}, the top part) have demonstrated the potential for Transformer-based networks to be used for medical image detection. {For applications that require generating bounding boxes, the Transformer encoder and decoder designs (\textit{e.g.}, DETR~\citep{carion2020end}) may be adopted to alleviate the need for expensive post-processing processes (\textit{e.g.}, non-maximal suppression).} Transformers have shown promise for detection applications, but since medical datasets are often modest in size, it may be necessary to tweak the network architecture or training strategy to accelerate convergence and reduce the amount of training data needed to develop an effective model. In other applications, pre-trained Transformer encoders on natural images may be viable for enhancing a neural network's ability to model long-distance feature dependencies without sacrificing the speed of convergence. In comparison with training from scratch, recent breakthroughs in self-supervised pre-training of Transformers for object recognition have shown significant performance improvements~\citep{dai2021up, dong2021peco}. In addition, studies have revealed that self-supervised pre-training strategies are useful for medical image segmentation and classification~\citep{matsoukas2021time, xie2021unified, tang2022self, karimi2021convolution}, thus we expect to witness more contributions on self-supervised learning for medical image detection.
% Base networks
% DETR:~\citep{carion2020end} ✔
% ViT:~\citep{dosovitskiy2020image} ✔

% DETR as backbone:
%  ~\citep{mathai2022lymph} DETR ✔ 
%  ~\citep{shen2021cotr} DETR ✔

% ViT as backbone:
%  ~\citep{li2022satr} ✔
%  ~\citep{zhu2022datr} ✔
%  ~\citep{ma2021Transformer} ✔
%  ~\citep{jiang2021rdfnet} ✔

% Oct. 22 new:
% ~\citep{tian2022contrastive} ✔
% ~\citep{windsor2022context} ✔

\subsection{Medical image registration} \label{sec.registration} 
%(0.5pages, 1 fig Junyu Chen / Ce Wang) 
Transformer is a viable choice for medical image registration since it has a better understanding of the spatial correspondence between and within images, and image registration is a process of establishing such correspondence between the moving and fixed images. The main advantages of applying Transformers over ConvNets to image registration are:
\begin{itemize}
    \item The self-attention mechanism in a Transformer has a large effective receptive field that encompasses the entire image (as shown in Fig. \ref{fig:ERFs}), enabling the Transformer to explicitly capture the long-range spatial relationships between points in the image~\citep{raghu2021vision, ding2022scaling}. [Property \hyperlink{M1}{$M_1$}]
    \item The majority of the learning-based deformable registration models adopt the spatial transformer network design \citep{jaderberg2015spatial}, which generates high-dimensional vector field mapping (\textit{i.e.}, one transformation for each spatial coordinate) with several million transformations per 3D volume. However, the commonly used CNN-based registration models are often of small parameters (\textit{e.g.}, VoxelMorph-1 \citep{balakrishnan2019voxelmorph} has about 0.3M parameters). Therefore, the Transformer's superior scaling behavior of a large-scale model size over that of ConvNets may contribute to the establishment of a more precise spatial correspondence [Property \hyperlink{C1}{$C_1$}]. 
\end{itemize}
Whereas with ConvNets, due to the limited receptive fields of convolution operations, these long-range spatial relationships can only be implicitly modeled in the deeper layers. As a result, Transformers is a more compelling contender than ConvNets for serving as the backbone for deep-learning-based image registration.

\begin{figure*}
\centering
\includegraphics[width=0.98\textwidth]{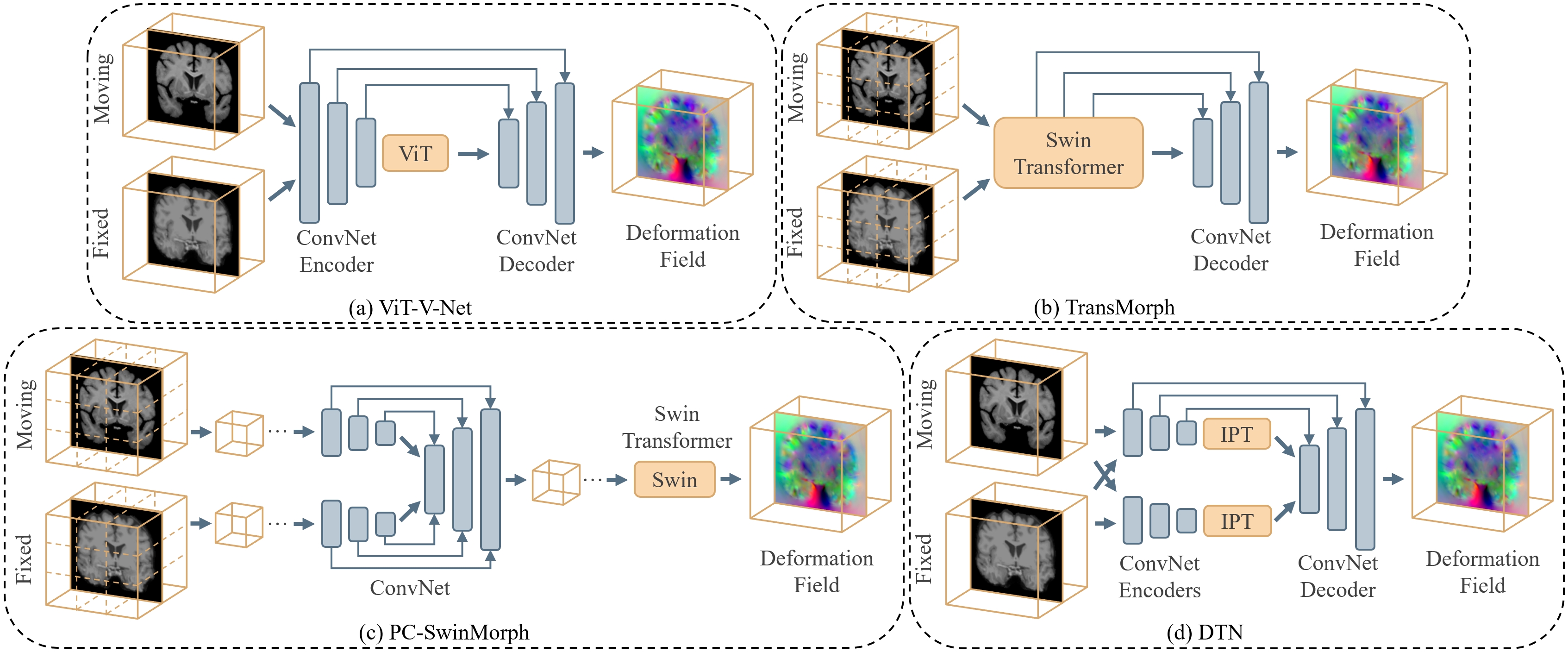}
\caption{The schematic illustration of the Transformer-based image registration networks. (a) ViT-V-Net~\citep{chen2021vit}. (b) TransMorph~\citep{chen2021transmorph}. (c) PC-SwinMorph~\citep{liu2022pc}. (d) DTN~\citep{zhang2021learning}. These network architectures are based predominately on the hybrid ConvNet-Transformer design.} \label{fig:trans_reg}
\end{figure*}

Transformers have been used predominantly for 3D registration applications such as inter-patient and atlas-to-patient brain MRI registration~\citep{chen2021vit, chen2021transmorph, zhang2021learning, liu2022pc}, as well as phantom-to-CT registration~\citep{chen2021transmorph}. As shown in Fig. \ref{fig:trans_reg}, Transformer-based registration networks primarily employ hybrid architectures, with Transformers used in the encoding stage to capture the spatial correspondence between the input moving and fixed images, and ConvNet encoders used to generate and refine the deformation fields. In the next subsection, we briefly summarize recent works on Transformer-based medical image registration (as listed in Table \ref{tab:detection}, the bottom part) and Fig.~\ref{fig:trans_reg} provides a schematic illustration of these approaches.

%\subsubsection{Transformers for Medical Image Registration} 
The use of Transformers for 3D medical image registration is first investigated by ~\citep{chen2021vit}. The authors propose a hybrid model, ViT-V-Net, in which the encoder is composed of convolutional layers, down-samplings, and a ViT~\citep{dosovitskiy2020image}, while the decoder is composed of consecutive convolutional layers and up-sampling operations. Long skip connections similar to those used in V-Net~\citep{milletari2016v} are used to maximize the flow of information between the encoding and decoding stages. This model first extracts high-level features from the concatenated image pair using the convolutional layers and down-samplings. Then, ViT is applied to capture the long-range spatial correspondence between the high-level features. The decoder then uses the ViT's output to generate a dense displacement field that warps the moving image. This model outperforms the widely used learning-based model VoxelMorph~\citep{balakrishnan2019voxelmorph} for inter-patient registration on an in-house brain MRI dataset, while using identical training procedures and having a comparable computational cost. Later, ~\citep{chen2021transmorph} extend this model and propose TransMorph by substituting a Swin Transformer~\citep{liu2021swin} for the encoder, resulting in more direct and explicit modeling of the spatial correspondences within the input image pairs. Additionally, the authors present the diffeomorphic and Bayesian variants of TransMorph, the latter of which integrates Monte-Carlo dropout layers~\citep{gal2016dropout} into the Swin Transformer encoder to enable registration uncertainty estimates. TransMorph is rigorously evaluated against a variety of baseline methods, including the traditional and ConvNet-based registration methods. Additionally, TransMorph is compared against several hybrid Transformer-ConvNet and pure Transformer network designs that demonstrate superior performances in other tasks (\textit{e.g.}, image segmentation). TransMorph outperforms the baseline methods in terms of Dice scores on two in-house datasets and the IXI\footnote{\url{http://brain-development.org/ixidataset/}} brain MR dataset. {\citep{shi2022xmorpher} propose XMorpher, in which a Swin-like Transformer is separately applied to moving and fixed images. Unlike Swin, XMorpher uses cross-attention to enable the exchange of information between a pair of features from moving and fixed images. Encoder and decoder of XMorpher are both Transformer-based, with the decoder being symmetric to the encoder while the patch merging layer being replaced by transposed convolution to increase the resolution of the features in the decoder. In a similar fashion, ~\citep{zhu2022swin} propose Swin-VoxelMorph, a pure Tranformer-based encoder and decoder network for inverse-consistent image registration. In contrast to XMorpher, Swin-VoxelMorph takes concatenated fixed and moving images as inputs and outputs two deformation fields for inverse and forward registration. In addition, the decoder of Swin-VoxelMorph uses patch expanding as opposed to transposed convolution to increase feature resolution.} Meanwhile, ~\citep{zhang2021learning} propose a dual Transformer network (DTN) for 3D medical image registration. DTN is similar to ViT-V-Net in that the Transformer is applied to the high-level features extracted by convolutional layers and down-sampling operations. However, in addition to the encoder that extracts \textit{inter}-image dependencies from the concatenated moving and fixed images, DTN employs two additional encoders with shared weights to extract \textit{intra}-image dependencies from each image. Each encoder of DTN is composed of a U-Net encoder and an Image Processing Transformer (IPT)~\citep{chen2021pre}. The output features from the three encoders are concatenated and sent to a ConvNet decoder to produce a dense displacement field. The authors evaluate DTN for the inter-patient registration task on the OASIS brain MRI dataset~\citep{marcus2007open}, for which it outperforms baseline methods in terms of Dice and deformation regularity. Taking a different route, Transformer is also used to refine deformation fields. In~\citep{liu2022pc}, Liu \textit{et al.} propose PC-SwinMorph, which is a patch-based image registration framework that uses contrastive learning on features extracted from the fixed and moving patches, followed by a ConvNet decoder that decodes the features and generates deformation field for the associated patch. The authors then employ two consecutive Swin Transformer blocks that learn to fuse and stitch patch-wise deformation fields together. {\citep{mok2022affine} propose C2FViT to tackle affine registration for brain MRI. C2FViT employs a multi-resolution strategy in which affine transformation parameters are estimated by a set of ViTs from low resolution input to high resolution. Comprehensive experiments reveal that C2FViT outperforms the comparative learning-based affine registration methods while being more robust to unseen datasets.}

Despite the promising potential demonstrated by the aforementioned Transformer-based registration methods, the application of Transformers to medical image registration is still in its infancy. Advanced Transformer training strategies and more complicated self-attention designs, both of which have been found to improve classification and segmentation performance~\citep{xie2021unified, tang2022self}, have not yet been evaluated for registration.

% Aspects:
%   Intuitions of using Transformers for image registration ✔
%   Hybrid Transformer CNN architectures ✔
%   2D vs 3D ✔
%   Pre-training ? (TBD)
% References:
%  ~\citep{chen2021vit} ✔
%  ~\citep{chen2021transmorph} ✔
%  ~\citep{zhang2021learning} ✔
% ~\citep{xu2022svort}
% ~\citep{zhu2022swin} ✔
% ~\citep{shi2022xmorpher} ✔

\subsection{Medical image reconstruction}
\label{sec.recon} 
\begin{table*}[t]
	\centering
	\resizebox{\textwidth}{!}{
		\begin{tabular}{p{0.05cm}llllrp{4cm}rp{7cm}}   
			\hline
			&Reference  &\multicolumn{1}{l}{Architecture} & \multicolumn{1}{l}{2D/3D}   & \multicolumn{1}{r}{\#Param}   & \multicolumn{1}{l}{Modality} & \multicolumn{1}{l}{Dataset}  &\multicolumn{1}{r}{ViT as Enc/Inter/Dec} & \multicolumn{1}{l}{Highlights}  \\
			%\hdashline
			\hline
			\multirow{30}{*}{\rotatebox{90}{ Reconstruction}}&ReconFormer~\citep{guo2022reconformer}& Conv-Transformer Hybrid&2D&1.414M&MRI&fastMRI~\citep{knoll2020fastmri}, HPKS~\citep{jiang2019identifying}&No/Yes/No& The Pyramid Transformer Layer (PTL) introduces a locally pyramidal but globally columnar structure.\\
			&DSFormer~\citep{zhou2022dsformer}&Conv-Transformer Hybrid&2D&0.18M&MRI&Multi-coil Brain Data from IXI*&No/Yes/No&The proposed Swin Transformer Reconstruction Network enables a self-supervised reconstruction process with lightweight backbone.\\
			&SLATER~\citep{korkmaz2022unsupervised}&Conv-Transformer Hybrid&2D&N&MRI& Single-coil Brain Data from IXI*, Multi-coil Brain Data from fastMRI~\citep{knoll2020fastmri}&No/Yes/Yes&An unsupervised MRI reconstruction design with the long-range dependency of Transformers.\\
                &DuDoCAF~\citep{lyu2022dudocaf}&Conv-Transformer Hybrid &2D &1.428M &MRI &fastMRI~\citep{knoll2020fastmri}, Clinical Brain MRI Dataset &No/Yes/No & The proposed recurrent blocks with transformers are employed to capture long-range dependencies from the fused multi-contrast features maps, which boosts target-contrast under-sampled imaging. \\
                &SDAUT~\citep{huang2022swin}& Conv-Transformer Hybrid &2D &N &MRI &Calgary Campinas dataset~\citep{souza2018open} & No/Yes/No & The proposed U-Net-based Transformer combines dense and sparse deformable attention in separate stages, improving performances and speed while revealing explainability.\\
			&MIST-net~\citep{pan2021mist}&Conv-Transformer Hybrid&2D&12.0M&CT&NIH-AAPM-Mayo~\citep{mccollough2016tu}&No/Yes/No& The Swin Transformer and convolutional layers are combined in the High-definition Reconstruction Module, achieving high-quality reconstruction.\\
			&DuDoTrans~\citep{wang2021dudotrans}&Conv-Transformer Hybrid&2D&0.44M&CT&NIH-AAPM-Mayo~\citep{mccollough2016tu}, COVID-19&No/Yes/No&The Sinogram Restoration Transformer (SRT) Module is proposed for projection domain enhancement, improving sparse-view CT reconstruction.\\
			&FIT~\citep{buchholz2021fourier} &Conventional Transformer&2D&N&CT&LoDoPaB~\citep{leuschner2021lodopab}&Yes/No/Yes&The carefully designed FDE representations mitigate the computational burden of traditional Transformer structures in the image domain.\\
            &RegFormer~\citep{xia2022transformer} &Conv-Transformer Hybrid &2D &N &CT&NIH-AAPM-Mayo~\citep{mccollough2016tu} & Yes/Yes/Yes & The unrolled iterative scheme is redesigned with transformer encoders and decoders for learning nonlocal prior, alleviating the sparse-view artifacts.  \\
			\hline
			\multirow{24}{*}{\rotatebox{90}{Enhancement}}&TransCT~\citep{zhang2021transct}&Conv-Transformer Hybrid&2D&N&CT&NIH-AAPM-Mayo~\citep{mccollough2016tu}, Clinical CBCT Images&No/Yes/No&Decomposing Low Dose CT (LDCT) into high and low frequency parts, and then denoise the blurry high-frequency part with the basic Transformer structure\\
			&TED-Net~\citep{wang2021ted}&Conv-like Transformer&2D&N&CT&NIH-AAPM-Mayo~\citep{mccollough2016tu}&Yes/Yes/Yes& Their design makes use of the tokenization and detokenization operations in the convolution-free encoder-decoder architecture.\\
			&Eformer~\citep{luthra2021eformer} &Conv-Transformer Hybrid&2D&N&CT&NIH-AAPM-Mayo~\citep{mccollough2016tu}&Yes/Yes/Yes&A residual Transformer is proposed, which redesigns the residual block in the denoising encoder-decoder architecture with nonoverlapping window-based Multi-head Self-Attention (MSA).\\
                &TVSRN~\citep{yu2022rplhr}&Conv-like Transformer &3D &1.73M &CT &RPLHR-CT$^\dag$ dataset &Yes/Yes/Yes &They design an asymmetric encoder-decoder architecture composed of pure transformers. The structure efficiently models the context relevance in CT volumes and the long-range dependencies. \\
			&T$^{2}$Net~\citep{feng2021task}&Conv-Transformer Hybrid&2D&N&MRI&Single-coil Brain Data from IXI*, Clinical Brain MRI Dataset&No/Yes/Yes&The task Transformer module is designed in a multi-task learning process of super-resolution and reconstruction, and the super-resolution features are enriched with the low-resolution reconstruction features\\
                &WavTrans~\citep{li2022wavtrans}&Conv-Transformer Hybrid &2D &2.102M &MRI &fastMRI~\citep{knoll2020fastmri}, Clinical Brain MRI Dataset &No/Yes/No & The Residual Cross-attention Swin Transformer is proposed to deal with cross-modality features and boost target contrast MRI super-resolution.\\
			\hline
			&*https://brain-development.org/ixi-dataset/\\
                &$^\dag$https://github.com/smilenaxx/RPLHR-CT/\\
		\end{tabular}%
	}
	\caption{The summarized review of Transformer-based model for medical image reconstruction and enhancement. ``N" denotes not reported or not applicable on model parameters.}
	\label{tab:recon01}
\end{table*}

\begin{figure*}[h]
	\begin{center}
		\includegraphics[width=0.97\textwidth]{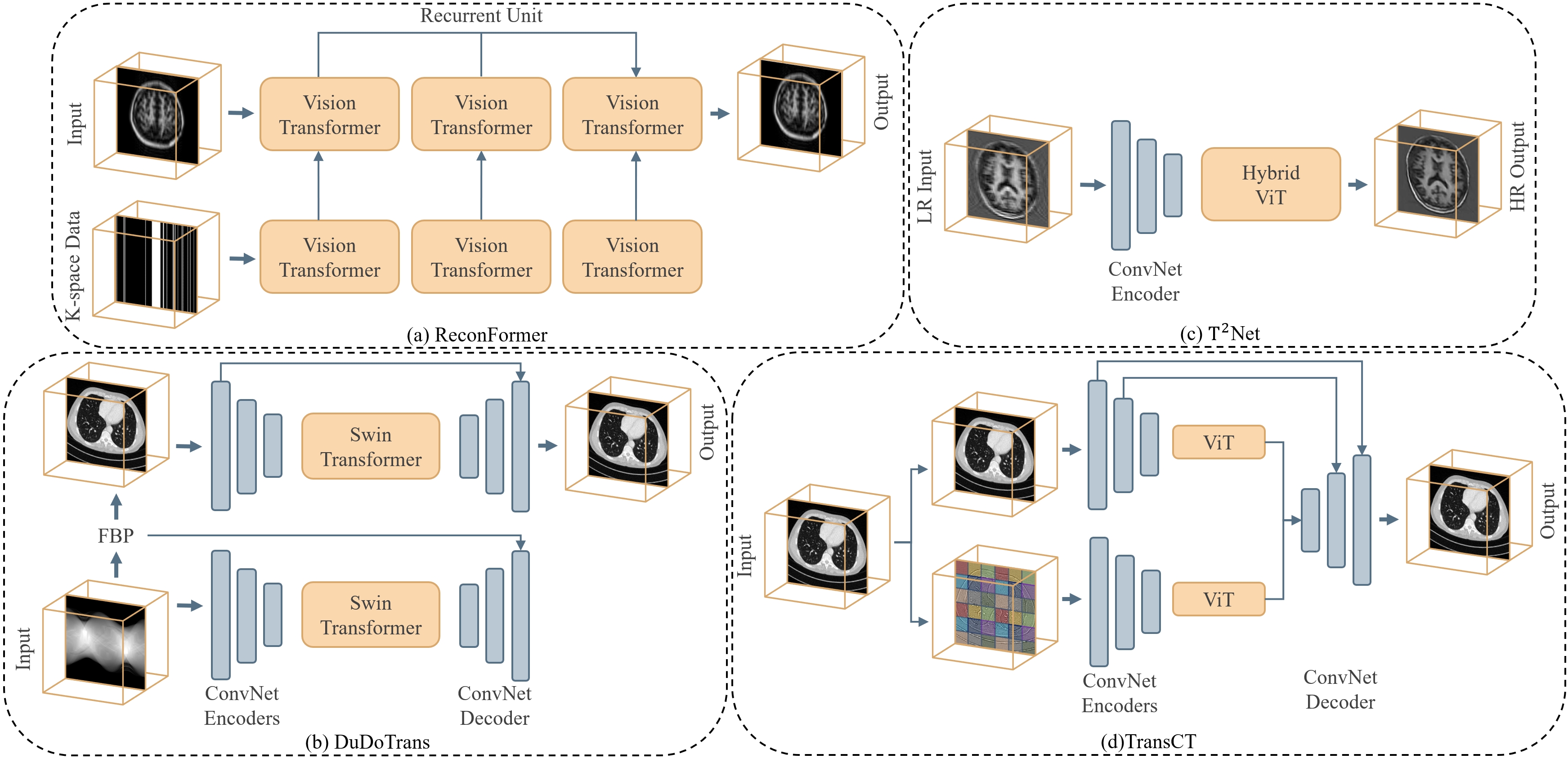}
		\caption{We illustrate the Transformer-based networks of (a) ReconFormer~\citep{guo2022reconformer} (b) DuDoTrans~\citep{wang2021dudotrans} (c) T$^{2}$Net~\citep{feng2021task} and (d) TransCT~\citep{zhang2021transct}. (a) and (b) are reconstruction models, (c) and (d) are for enhancement. These structures are based on the hybrid ConvNet-Transformer design.}
		\label{recon_framework}
	\end{center}
\end{figure*}

\begin{figure}[h]
	\centering
	\includegraphics[width=0.45\textwidth]{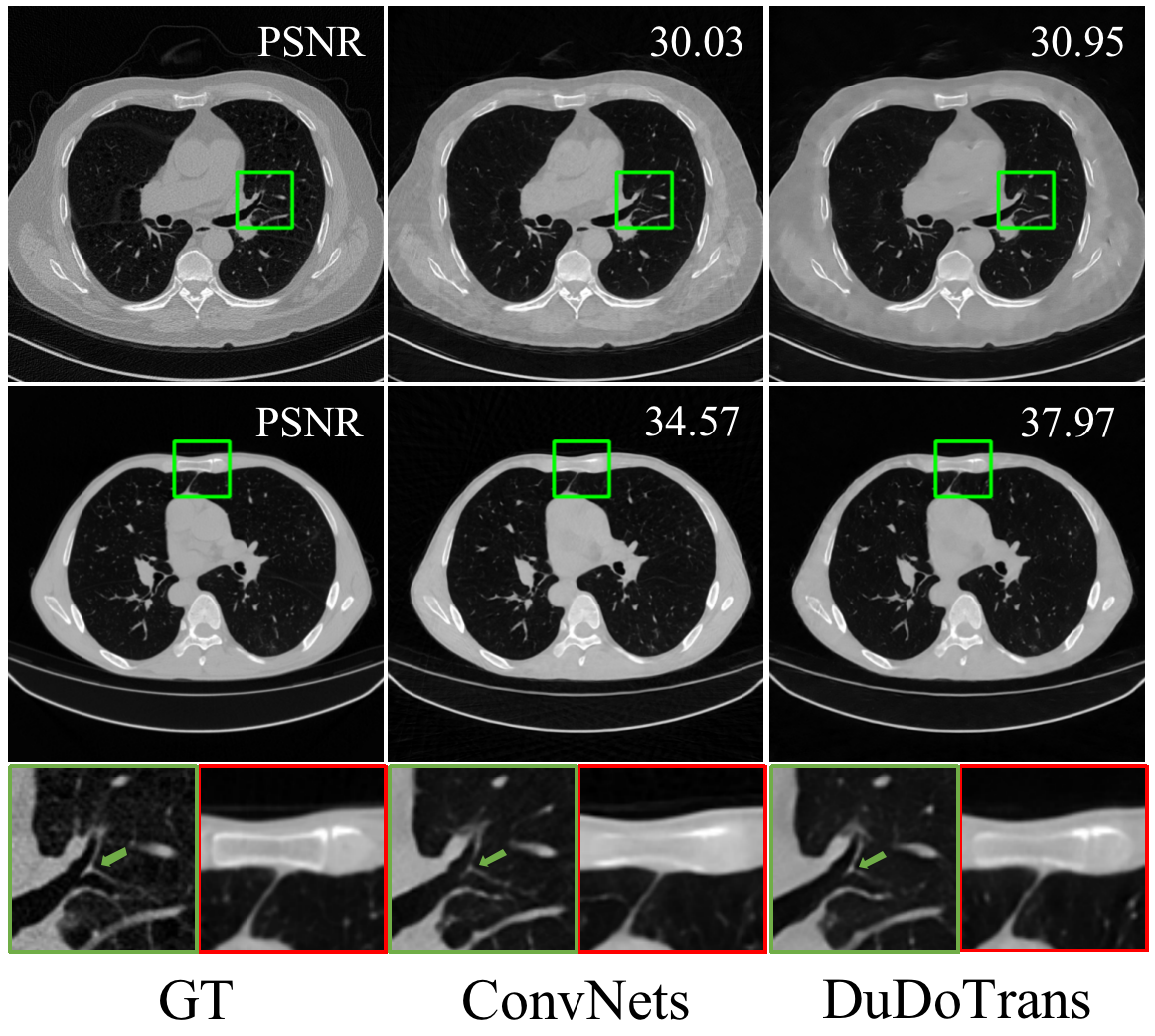}
	\caption{We visualize reconstructions of Transformer-based DuDoTrans~\citep{wang2021dudotrans} versus ConvNet with 72 and 96 sparse views on NIH-AAPM-Mayo~\citep{mccollough2016tu} dataset, and the zoom-in images are shown in the last row. With the included Property \protect\hyperlink{M2}{$M_2$}, Transformer-based DuDoTrans obtains better overall performances, especially on bones, and alleviates the FBP artifacts. While the recovered soft tissues are not as sharp as ConvNets results.}
	\label{recon_vis}
\end{figure}

%\section{Medical Image Reconstruction and Enhancement}
As the fundamental precursor to downstream medical image analysis tasks, image reconstruction aims to generate high-quality structural representations or images of external or internal tissues of the human body. However, the practical MRI and CT imaging systems suffer from either a long acquisition time or an induced radiation in the imaging process, which causes an additional stress for patients. To alleviate the above problems, downsampling the acquired signals is commonly used; however it induces a very ill-posed problem and challenges the reconstruction algorithms. With the recent development of Transformer architectures and their capability of effectively characterizing global features, as well as the dense modeling of local patches that preserves more context details, Vision Transformer have attracted researchers and shown remarkable performances in medical image reconstruction.% which can be categorized into the following two aspects:
%\emph{Medical Image Reconstruction} and \emph{Medical Image Enhancement}.

%\subsubsection{Medical Image Reconstruction}
While the under-sampling procedure alleviates the aforementioned problems, the accompanying artifacts prevent accurate clinical diagnosis; therefore, various iterative and convolutional models are proposed to suppress the artifacts. Although CNN-based post-processing and deep-unrolling methods show satisfactory performance, the global context in the structural representation is not fully captured by the spatial-split kernels, especially when context details are absent in the under-sampling scenarios. This motivates the exploration of the following key Transformer properties available for reconstruction:
\begin{itemize}
	\item The long-range dependency modeling ability of Transformer is rather valuable. As is well known, medical images, different from natural images, consist of organ anatomies and represent 2D/3D information of a human body. The global correlation is much higher than that of natural images and is thus critical to be captured. [Property \hyperlink{M1}{$M_1$}]
	\item As the fundamental procedure for diagnosis, reconstruction needs clearer anatomies. Towards this purpose, the dense modeling property and attention mechanism in Transformers assist in locating the most valuable features within the context of the whole image. [Property \hyperlink{M2}{$M_2$}]
	\item Towards combining the dense modeling of Transformer and the local context modeling of CNN, mixing them as sub-modules in a hybrid model to accommodate different modeling requirements is flexible. [Property \hyperlink{C2}{$C_2$}]
\end{itemize}
%Next, we briefly review and analyze recent works pioneering the under-sampling MRI and CT reconstruction.

%\subsubsection
\underline{Under-sampled MRI reconstruction:}
% ReconFormer
Recently, motivated by a lack of attention paid to the intrinsic multi-scale information of MRI,  ReconFormer~\citep{guo2022reconformer} designs a Pyramid Transformer Layer (PTL), which introduces a locally pyramidal but globally columnar structure. Then, via recurrently stacking the basic layer, the ReconFormer is capable of scaling the model and exploiting deep feature correlation through recurrent states in the model. Due to the use of recurrent structure, ReconFormer is lightweight and parameter-efficient, which alleviates the bottleneck that exists in previous Vision Transformer methods.
% DSFormer
To facilitate the exploration of the relative information between multi-contrast images in MRI reconstruction, DSFormer~\citep{zhou2022dsformer} proposes a novel Swin Transformer Reconstruction Network, which is based on the lightweight Swin Transformer~\citep{liu2021swin} with the backbone structure in a self-supervised reconstruction process. They use hybrid operations with both the convolutional layers and the involved Swin Transformer blocks, and condition the model with information from the reference contrast image, achieving a performance comparable to that of supervised reconstruction methods.
% SLATER
SLATER~\citep{korkmaz2022unsupervised} pioneers the unsupervised MRI reconstruction using the long-range dependency of Transformers. It decouples the traditional imaging process into a phase of deep-image-prior learning and a subsequent phase of zero-shot inference. In the former phase, the proposed adversarial Transformer model is trained to capture a prior on coil-combined, complex MR images obtained from fully-sampled acquisitions since the previously equipped CNNs prevent capturing the long-range relationship prior~\citep{zhang2019self, chen2021transunet}. In the later phase, they reconstruct the target MRI via an iterative procedure to ensure the consistency between the reconstruction and the acquisition. The method renders the potential of Transformers in purely unsupervised reconstruction setting.
% DuDoCAF
{
For efficiently reconstructing the under-sampled target-contrast MR images, DuDoCAF~\citep{lyu2022dudocaf} takes advantage of the long-range dependency modeling capability of transformers to fuse features of a reference contrast MR image. Specifically, they propose the CAF and RRT modules composed of transformer structures to first bridge the cross-modality relationship between reference and target k-space data. Then, with recurrent dual-domain learning, they gain remarkable performances and fast imaging speed.
% SDAUT 
As known, the high-computational cost of self-attention in Transformer hinders its further development in medical imaging. To tackle the issue, SDAUT~\citep{huang2022swin} proposes a U-Net-based Transformer that combines dense and sparse deformable attention in separate stages. These two involved deformable attention works together to efficiently model long-range dependencies. Further, they achieve state-of-the-art performances and fast imaging speed, while still revealing model explainability.
}

%\noindent\textbf
%\subsubsection
\underline{Under-sampled CT reconstruction:}
% MIST-net
MIST-net~\citep{pan2021mist} proposes the multi-domain integrative Swin Transformer network for improved sparse-view CT reconstruction. Considering the information loss in the projection domain and data inconsistency between image and projection domains, it begins by using an encoder-decoder structure to give an initial estimation. Then, a carefully designed High-definition Reconstruction Module is proposed, which is realized through the combination of Swin Transformer~\citep{liu2021swin} and convolutional layers. The post-processing Transformer structure indeed helps in reducing artifacts caused by the aforementioned problems.
% DuDoTrans
With an aim to further investigate the relationship between the sampling nature of projections and the global modeling capability of Transformers, DuDoTrans~\citep{wang2021dudotrans} proposes a Sinogram Restoration Transformer (SRT) Module for projection domain enhancement. The model achieves satisfactory sparse-view reconstruction performance when combined with a similarly designed post-processing module in the image domain.
% RegFormer
{
Targeting to explore a more general prior with the local \& nonlocal regularizations, RegFormer~\citep{xia2022transformer} unrolls the gradient descent algorithm, followed by the designed iterative blocks composed of ConvNet and Transformer structures to model local and nonlocal characteristics, respectively. With such a hybrid architecture embedded into the iterative reconstruction scheme, the model reduces artifacts and preserves image details successfully.}
% hat Fourier Image Transformer (FIT)
FIT~\citep{buchholz2021fourier} instead proposes to process the sinogram and the low-quality reconstruction, realized with Filtered Backprojection~\citep{wang2019machine}, in the Fourier domain with the proposed Fourier Domain Encodings (FDEs). Then the two FDE representations are fed into the Fourier Image Transformer, an encoder-decoder Transformer structure, for predicting all Fourier coefficients. Following that, the inverse Fourier transformation is applied to restore the high-quality reconstruction. The carefully designed FDE representations are shown to reduce the computational burden on conventional Transformer structures.

As illustrated in Fig.~\ref{recon_framework} (a) and (b), these model designs benifit from the combination of ConvNet encoding and ViT media-processing, and achieves image context recovery. Besides, we compare the visualizations of Transformer-based method and pure ConvNet method in Fig.~\ref{recon_vis}. ConvNets gives sharper soft tissue reconstructions and Transformer-based methods recover the whole image better. 
Although the aforementioned sparse-view CT reconstruction methods (also listed in Table~\ref{tab:recon01}, the top part) have been proposed to explore the capability of Transformer versus CNNs in both image and projection domains, few works combines the dense modeling property of Transformer, which helps preserve clinical patterns from input low-quality images and down-sampled projections. Additionally, the limited-angle scenario is overlooked, but the relative consistency between in- and out-of-range projections may be modeled using the Transformer's powerful global-modeling capability. %All these encourage more researchers to explore the field.

\subsection{Medical image enhancement} \label{sec.enhancement} 
%(1.5pages, 1fig, Ce Wang / Junyu)
%\subsubsection{Medical Image Enhancement}
Image enhancement is generally utilized as the subsequent procedure after reconstruction, aiming to remove noise artifacts and enhance medically concerned patterns. Different from high-level vision tasks (e.g., classification), the enhancement process requires maintaining details for the final pixel-level image. For this purpose, the commonly used pooling and strided-convolutional operations in the popular CNN architectures are undesired because of the loss of details. Additionally, the locality nature of convolutional operation constrains its potential to recover with more global contexts. In contrast, Transformer has shown the two attractive key properties:
\begin{itemize}
	\item Transformers facilitate the modeling of global features by promoting a wider reception fields (as shown in Fig. \ref{fig:ERFs}), which establishes the intra-relationships throughout the whole image and provides abundant information for restoration. [Property \hyperlink{M1}{$M_1$}]
	\item Within a whole image, enhancement targets to alleviate artifacts and blur for latter tasks while keeping else context. The involved self-attention mechanism guides the models to focus on the enhancement-related features, and the dense modeling maintains clear context. [Property \hyperlink{M2}{$M_2$}]
\end{itemize}

% \noindent TransCT[MICCAI2021] 
%Based on the early success of various Transformers, 
TransCT~\citep{zhang2021transct} first decomposes a Low Dose CT (LDCT) into high and low frequency components, and denoises the noisy high-frequency component using the basic Transformer structure composed of the MSA and MLP layers, simultaneously assisted by the features of the noise-free low-frequency part. It pioneers the use of Transformer in denoising CT images, and numerically proves that the global modeling ability indeed aids in context preservation.
%\\TED-Net 
To a different extent, TED-Net~\citep{wang2021ted} is proposed and studied in LDCT denoising to explore the convolution-free Transformer structure. Their design makes use of the tokenization and detokenization operations in the encoder-decoder architecture, which aims to entirely evaluate the spatial information extraction capability of the Transformer. Such a design helps understand the difference between the convolution-free features and hybrid features in LDCT denoising, as well as the respective benefits of the two genres in clinical pattern recovery.
%\\ Eformer[ICCV2021Workshop] 
To further combine the global modeling capability of Transformer and the successfully applied residual learning in low-level vision tasks, Eformer~\citep{luthra2021eformer} investigates a residual Transformer that redesigns the residual block in the denoising encoder-decoder architecture with non-overlapping window-based MAS. Additionally, it utilizes strided-convolutions and -deconvolutions instead of downsampling and upsampling operations to preserve image context. The re-design of the previously validated structure, \textit{i.e.}, the residual learning here with Transformer as the basic block instead of convolutional layers, contributes a new perspective to the comparison of Transformer and CNNs.
% TVSRN
{
Although recent works focus on volumetric CT super-resolution, the conducted low-resolution (LR) volumes are most degraded from high-resolution CT volumes, which brings a domain gap between real-LR and such pseudo-LR volumes. \citep{yu2022rplhr} thus releases RPLHR-CT paired real-world LR-HR volumes, and proposes the transformer-based TVSRN for volumetric CT super-resolution. Considering the remote correlation between slices, TVSRN designs an asymmetric encoder-decoder architecture composed of pure transformers. Such a structure enables the long-range dependencies modeling capability and the employed Swin Transformer~\citep{liu2021swin} reduces computational costs.
}
% Task Transformer Network for Joint MRI Reconstruction and Super-Resolution[MICCAI2021]
For obtaining improved super-resolution MR Images, T$^{2}$Net~\citep{feng2021task} specifically designs a task Transformer module in a multi-task learning process of super-resolution and reconstruction. It inserts the module between the iterative recovering processes of the two tasks, and utilizes the module to share informative features. In this way, the super-resolution features are enriched with the low-resolution reconstruction features, resulting in a context devoid of motion artifacts with the detail-preserving Transformer.
% WavTrans
{
WavTrans~\citep{li2022wavtrans} proposes to impose anatomy information from reference contrast MR images for boosting super-resolution performances. They first use Wavelet transforms to obtain details of reference images, followed by a carefully designed hybrid structure composed of ConvNet and Residual Cross-attention Swin Transformer~\citep{liu2021swin} module to extract and upsample images. The introduced transformer explores nonlocal features and promotes long-range dependencies between feature maps.
}

These involved methods, as shown in Fig.~\ref{recon_framework}, take advantage of the hybrid design that globally models the whole image context and locally models the fore/back-ground objects.
In spite of these carefully designed works for image enhancement (also listed in Table~\ref{tab:recon01}), there is still no discussion of the relationship between the intrinsic properties of Transformer structure and image recovery process, leaving the reaction of Transformers on this task as a ``black box" as deep learning. Future architectural design should place a higher emphasis on the interpretability of model mechanisms.

\section{Future Perspectives} \label{sec.future}
%(1 page, 10refs & Kevin/ ALL)\\
Returning back to the initial question: Can Transformers transform medical imaging? The answer is likely dichotomous. This is because Transformer, albeit powerful, belongs to  machine learning, deep learning in particular, and hence it inherits the pros and cons of machine / deep learning. 

The answer is likely positive because it is evident as shown in Section \ref{sec.progress} that Transformer, one of the latest technological advances of deep learning, is picking up its momentum in medical imaging. It is predictable that more and more research will be devoted to innovating the architecture of transform and applying it to more medical imaging tasks. %, and incorporating more medical image domain knowledge into Transformer. 

The answer is likely negative too. %due to the following challenges and limitations that Transformer is yet to address.
%\subsection{Challenges}
In~\citep{zhou2021review}, Zhou \textit{et al.} illustrate some of key traits of medical imaging: multi-modal with a high resolution, non-standard acquisition and data silo, noisy and sparse labeling, imbalanced samples, long-tail disease prevalence, etc. These traits are accompanied with challenges to be solved.

\subsection{Challenges}
{\underline{Annotation intensiveness}. The Transformer or deep learning in general requires large-scale datasets~\citep{cheplygina2019not}.  Empirically, transformer-based models can achieve higher performance trained on larger datasets~\citep{chen2021empirical}, and their performances degrade when data or annotations are sparse. To address the challenge, self-supervised transformers are promising tools. Using unlabeled data, proxy tasks such as contrastive learning and reconstruction can be leveraged to boost representation learning capability of transformers. Self-Supervised SwinUNETR~\citep{tang2022self} and unified pre-training~\citep{xie2021unified}, these medical pre-training frameworks show that training with large-scaled unlabeled 2D or 3D images is beneficial to fine-tuning model with smaller datasets. However, we observe that employing pre-training is computationally exhaustive. Future works can be targeted to simplify and evaluate the efficiency of the pre-training framework and fine-tuning it to smaller datasets.}

{\underline{Data bias, domain adaptation, and model fairness}. In addition to the superior performance, scalability is an advantage brought by transformer models. The robustness to scaling datasets and model complexity are useful properties to address data bias, domain gaps, and fairness. By effectively modeling larger datasets, transformer models~\citep{xie2021unified, tang2022self} can learn diverse datasets, including different modalities, different body components, variant imaging protocols and reconstructions. 
Regarding these domain gaps, there are adaptation methods~\citep{guan2021domain}, which aim to overcome the distribution shift between source and target domains. Meanwhile, the other approach~\citep{caton2020fairness} addresses model fairness. For example, if a model is trained by exclusively male subjects, its performance on female subjects is unknown to the least extent and even worse, the model appears with a gender discrimination. We envision the transformers, with superior scalability, can be used to provide solutions to fairness and social affairs.}

{\underline{Incorporating domain knowledge}. Medical imaging is full of domain knowledge arising from different sources, including anatomical structures, imaging physics, geometric constraints, disease knowledge base, etc. All these knowledge governs the data generation process or serves strong priors for regularized. Visual quantitative analysis of anatomic structures remains a complex task for radiologists. Some of the histomorphometry features of regions of the organs/tissues (e.g. textural or graph features) are poorly adapted for manual identifications~\citep{anandarajah2005validity}. In this study, transformer networks are shown to provide a moderately better solution that achieves consistently robust performance with variate of anatomies. Compared with previous CNNs~\citep{isensee2021nnu}, transformer approaches~\citep{chen2021transmorph,yu2022characterizing} facilitate better derivation of the visual and quantitative results. In addition, efficient modeling is essential for clinical practice in deploying AI networks. We observe, current medical datasets~\citep{wasserthal2022totalsegmentator} can be different in terms of imaging protocols, patient morphology, and institutional variations, which lead to challenging target tasks. Transformer models are yet to unleash the potential to tackle challenges of sensitivity and adapt abnormal primitives.}

%\underline{Domain adaptation}.
{\underline{Task scalability}. Representation learning with medical images is challenging due to it heterogeneity nature~\citep{zhang2015fusing}. Prior studies typically focus solving single medical task, transformer model, especially with self-supervised learning, are superior at learning heterogeneous tasks~\citep{li2022transbtsv2}. The advanced scaling property empowers transformer the ability to tackling multi-domain tasks. In addition, by scaling up transformer networks~\citep{zhai2021scaling}, models can fit variate datasets, researchers can adapt a model at training from a low-data regime~\citep{tang2022self} to larger scales.}

{\underline{Data scalability}. The lack of inductive bias in the original ViT \citep{dosovitskiy2020image} results in subpar performance when trained on a small amount of data (see \hyperlink{M3}{$M_3$} and \ref{sec:ind_bias}). If a large amount of data is available, Transformers can surpass inductive bias by using various pre-training strategies \citep{li2022exploring, Zhai2022Scaling}. In the field of medical imaging, pre-training strategies are also shown merits in improving Transformers' performances \citep{xie2021unified, tang2022self}. However, it is not always practical to collect a large amount of data in medical imaging due to patient privacy concerns and labor-intensive manual annotations. Obtaining a large amount of data for imaging modalities or protocols currently under development is even more challenging. Therefore, it is necessary to develop less data-intensive Transformer models for medical imaging applications by introducing inductive bias into Transformer architectures. Several works have been proposed for both natural images \citep{touvron2021training, liu2021swin, xu2021vitae, d2021convit} and medical images \citep{jose2021medical, gao2021utnet, jang2022m3t, xie2021cotr} to address this issue.} 

{\underline{Black box and interpretability}.
%black box and interpretability: Deep learning is known as a black-box approach and lacks interpretability~\citep{zhang2018visual}. Though Transformer uses self-attention which mimics some human functions, still it is a black box. Given that medical image analysis is keen to a model’s interpretability, it is important to study the interpretability of a Transformer model. Recently, studies~\citep{krishna2022disagreement,kanbrain} are proposed and can be used with transformer backbones for investigating intepretebility.
Deep learning is known as a black-box approach and lacks interpretability ~\citep{zhang2018visual}. Though Transformer uses self-attention which mimics some human functions, still it is a black box and unable to provide insights on how variables are being combined to make decisions. Given that medical image analysis is keen to a model’s interpretability, it is important to study the interpretability of a Transformer model. A common practice to visualize Transformers is to compute relevancy score from single or multiple attention layers. The multi-head self-attention mechanism provide a direct connections among tokens, an intuitive clue on decision-making. There are several methods to visualize transformer in natural images, raw-attention~\citep{hao2021self}, rollout~\citep{xu2022attribution}, GradCAM~\citep{li2022transcam}, LRP~\citep{chefer2021transformer}, etc. Besides, studies ~\citep{krishna2022disagreement,kanbrain} are proposed by using Transformer backbones in investigating interpretability. Specifically, the self-attention on the last layer of ViTs, trained by a teacher-student style, is visualized. The visualization contains object segmentation, which is not clearly observed in supervised ViTs, nor in CNN ~\citep{caron2021emerging}. Recent efforts ~\citep{mondal2021xvitcos,matsoukas2021time} in visualizing vision transformers on medical images conform to conventional methods as those on natural images. From a perspective of practical medical scenario, interpretability is not a property of the algorithm but a model affordance for clinical users ~\citep{chen2022explainable}. The model visualization methods, currently used extensively, are depicting  the interpretability purely computational. It should vary in methods and forms as contexts and users change. It remains a challenging and open problem, which would be an essential factor in convincing physicians and supporting the deployment of algorithms.
}

{\underline{3D modeling}. Most of medical image tasks need to process 3D volumetric data, however, vision Transformer models are known to be computationally intensive and memory-demanding. Efficiently and effectively handling 3D data is a key challenge for adopting Transformers in medical image analysis, UNETR~\citep{hatamizadeh2022unetr}, TransBTS~\citep{wang2021transbts}, CoTr~\citep{xie2021cotr}, nnFormer~\citep{zhou2021nnformer} and many pioneering works have been proposed to address challenges of modeling spatial features. Though, there are still difficulties at preserving 3D positional information between patches in 1D sequences, and loss of local positional information can lead to sub-optimal performance when dealing heterogeneous tissues in 3D medical image segmentation. Current works have shown great progresses in segmentation, classification, detection, registration, reconstruction or enhancement tasks with 3D radiographic images or videos.}

{\underline{Computational complexity}. As seen in \ref{sec:comp_complex}, Transformers are typically computationally complex owing to the computation of self-attention, which is typically quadratic to input image size. While this seems to be less of an issue with natural images, it is a major concern with medical images. This is due to the fact that medical images tend to be far more substantial in size than the size that is common to natural image datasets. For example, a brain MRI image from the BraTS challenge \citep{menze2014multimodal} has a size of $240\times240\times155$, whereas a natural image from ImageNet \citep{deng2009imagenet} has an average size of around $450\times400$. As a result, Transformers used in medical imaging tend to be more compact and trained using a smaller batch size or patched input than their counterparts used for natural images. Many of the existing Transformers used in medical imaging applications are either constructed on top of a SWin Transformer \citep{liu2021swin} (e.g., SWin-UNETR \citep{tang2022self, hatamizadeh2022swin}, SWin-UNet \citep{cao2021swin}, nnFormer \citep{zhou2021nnformer}, and TransMorph \citep{chen2021transmorph}) or rely on a CNN to extract and down-sample feature maps before feeding them into a Transformer (e.g., TransUNet \citep{chen2021transunet} and ViT-V-Net \citep{chen2021vit}). Some exciting explorations have shown that it may be possible to bypass Softmax in order to linearize the computation of self-attention \citep{choromanski2021rethinking, qin2022cosformer, wang2020linformer, xiong2021nystromformer, lu2021soft}, but so far, none of these methods have been applied to medical imaging. We foresee more future research in this area for medical imaging applications.}

%Apart from the above challenges, the properties listed in Section \ref{sec.theory}.D are not fully exploited in medical imaging, other than long-term dependency, dense modeling, scaling behavior, and easy integration. More research is needed to put these remaining properties into real use. Finally, Vision Transformer models are computationally intensive and memory-demanding. Most of existing methods are based on 2D inputs or multi-slice inputs. However, medical images are 3D by nature. How to extend Transformer from 2D to 3D will be an evolving challenge. 

%\subsection{Debate} %over Transformer's properties}
\subsection{{Discussion and concluding thoughts}}
\subsubsection{Debate}
\label{sec:debate}
Despite the promising potential that the Transformers have brought to medical imaging, there have been continuing discussions %in computer vision 
over which properties of Transformers (listed in Section \ref{sec.theory}.D) are particularly beneficial.
\begin{enumerate}
    \item In \citep{raghu2021vision}, the authors discover that the self-attention mechanism enables the early aggregation of global information (i.e., the modeling of long-range dependencies) and that the residual connections help propagate global features throughout the Transformer. %long-range dependencies
    \item \citep{ding2022scaling} believes that the superiority of Transformers is due to their large effective receptive fields, where they experimentally reveal that incorporating convolution operations with large kernels could help close the performance gap between Transformers and CNNs.
    \item Contrarily, in \citep{park2021vision}, the authors observe that the modeling of long-range dependency could hinder the training of Transformers, and experimentally demonstrate that constraining locality rather than employing global computations improves Transformer performance. They argue that data specificity, not long-range dependency, is the critical feature of the self-attention mechanism. Additionally, they suggest that although Transformers encourage flatter loss landscapes, their weak inductive bias results in non-convex losses that disturbs training. % data-specificity; weaker inductive bias affect performance
    \item \citep{liu2022convnet} believe the superior performance of Transformers over CNNs is the result of larger model sizes and training datasets (\textit{i.e.}, the scaling behavior). The authors reveal that by carefully tweaking the CNN designs in accordance with Transformers, CNN outperform Transformers with the help of larger model sizes and training datasets. %scaling behavior (large model size and large dataset)
\end{enumerate}

\subsubsection{Comparative models}

\paragraph{CNNs} {Since the introduction of ViT \citep{dosovitskiy2020image}, many advancements to ViT have attempted to reinstate convolution-like behaviors, e.g., Swin Transformer \citep{liu2021swin}, CVT \citep{wu2021cvt}, CeiT \citep{yuan2021incorporating}, and CMT \citep{guo2022cmt}. Going a different route, efforts have been made to improve CNNs based on the rationale behind the success of Transformers. These CNN models may attain performances similar to those of Transformers. Liu et al. propose ConvNeXt \citep{liu2022convnet}, which modifies a standard CNN with Transformer-inspired components, such as depthwise convolution, layer normalization \citep{ba2016layer}, GELU activation \citep{hendrycks2016gaussian}, and so forth. ConvNeXt exhibits favorable performance and scalability to the competing Transformers while maintaining a CNN-only architecture. In \citep{ding2022scaling}, Ding et al. draw inspiration from the large kernel size of the self-attention operation in a Transformer. They introduce RepLKNet, which substitutes the typically used small convolution kernel (e.g., $3\times3$ or $5\times5$) with large kernels up to $31\times31$. RepLKNet's performance is competitive to that of the competing Transformers, and it demonstrates excellent scalability to large data and model sizes. In a similar fashion, \citep{guo2022visual} present VAN that takes advantage of both convolution and self-attention. VAN employs depth-wise convolutions with large kernel sizes to mimic self-attention, and it outperforms the comparative Transformers and CNNs on several computer vision tasks. As seen from these CNN models, the odyssey of CNN design has recently taken on resembling the characteristics of Transformers. These CNNs have benefited significantly from components like depthwise convolution and large kernel sizes, where the former is analogous to the weighted sum operation in self-attention \citep{liu2022convnet} and the latter resembles the large effective receptive field of Transformers \citep{ding2022scaling}. Similar trends can be observed in the field of medical imaging, where the integration of these Transformer-like components into CNN designs is gaining increased attention \citep{lin2022contrans, liu2022coordinate, jia2022u, han2022convunext}.}

\paragraph{MLPs} {Similar to the aforementioned CNNs, MLP-based models are influenced by Transformers but diverge from Transformers and CNNs. In \citep{tolstikhin2021mlp}, Tolstikhin et al. first demonstrate that, although being beneficial, convolution and self-attention are not required for superior performance. They proposed MLP-mixer, a pure MLP architecture that attains competitive performances on image classification benchmarks. Since then, MLP-mixer has sparked research on developing MLP-based models that can compete with the well established CNNs and Transformers. In general, the architecture of MLP-based models resembles that of Transformers: first, the input image is divided into equal-sized patches; then, the patches are linearly projected to form tokens; and then, two types of MLP layers are repeatedly applied across either spatial locations or embedding channels. On the basis of this concept, models such as ResMLP \citep{touvron2021resmlp}, S$^2$-MLP \citep{yu2022s2}, CycleMLP \citep{chen2022cyclemlp}, Dynamixer \citep{wang2022dynamixer}, Hire-MLP \citep{guo2022hire} have shown promising results in a variety of computer vision applications. MLP-based models have several appealing advantages over Transformers and CNNs, including their simplicity of implementation, more stable training due to the absence of self-attention, their ability to capture long-range interactions, the visibility of the linear layers, and the alleviation of positional embedding \citep{tolstikhin2021mlp, touvron2021resmlp}. However, the use of MLP-based models in the medical imaging field is still in its infancy, with only a small number of models proposed \citep{valanarasu2022unext}.}

{The models discussed above aim to improve upon conventional CNNs and MLPs by making special modifications inspired by the properties of Transformers. Likewise, to develop an efficient model for medical imaging, it is necessary to understand which Transformer properties are particularly advantageous for specific medical imaging applications. In the next section, we discuss briefly the key Transformer properties for each medical imaging application. }

\subsubsection{{Which properties are beneficial for medical imaging applications?}}
It is worth noting that the majority of the findings about Transformers listed in \ref{sec:debate} are derived from image classification tasks. However, the applications of medical imaging are not limited to classification. The properties listed in Section \ref{sec.theory}.D are still under-exploited in all medical imaging applications. \textit{The majority of Transformer-based methods in medical imaging do not investigate the properties adequately and instead take the performance improvement from Transformers for granted.} This paper surveys the applications of Transformers in medical imaging, including segmentation, classification, detection, registration, enhancement, and reconstruction. Yet, it remains to be a question of which Transformer properties are beneficial for which application. Further research is needed to establish the efficacy of these properties and put them into practical use, maybe along the following routes.
%Next, we pioneer the question and discuss the most relevant key properties in Section.~\ref{key_properties} of each task in future tendency:
\begin{itemize}
\item \underline{Segmentation.}
Medical image segmentation is typically with high-resolution, high-dimensional images, which requires modeling capability of visual semantics in dense prediction. That means, unlike the language tokens that used as the basic word sequence in  Transformers, visual contexts in segmentation task vary substantially in scale. ViT-based methods, especially hierarchical structures such as swin Transformer, are designed for efficient modeling of multiscale features (Properties \hyperlink{M1, M2}{$M_1, M_2$}). Furthermore, Transformer-based segmentation networks show futuristic scaling behavior (Property \hyperlink{C1}{$C_1$}) of exploiting large-scale pre-training dataset with self-supervised learning, which provide effective solutions to the difficulties of acquiring expert annotated labels. We believe that the efficiency of modeling hierarchical contexts in medical images, and the effectiveness of pre-training strategy can pave the way for the future work of Transformer-based medical image segmentation. 

\item \underline{Recognition and classification.} 
As the fundamental task evaluated by the original ViT \citep{dosovitskiy2020image}, the properties of Transformers for image classification have been intensively investigated in computer vision. Although medical images are very dissimilar to natural images, Transformers for medical image classification are expected to share similar properties with those deemed beneficial in natural image classification tasks  ({\it i.e.}, Properties \hyperlink{M1}{$M_1$}, \hyperlink{M3}{$M_3$}, \hyperlink{M4}{$M_4$}, \hyperlink{M5}{$M_5$}, and \hyperlink{C1}{$C_1$}). Among these properties, Transformers' superior scaling behaviour (\textit{i.e.}, pre-training using large-scale datasets, Property \hyperlink{C1}{$C_1$}) has been validated for various medical classification applications. In general, the applications of Transformers for medical image classification are mostly limited to 2D, it will be necessary in the future works to expand Transformers to 3D applications given the volumetric nature of most medical images, which is related to Property \hyperlink{C3}{$C_3$}.

\item \underline{Detection.}
Detection is the task of localizing and categorizing lesions and abnormalities. Such a task relies heavily on the comprehension of contextual information about abnormalities and organs. Consequently, the capability of Transformers to model and aggregate long-range dependencies (Property \hyperlink{M1}{$M_1$}) may be the most critical property among other properties for medical image Detection.

\item \underline{Registration.} 
In addition to the flatter loss landscape of Transformer-based registration models (as seen in Fig. \ref{fig:loss-landscape} and Property \hyperlink{M4}{$M_4$}), the large model size of Transformers (Property \hyperlink{C1}{$C_1$}) may also aid in generating accurate high-dimensional vector fields, hence improving registration performance. Moreover, CNN-based models are often of small kernel sizes (\textit{e.g.}, $3\times3$ or $5\times5$), while the deformation or displacement in common registration applications often exceeds their kernel size. Therefore, CNNs may not recognize the proper spatial correspondence until the deeper layers. On the other hand, Transformers aggregate contextual information with large kernels starting from the first layer of the network (Property \hyperlink{M1}{$M_1$}), which may play a crucial role in the improved performance.

\item \underline{Reconstruction.} As discussed, Properties \hyperlink{M1}{$M_1$} and \hyperlink{M2}{$M_2$} have been explored in reviewed works. Further considering the physical imaging system in a real-time clinical diagnosis, the photon noises blur the images and the imaging time troubles the waited patients. Therefore, [Properties \hyperlink{M5}{$M_5$} and \hyperlink{C3}{$C_3$}] need to be further concerned in later model design when introducing ViT in reconstruction.

\item \underline{Enhancement} With the low-resolution or downsampled medical images, the Region of Interest (RoI), such as anatomy boundaries, seems the most important in the diagnosis. Thus, it's worth exhausting to improve the RoI quality while tolerating the else image context less enhanced. Towards this target, exploring the relations from the locality of pixels [Properties \hyperlink{M3}{$M_3$}] is necessary in a Transformer architecture design. Meanwhile, a considerable balance between the global modeling and local modeling of a hybrid model really matters in medical image enhancement.
\end{itemize}

%\underline{Statistical correlation not causality}.

%\input{The lastest advances}

\appendix
\section*{Appendix}
\subsection{Inductive bias}
\label{sec:ind_bias}
{Because of the convolution and pooling operations, CNN architectures impose a strong intrinsic inductive bias. A CNN is analogous to a fully-connected network but with an infinitely strong prior over the weights. The convolution operation constrains the weights of one hidden unit to be equivalent to the weights of its neighbor but spatially shifted. Similarly, the pooling operation constraints that each weight should be invariant to small translations \citep{Goodfellow-et-al-2016}. These priors, known as the intrinsic inductive bias, make CNNs more data- and parameter-efficient \citep{Goodfellow-et-al-2016, scherer2010evaluation}. Additional inductive bias, on top of the intrinsic inductive bias, may further improve the efficacy of CNN-based generative models~\citep{xu2021positional}. Despite inductive bias is of great importance, the original ViT~\citep{dosovitskiy2020image} lacks it since the self-attention operations are global and the positional embedding is the only manually introduced inductive bias. Therefore, ViT yields inferior performance when trained on insufficient amounts of data. However, it is demonstrated that training Transformers on large-scale datasets may surpass inductive bias. When pre-trained using sufficiently large amount of data, Transformers achieve superior performances on tasks with less data \citep{han2020survey, zhai2021scaling, chen2021pre, dosovitskiy2020image, liu2022convnet, naseer2021intriguing}. Alternatively, there have been attempts to introduce locality into Transformers \citep{liu2021swin, xu2021vitae} or distill the inductive bias from CNNs to Transformers \citep{touvron2021training, ren2022co} have been proposed. It has also been shown that combining CNNs with Transformers to construct hybrid models imposes convolutional inductive bias on network architecture \citep{dosovitskiy2020image, d2021convit, wu2021cvt}.}

\subsection{Loss landscapes}
\label{sec:loss_land}
{The sharpness or flatness of a loss landscape is often used as a measure of the trainability and generalizability of a network architecture or optimizer \citep{li2018visualizing, keskar2017on}. A loss landscape is generated relative to the parameters of a neural network. Here, we provide a brief introduction to the computation of loss landscapes and direct interested readers to the corresponding references for further information. We first use a pre-trained model with network parameters, $\theta$, to generate a loss value, which corresponds to the minimum value in the resulting loss landscape. Then, $\theta$ is perturbed using two random direction vectors, $\delta$ and $\eta$, with the corresponding step sizes of $\alpha$ and $\beta$. A loss landscape can be depicted as a plot of the form:
\begin{equation}
    f(\alpha, \beta)=\mathcal{L}(\theta+\alpha\delta+\beta\eta),
\end{equation}
where $\mathcal{L}(\cdot)$ denotes the loss value given the perturbed network parameters.}

{The flatness of a loss landscape translates to how sensitive the network parameters are to the perturbations. There have been substantial theoretical and empirical attempts to understand the relationship between the sharpness of the loss landscape and the generalizability of the neural network \citep{foret2021sharpnessaware, dinh2017sharp, li2018visualizing, dziugaite2017computing, jiang2019fantastic}. Sharp minimizers are more sensitive to noise in the parameter space, resulting in poor generalizability in general \citep{keskar2017on, hochreiter1997flat}. A recent study suggests that ViTs tend to promote flatter loss landscapes than CNNs and thus generalize better on unseen data \citep{park2022vision}. In this work, we empirically confirm this finding by depicting the loss landscapes for CNNs versus Transformers on two tasks, registration and segmentation, as shown in Fig. \ref{fig:loss-landscape}.}

\subsection{Computational complexity of Transformers}
\label{sec:comp_complex}
{Transformers are generally computationally complex, with the self-attention mechanism standing as the main bottleneck. In a self-attention mechanism, each token is updated by attending it relative to all other tokens. Although the computation of self-attention is discussed in length in section 2, we repeat the its equation here for clarity:
\begin{equation}
     {\rm SA}(Q, K, V) = {\rm Softmax} (\frac{Q\times K^\top}{\sqrt{d}}) \times V.
\label{eq:self-att_append}
\end{equation}
Suppose $Q, K, {\rm\ and\ } V$ all have the same size of $n\times d$, where $n$ is the sequence length and $d$ denotes the embedding size, both matrix multiplications in the above equation (i.e., $Q\times K^\top$ and ${\rm Softmax} (\cdot)\times V$) have the complexity of $O(n^2d)$. Consequently, the computational complexity of computing self-attention is quadratic to the sequence size, i.e., $O(n^2)$. In comparison, a convolution operation in CNNs has a linear complexity of $O(n)$. For this reason, training Transformers often requires more time and resources than training CNNs. In light of this shortcoming, modifications to self-attention computation have been proposed to lower its computational complexity. For example, consider Eqn.~(\ref{eq:self-att_append}) without the softmax operation, the complexity of Eqn.~(\ref{eq:self-att_append}) can then be reduced by using the associative property of matrix multiplication, i.e, $Q\times (K^\top \times V)$ as opposed to $(Q\times K^\top) \times V$, where the former has approximately linear complexity while the latter has quadratic complexity. Based on this idea, Choromanski et al. \citep{choromanski2021rethinking} and Qin et al. \citep{qin2022cosformer} linearize the matrix multiplication by avoiding the direct usage of softmax, and afterwards compute self-attention by approximating the softmax attention kernels. Wang et al. \citep{wang2020linformer} propose decomposing self-attention into several smaller attentions by means of linear projections, motivated by the finding that self-attention is of low rank. Xiong et al. \citep{xiong2021nystromformer} reduce the complexity of self-attention computation by leveraging the Nystr{\"o}m method, which samples a subset of columns or rows to approximate a softmax matrix. Similarly, Lu et al. \citep{lu2021soft} propose a Softmax-free Transformer that leverages Gaussian kernel, instead of softmax, to define self-attention. In the meanwhile, Liu et al. \citep{liu2021swin} and Wang et al. \citep{wang2021pyramid} develop hierarchical Transformers that confine self-attention locally rather than globally, thereby reducing complexity and introducing spatial inductive bias that conventional Transformers lack.
}

\bibliographystyle{model2-names.bst}
\biboptions{authoryear}
\bibliography{bib_general}
\end{document}